\pdfoutput=1
\documentclass[10pt,logo,copyright]{nvidiatechreport}
\makeatletter
\renewcommand{\abscontent}{%
  \noindent{\large\bfseries\headingfont Abstract}\par\vspace{0.5em}%
  {\absfont \theabstract\par}%
  \@ifundefined{@keywords}{}{%
    \vskip1em \noindent \keywordsfont Keywords: \@keywords}%
}
\makeatother

\usepackage{algorithm}
\usepackage{algpseudocode}

\usepackage{adjustbox}

\usepackage[authoryear,sort&compress,round]{natbib}

\usepackage[utf8]{inputenc} %
\usepackage[T1]{fontenc}    %

\usepackage{glossaries}
\usepackage[printonlyused,nohyperlinks]{acronym} %
\usepackage{soul, color}    %
\usepackage{parskip}        %
\usepackage{url}            %
\usepackage{hyperref}       %
\usepackage{booktabs}       %
\usepackage{amsfonts}       %
\usepackage{titlesec}
\usepackage{nicefrac}       %
\usepackage{microtype}      %
\usepackage{xcolor}         %
\usepackage[dvipsnames]{xcolor} %
\usepackage{graphicx}
\usepackage{animate}        %
\usepackage{comment}
\usepackage{subcaption}
\usepackage{tabularx}
\usepackage{makecell}
\usepackage{adjustbox}
\usepackage{setspace}
\newcolumntype{M}[1]{>{\centering\arraybackslash}m{#1}}
\usepackage{float}
\usepackage{tikz}
\usetikzlibrary{positioning,shapes,arrows}
\usepackage{amsmath,amsfonts,bm, bbm,leftindex}
\usepackage{multirow}
\usepackage{multicol}
\usepackage{comment}
\usepackage{gensymb}
\usepackage{lipsum}
\usetikzlibrary{arrows.meta, positioning, fit}
\usepackage[para]{threeparttable}
\usepackage{tikz}
\usetikzlibrary{tikzmark}

\usepackage{CJKutf8}
\usepackage[stable]{footmisc}

\usepackage{overpic}
\newcommand{\nvbloxtorch}{\textit{nvblox}\xspace}

\definecolor{backcolour}{rgb}{0.95,0.95,0.92}

\def\eqref#1{equation~\ref{#1}}

\def\1{\bm{1}}

\DeclareMathAlphabet{\mathsfit}{\encodingdefault}{\sfdefault}{m}{sl}
\SetMathAlphabet{\mathsfit}{bold}{\encodingdefault}{\sfdefault}{bx}{n}

\makeatletter
\let\save@mathaccent\mathaccent
\newcommand*\if@single[3]{%
  \setbox0\hbox{${\mathaccent"0362{#1}}^H$}%
  \setbox2\hbox{${\mathaccent"0362{\kern0pt#1}}^H$}%
  \ifdim\ht0=\ht2 #3\else #2\fi
  }
\newcommand*\rel@kern[1]{\kern#1\dimexpr\macc@kerna}
\newcommand*\widebar[1]{\@ifnextchar^{{\wide@bar{#1}{0}}}{\wide@bar{#1}{1}}}
\newcommand*\wide@bar[2]{\if@single{#1}{\wide@bar@{#1}{#2}{1}}{\wide@bar@{#1}{#2}{2}}}
\newcommand*\wide@bar@[3]{%
  \begingroup
  \def\mathaccent##1##2{%
    \let\mathaccent\save@mathaccent
    \if#32 \let\macc@nucleus\first@char \fi
    \setbox\z@\hbox{$\macc@style{\macc@nucleus}_{}$}%
    \setbox\tw@\hbox{$\macc@style{\macc@nucleus}{}_{}$}%
    \dimen@\wd\tw@
    \advance\dimen@-\wd\z@
    \divide\dimen@ 3
    \@tempdima\wd\tw@
    \advance\@tempdima-\scriptspace
    \divide\@tempdima 10
    \advance\dimen@-\@tempdima
    \ifdim\dimen@>\z@ \dimen@0pt\fi
    \rel@kern{0.6}\kern-\dimen@
    \if#31
      \overline{\rel@kern{-0.6}\kern\dimen@\macc@nucleus\rel@kern{0.4}\kern\dimen@}%
      \advance\dimen@0.4\dimexpr\macc@kerna
      \let\final@kern#2%
      \ifdim\dimen@<\z@ \let\final@kern1\fi
      \if\final@kern1 \kern-\dimen@\fi
    \else
      \overline{\rel@kern{-0.6}\kern\dimen@#1}%
    \fi
  }%
  \macc@depth\@ne
  \let\math@bgroup\@empty \let\math@egroup\macc@set@skewchar
  \mathsurround\z@ \frozen@everymath{\mathgroup\macc@group\relax}%
  \macc@set@skewchar\relax
  \let\mathaccentV\macc@nested@a
  \if#31
    \macc@nested@a\relax111{#1}%
  \else
    \def\gobble@till@marker##1\endmarker{}%
    \futurelet\first@char\gobble@till@marker#1\endmarker
    \ifcat\noexpand\first@char A\else
      \def\first@char{}%
    \fi
    \macc@nested@a\relax111{\first@char}%
  \fi
  \endgroup
}
\makeatother

\usepackage{algorithm}
\usepackage{algpseudocode}
\usepackage{subcaption}

\usepackage{xcolor} %

\newcommand{\BlockComment}[1]{%
  \Statex \hspace{\algorithmicindent} \textcolor{gray}{\texttt{#1}}%
}

\newcommand{\sectionowner}[2][]{%
  \textcolor{red}{
    \textbf{Section Owner:} #2%
    \ifthenelse{\equal{#1}{}}{}{ \textit{(#1)}}%
  }
}

\newcommand{\todo}[2][]{%
  \textcolor{blue}{
    \textbf{TODO:} #2%
    \ifthenelse{\equal{#1}{}}{}{ \textit{(#1)}}%
  }
}

\usepackage{pifont}
\usepackage{wrapfig}

\usepackage[nameinlink]{cleveref}
\crefname{equation}{Eq.}{Eqs.}
\crefname{figure}{Figure}{Figures}
\crefname{section}{Section}{Sections}
\crefname{appendix}{App.}{App.}
\crefname{table}{Table}{Tables}
\crefname{algorithm}{Algorithm}{Algorithms}
\crefname{thm}{Thm}{Thm}
\Crefname{thm}{Thm}{Thm}
\crefname{prop}{Prop}{Prop}
\usepackage{longtable}

\title{Isaac Lab: A GPU-Accelerated Simulation Framework for Multi-Modal Robot Learning}

\author{NVIDIA\footnote{A detailed list of contributors and acknowledgments can be found in~\cref{sec:contributors} of this paper.}}

\begin{abstract}
We present Isaac Lab, the natural successor to Isaac Gym, which extends the paradigm of GPU-native robotics simulation into the era of large-scale multi-modal learning. Isaac Lab combines high-fidelity GPU parallel physics, photorealistic rendering, and a modular, composable architecture for designing environments and training robot policies.
Beyond physics and rendering, the framework integrates actuator models, multi-frequency sensor simulation, data collection pipelines, and domain randomization tools, unifying best practices for reinforcement and imitation learning at scale within a single extensible platform.
We highlight its application to a diverse set of challenges, including whole-body control, cross-embodiment mobility, contact-rich and dexterous manipulation, and the integration of human demonstrations for skill acquisition. Finally, we discuss upcoming integration with the differentiable, GPU-accelerated Newton physics engine, which promises new opportunities for scalable, data-efficient, and gradient-based approaches to robot learning.
We believe Isaac Lab’s combination of advanced simulation capabilities, rich sensing, and data-center scale execution will help unlock the next generation of breakthroughs in robotics research.

Isaac Lab is open sourced on GitHub. Code and documentation are available here: \url{https://github.com/isaac-sim/IsaacLab}
\end{abstract}

\begin{document}
\maketitle
\abscontent

\vspace{10pt}
\begin{figure}[htp]
    \centering
    \includegraphics[width=0.99\linewidth]{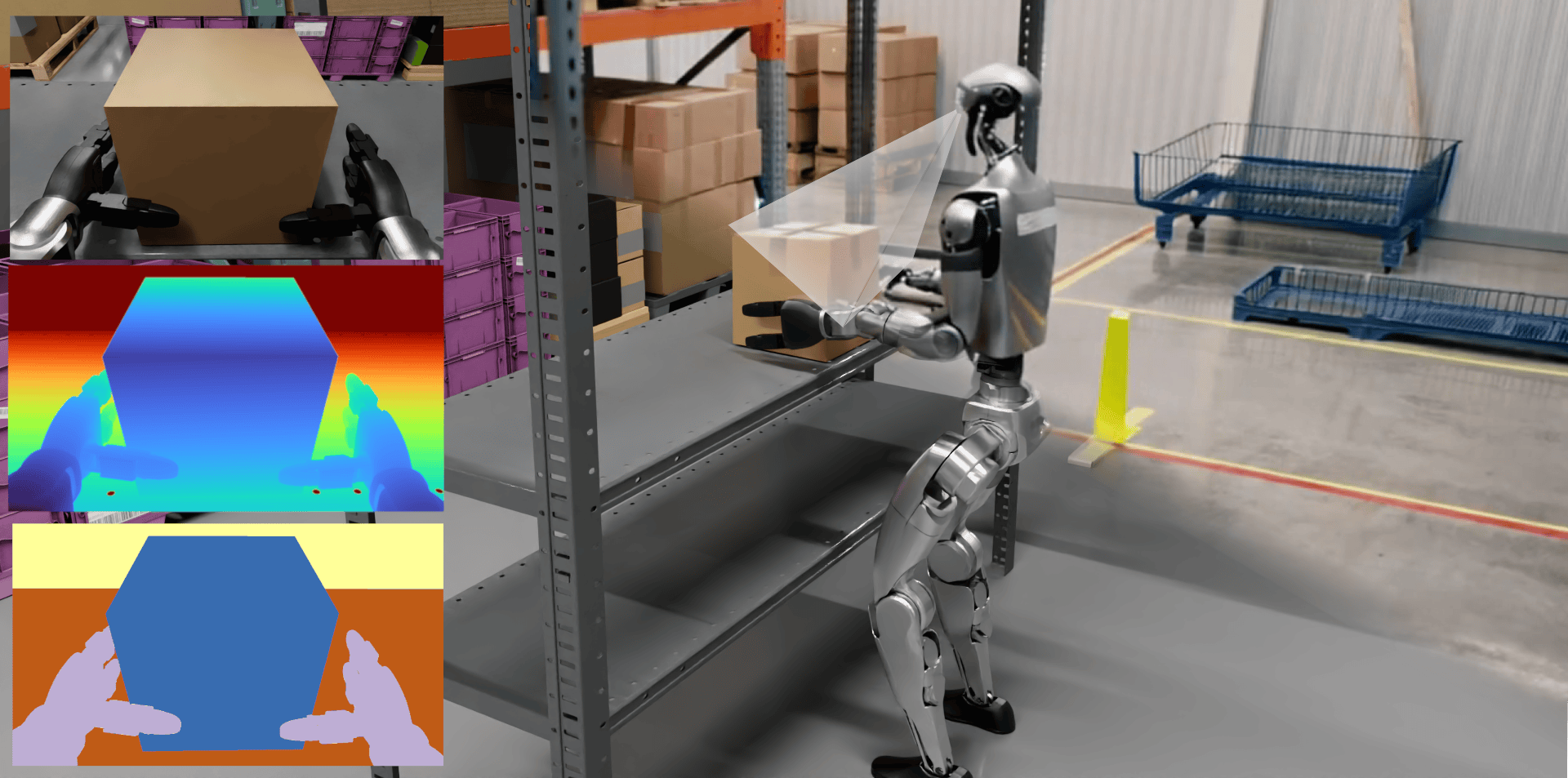}
    \caption{Isaac Lab supports diverse robotic applications with exteroceptive observation inputs. It provides a user-friendly API for experimentation and includes features to facilitate sim-to-real transfer. The framework also supports multiple learning paradigms, including reinforcement learning and imitation learning.}
    \label{fig:hero}
\end{figure}

\clearpage
\section{Introduction} %

The development of robust and intelligent robotic systems increasingly depends on the ability to evaluate their performance in complex real-world environments.
While the physical world remains the definitive testbed, the acquisition of physical interaction data with robots is expensive, time-consuming, and often necessitates specialized instrumentation.
These limitations are especially acute in rare but safety-critical situations.
Events such as high-speed collisions, hardware malfunctions, or navigation in unpredictable human environments are difficult to reproduce and pose significant risks to equipment and human safety.
Moreover, real-world data collection is inherently biased toward normative conditions, leaving robotic systems insufficiently prepared for atypical or extreme situations.
Simulation provides a compelling alternative by offering controlled, reproducible, and risk-free environments in which robotic systems can be developed and evaluated rigorously.
High-fidelity simulators extend these advantages by modeling physics, sensors, and environmental complexity with greater realism.
This enables large-scale data collection, systematic stress testing, and the development of algorithms that transfer more effectively to real-world systems.

The emergence of \acs{GPU}-accelerated, physics-based simulators has democratized robotics research by making scalable training feasible on consumer-grade hardware.
Traditional \acs{CPU}-based simulators~\citep{todorov2012mujoco, coumans2021bullet,Lee2018dart} often struggle to meet the computational demands of high-fidelity physics, complex sensor models, and large-scale parallelization.
Scaling such simulations typically requires clusters with high-core CPUs, which are costly and less widely available.
In contrast, modern GPU-based simulators~\citep{makoviychuk2021isaac,tao2024maniskill3,zakka2025mujoco} exploit massive parallelism to efficiently execute a larger number of concurrent environments, dramatically accelerating the training of complex robotic policies.
By running the agent-environment interaction loop entirely on the GPU, these frameworks avoid inefficiencies associated with frequent CPU-GPU data transfers.
This approach is particularly advantageous for on-policy \ac{RL}, which benefits from large batch sizes during training.
Beyond improving efficiency, GPU-based simulation lowers the entry barrier for researchers, allowing training and development of sophisticated robotic systems without access to specialized supercomputer resources.

A landmark contribution in this space came from NVIDIA Isaac Gym~\citep{makoviychuk2021isaac}, which demonstrated for the first time that end-to-end \ac{RL} for complex robotic tasks could be performed entirely on a single GPU. Isaac Gym uses NVIDIA PhysX, a GPU-accelerated physics engine capable of simulating high-fidelity rigid body dynamics at massive scales. By exposing the physics simulation results directly as PyTorch tensors, Isaac Gym provides a GPU-native pipeline that reduces training times from days to hours. Since its release, the framework has been used successfully in a large number of projects, such as locomotion~\citep{rudin2022learning, agarwal2023legged}, whole-body control~\citep{fu2023deep,he2024omnih2o}, in-hand manipulation~\citep{allshire2021transferring,handa2023dextreme}, dexterous grasping \citep{lum2024get, wang2023dexgraspnet},  and industrial assembly~\citep{narang2022factory}. In doing so, it has established a new standard for scalable, high-performance robotic simulation and laid the foundation for subsequent GPU-based simulation frameworks.

\textbf{Isaac Lab is the natural successor of Isaac Gym}, carrying forward the paradigm of GPU-native robotics simulation into the era of large-scale multi-modal learning. Built on NVIDIA Isaac Sim, Isaac Lab combines RTX rendering for photorealistic, scalable visuals with PhysX for high-fidelity physics simulation. It uses \ac{USD} as the core data layer for structured world authoring, simplifying the design of complex sensor-rich scenes. The physics engine adds numerous enhancements over the one in Isaac Gym, such as filtered contact reporting, mimic joint systems, closed-loop kinematic chains, deformable objects (cloth and soft bodies), and coupled solvers for rigid and deformable bodies. High-throughput GPU-accelerated rendering supports large-scale generation of RGB, depth, and segmentation data, facilitating policy training and sim-to-real transfer using exteroceptive information. Together, these capabilities scale efficiently across multi-GPU and multi-node setups.

Based on the design from~\cite{mittal2023orbit}, Isaac Lab provides users with more than just the outputs of the underlying physics and rendering engines. Since the adoption of Isaac Gym, various paradigms have emerged to facilitate robot learning and sim-to-real transfer. Often, these practices have been independently re-implemented across projects, leading to significant duplication of effort. Isaac Lab addresses this challenge by unifying these practices within a modular and extensible framework for robotics research. Key features include integration of non-linear actuator models, multi-frequency sensor simulation, interfaces for low-level controllers, and tools for procedural environment generation and domain randomization. The framework also supports custom sensors beyond rendering, such as raycast-based \acs{LiDAR}, height scan, and visuo-tactile sensors. At its core, Isaac Lab designs a \emph{manager-based} \acs{API} that organizes environment design into reusable and composable components, allowing consistent workflows across diverse research projects. However, the use of this \acs{API} is optional, and researchers can also structure their simulation environments with simple single-script setups if preferred. In addition, Isaac Lab offers data collection pipelines to record expert demonstrations, access to a wide range of \ac{RL} libraries, and a large suite of robotic environments (shown in~\cref{fig:lab_envs}).

This technical report presents the core capabilities and features of Isaac Lab, providing insight into its design decisions and implementation. It examines the core technologies underlying Isaac Lab (USD for scene authoring, PhysX for high-fidelity physics, and RTX rendering for photorealistic rendering), highlighting their importance for advancing simulation. Building on this foundation, the report details Isaac Lab’s unique enhancements, including custom sensors, actuator models, motion generation pipelines, teleoperation devices, and the environment design framework. It further describes the various learning workflows supported by Isaac Lab and showcases the wide range of robotic applications, from locomotion and navigation to contact-rich manipulation, each benefiting from the framework's capabilities and modularity. Finally, the report concludes with future directions, including the Newton engine~\citep{nvidia2025newton}, and a roadmap for Isaac Lab as a platform for next-generation robotics research.

\begin{mdframed}[
    linecolor=black!20,
    outerlinewidth=1pt,
    roundcorner=10pt,
    innertopmargin=1ex,
    innerbottommargin=.5\baselineskip,
    innerrightmargin=1em,
    innerleftmargin=1em,
    shadow=false,
    shadowsize=6,
    shadowcolor=black!20,
    frametitle={Key contributions of Isaac Lab},
    frametitlebackgroundcolor=backcolour,
    frametitlerulewidth=10pt%
]
    \begin{itemize}
        \setlength{\itemsep}{0.5em} %
        \item \textbf{Modular and scalable framework:} Built on NVIDIA Omniverse, enabling high-fidelity, GPU-accelerated simulation for complex robots and tasks.
        \item \textbf{Advanced sensor simulation:} Supports tiled RTX rendering, Warp-based custom sensors, and physics-based data for rich observation spaces.
        \item \textbf{Seamless teleoperation and data collection:} Integrates spacemouse, VR headsets, and other devices for large-scale demonstration capture.
        \item \textbf{Extensive environment suite:} Provides diverse, ready-to-use environments for reinforcement learning, imitation learning, and sim-to-real research.
    \end{itemize}
\end{mdframed}

\section{Core Simulation Infrastructure}
\label{sec:sim-tech}

\subsection{USD for Robotics}
\label{sec:sim-tech-usd}

The robotics simulation ecosystem remains highly fragmented, with developers managing diverse data sources that include CAD models, kinematic and dynamic descriptions, sensor parameters, and more. Additionally, simulators often rely on multiple specialized attributes for physics and rendering, which further increases the complexity of asset data.
Existing tools illustrate these challenges.
Gazebo~\citep{koenig2004gazebo} utilizes the flat and inflexible XML-based \ac{SDF}.
This format limits the ability to generate large, photorealistic worlds and collaborate on scene variations. It only defines descriptions at the component level, making complex subsystem behaviors, such as closed-loop kinematic chains, difficult to represent.
Other widely used XML-based robotics formats, such as \acs{URDF} and \acs{MJCF}, face similar limitations. 
Modern game engines such as Unity and Unreal are increasingly used in robotics. They provide photorealistic rendering, integrated physics, and better scene authoring capabilities. However, these tools were originally designed for entertainment and use paradigms that differ from traditional robotics workflows. 
AirSim~\citep{airsim2017fsr} attempts to lower the barrier to entry for robotics researchers, but its approach of compiling simulations into a monolithic "game" introduces significant limitations: Any modifications to the simulation often require returning to the underlying game engine.
These limitations highlight the need for a flexible unifying format that can integrate diverse data types, support complex subsystem modeling, and facilitate collaborative workflows for digital content creation.

\begin{figure}
    \centering    \includegraphics[width=\linewidth]
    {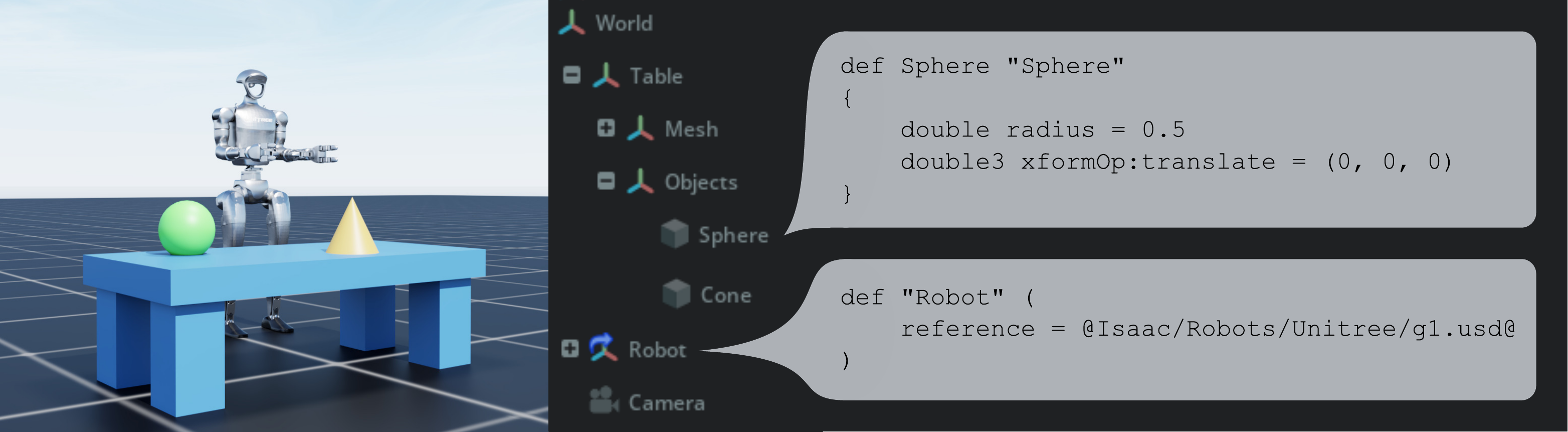}
    \caption{Isaac Lab uses OpenUSD to define rich, complex simulation scenes for robotics. Robots, objects, and sensors are arranged in hierarchical scene graphs, where parent–child relationships manage spatial organization, coordinate frames, and groupings. Meanwhile, simulation-specific schemas capture visual appearance, collision geometry, physical properties, semantic IDs, and sensor configurations.}
    \label{fig:usd-example}
\end{figure}

Addressing these challenges, Open \acl{USD} (OpenUSD)~\citep{OpenUSD} is an open-source format for robust and scalable authoring of complex 3D scenes composed of numerous elements. It provides a comprehensive set of tools and Python/C++ APIs to define, organize, and edit 3D data. USD represents a 3D scene as a hierarchical scene graph (or \emph{stage}), with data arranged in namespaces of primitives (or \emph{prims}). Each prim can contain child prims and have attributes or properties, allowing transformations and properties to be inherited from parent prims. USD's \emph{schema}-based system helps define structured properties for geometry, materials, physics, and more, while \emph{layering} and non-destructive composition facilitate multiple collaborators working on a scene simultaneously without overwriting changes. Additionally, \emph{references} and \emph{instancing} let users combine multiple USD files and reuse repeated elements efficiently. These features make USD a flexible, scalable, and collaborative foundation that naturally ties into robotics simulation workflows. An example of a USD scene leveraging referencing is demonstrated in~\cref{fig:usd-example}.

The \href{https://aousd.org/}{\ac{AOUSD}} is an open, non-profit organization that aims to advance 3D data interoperability through OpenUSD.  
Within this effort, a key development for robotics is the USDPhysics schema. 
The USDPhysics schema extends OpenUSD with a standardized way to describe physical properties such as rigid bodies, collisions, joints, and materials. By providing a common representation for physics simulation, it enables robotics and simulation tools to share and interpret scenes consistently across different engines and workflows.
While these definitions are designed to generalize across multiple simulation backends, OpenUSD also allows a straightforward extension with engine-specific schemas. For example, the PhysxSchema provides parameters used in NVIDIA PhysX, while the MjcPhysics schema, currently under development in collaboration between NVIDIA and Google DeepMind, extends USD for MuJoCo.

Beyond physics, OpenUSD also provides complementary schemas that enrich how scenes can be represented and exchanged across domains. The Semantics schema annotates prims with categorical labels, enabling tasks in perception and learning. Camera prims allow the description of virtual sensors directly within a scene, ensuring that viewpoints and sensor models are preserved consistently across tools. Similarly, Material schemas capture surface properties like textures, reflectance, and friction in a standardized way, bridging the needs of both rendering and physics. 
These schemas unify geometry, dynamics, semantics, sensing, and appearance within a single scene description. This integrated representation overcomes the limitations of existing robotics formats such as MJCF (physics-focused, limited scene richness) and URDF (kinematics/dynamics with Gazebo-specific tags for sensors).

In Isaac Lab, USD serves as the foundation for robotics simulation. While USD’s domain-agnostic design provides flexibility, it necessitates domain-specific conventions to maintain structured scene graphs and develop reliable APIs.
Robotics uses meters and a Z-up convention in the USD stage, where other domains, such as computer graphics, follow loosely defined conventions.
Prims must follow specific hierarchy rules. For instance, two prims with rigid body properties cannot be nested under each other.
Additional considerations, such as instancing, are necessary to ensure efficient stage creation for large-scale simulation scenes.
To address these requirements, Isaac Lab provides USD converters for widely used formats, including URDF, MJCF, and meshes (e.g., OBJ, DAE). It also offers high-level wrappers around USD APIs, allowing users to configure attributes and create prims via simple configuration objects. These wrappers handle stage-modification nuances, such as cloning prims efficiently for large-scale scenes and altering USD properties for domain randomization. 
Additionally, thanks to ongoing \href{https://developer.nvidia.com/omniverse/simready-assets}{SimReady asset} creation efforts, Isaac Lab includes various photorealistic and physically accurate robot and object assets that are ready for use in simulation.

\subsection{Physics Simulation}
\label{sec:sim-tech-physx}

\begin{figure}
    \centering
    \includegraphics[width=\linewidth]{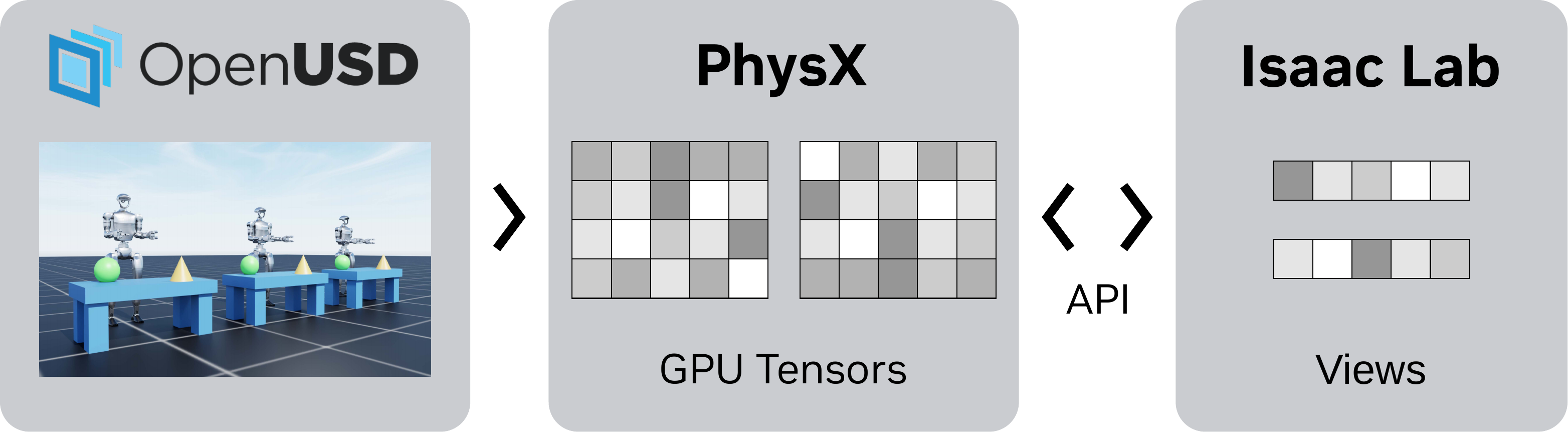}
    \caption{Integration of USD with PhysX in OmniPhysics. The \ac{USD} stage provides a hierarchical representation of all objects and robots in the scene. This scene graph is parsed into PhysX, which allocates GPU tensors representing the internal simulation state. Unlike Isaac Gym, where users accessed raw buffers and manually indexed per simulation object, OmniPhysics exposes these states through the View APIs. These APIs allow read or write access to subsets of objects in the scene while simplifying data management and improving usability.
    }
    \label{fig:omniphysics-views-api}
\end{figure}

The \href{https://developer.nvidia.com/physx-sdk}{\textbf{NVIDIA PhysX SDK}} is an open-source, multi-physics simulation engine designed to meet the demanding needs for robotics and industrial applications.
With the release of PhysX 5, the engine incorporates capabilities from the former \href{https://developer.nvidia.com/flex}{NVIDIA Flex} library, including \ac{FEM}–based soft-body dynamics, as well as \ac{PBD} for liquids and inflatable objects. 
It supports a wide range of simulation types, from rigid and articulated bodies to deformable bodies such as cloth, fluids, and soft bodies. These simulation objects can interact with each other through two-way coupling between specialized solvers. As an example, the \ac{FEM} cloth solver exchanges impulses with the Featherstone-based articulation solver to support efficient and accurate mixed-physics interactions.
Building on this foundation, PhysX has continued to evolve towards robotics use cases, collaborating with robotics engineers and researchers to develop domain-specific features, including emulation of force–torque load cells, advanced actuator and friction models for an articulation's joints, and collision handling improvements such as Signed Distance Field (SDF) methods, which are particularly valuable for high-precision, non-convex geometries encountered in robotic assembly tasks.

PhysX can run on both CPU and GPU, providing flexibility for different simulation scenarios. For high-performance, large-scale robot learning workflows, GPU execution delivers the parallelism and throughput necessary to scale efficiently.
PhysX’s Direct-GPU API provides direct read and write access to simulation state and control data in GPU memory. The resulting CUDA tensors can then be processed efficiently using user-defined GPU kernels for downstream applications, such as computing observations for robotics learning workflows.
This end-to-end GPU pipeline eliminates the performance bottlenecks from CPU–GPU data transfers seen in traditional simulators.
The benefit of this GPU-first simulation approach was first demonstrated for \ac{RL} in Isaac Gym~\citep{makoviychuk2021isaac}, where training times for complex robotic tasks were reduced from days to hours.
It is important to note that only the simulation state and control can currently be accessed directly on the GPU device. Simulation parameters, such as friction coefficients, rigid-body masses, and joint properties, must still be set via the PhysX CPU APIs due to current design constraints.

\textbf{NVIDIA Omniverse Physics} (OmniPhysics) serves as the integration layer for the PhysX simulation engine within NVIDIA Omniverse. It extends the OpenUSD framework by introducing user-defined data types under PhysxSchema, which enables the representation and control of physics simulations. OmniPhysics parses these attributes from USD, maps them into PhysX, runs the simulation, and writes the simulation output back to USD. It also supports state updates by monitoring changes in USD and updating the active simulation accordingly.
In Isaac Lab, where high-performance and large-scale simulation scenes are common, the USD read–write cycle during simulation is often a performance bottleneck and is therefore bypassed. Instead, the simulation data is accessed through OmniPhysics Tensor APIs, which internally rely on PhysX Direct GPU APIs.
For workflows that require rendering, such as vision-in-the-loop training, OmniPhysics also maintains efficient synchronization between the simulation state and the renderer.

\begin{algorithm}[t]
    \caption{OmniPhysics Workflow in Isaac Lab}
    \label{alg:omniphysics-workflow}
    \begin{algorithmic}[1]
        \State Prepare USD stage \Comment{All modifications possible through USD APIs}
        \State Start simulation \Comment{Parse USD stage and initialize PhysX objects}
        \State Create physics simulation views \Comment{Define views via USD prim path patterns}
        \State Access simulation state using Tensor API views \Comment{Read/write simulation state efficiently}
        \If{direct GPU API disabled}
            \State Use standard USD APIs with OmniPhysics monitoring \Comment{Fallback to standard USD workflow}
        \Else
            \State Access PhysX exclusively via Tensor API views \Comment{USD read/write suppressed for performance}
        \EndIf
    \end{algorithmic}
\end{algorithm}

\cref{alg:omniphysics-workflow} summarizes the OmniPhysics workflow in Isaac Lab. The user sets up the USD stage through USD APIs. Isaac Lab provides configurable interfaces and APIs to facilitate programmatic creation of prims. These interfaces support spawning prims of different types, modifying their USD attributes, and efficiently duplicating a prototype scene to multiple environment instances on the USD stage. Once the stage is ready for simulation and the simulation is \emph{played}, OmniPhysics parses the USD stage to create a set of PhysX simulation objects (\eg robots, rigid objects, or static colliders) that mirror the USD scene.
To efficiently scale a prototype environment to thousands of instances, OmniPhysics provides replication APIs that duplicate the prototype’s PhysX simulation objects across many parallel training environments. During this replication process, a mapping is also established between the PhysX objects and their USD prim paths, which is subsequently used to construct \textbf{Tensor API} views.

The OmniPhysics \textbf{Tensor API} presents simulation data as batched, device‑resident arrays organized into views, as shown in~\cref{fig:omniphysics-views-api}.
Instead of operating each physics actor individually, users can interact with a collection of actors through a \texttt{SimulationView}, which links the chosen tensor framework (\ie NumPy, PyTorch, or NVIDIA Warp) to the physics simulation backend. A simulation view can then be used to create specialized views for different types of physics objects, including rigid bodies (\texttt{RigidBodyView}) and articulated systems (\texttt{ArticulationView}).
Views are defined using USD prim path pattern matching.
For instance, if a prototype scene contains a robot prim at "\texttt{/World/envs/env\_0/Robot}", and this scene is cloned $N$ times, then all robots across the $N$ environments can be collected into a single articulation view using the pattern "\texttt{/World/envs/*/Robot}". The resulting view exposes simulation data as arrays with the first dimension corresponding to the number of robots $N$.

In Isaac Lab, the Tensor API-based view classes provide efficient access to underlying simulation data for user-facing asset classes, such as \texttt{Articulation} and \texttt{RigidObject}, as well as physics-based sensor classes like \texttt{ContactSensor} and \texttt{IMU}. These views enable GPU-accelerated batched operations and serve as the foundation for additional asset- and sensor-level functionalities. For example, the \texttt{Articulation} class extends the OmniPhysics \texttt{ArticulationView} to support user-defined actuator models, store default physical properties for domain randomization, and implement efficient buffering mechanisms to minimize redundant read-write operations. These API-level features are described in detail in~\cref{sec:features}.

\subsection{Rendering}
\label{subsec:rendering}

The \href{https://docs.omniverse.nvidia.com/materials-and-rendering/latest/rtx-renderer.html}{\textbf{Omniverse RTX}} renderer simulates RGB cameras and synthetic ground truth sensors, including depth, surface normals, and semantic segmentation, using physically based ray tracing. It supports real-time and offline path tracing, direct lighting, and denoising on top of hardware-accelerated ray tracing. To improve rendering efficiency, the RTX renderer leverages \href{https://developer.nvidia.com/rtx/dlss}{\ac{DLSS}}, which upscales images from a lower internal resolution using high-quality temporal super-resolution, making it well-suited for high-resolution perception tasks.
The RTX renderer allows users to selectively disable \ac{DLSS} in very low-resolution scenarios where there is insufficient visual data for the upscaler to make use of. \ac{DLSS} only affects RGB outputs; other ground truth data, such as depth and segmentation, are always rendered at their native resolution.

\begin{figure}[t]
    \centering
      \begin{minipage}{0.7\linewidth}
        \includegraphics[width=\linewidth]{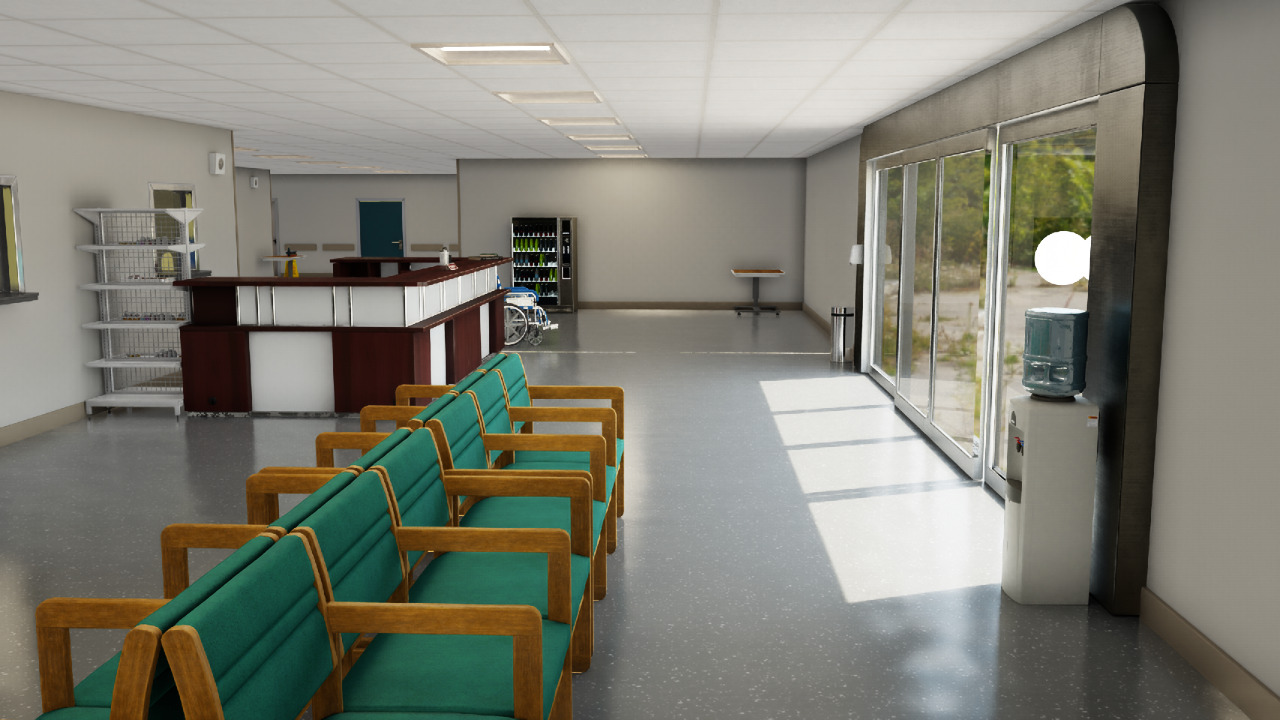}
    \end{minipage}\hfill
    \begin{minipage}{0.29\linewidth}
        \includegraphics[width=\linewidth, trim ={0 0 900 0}, clip]{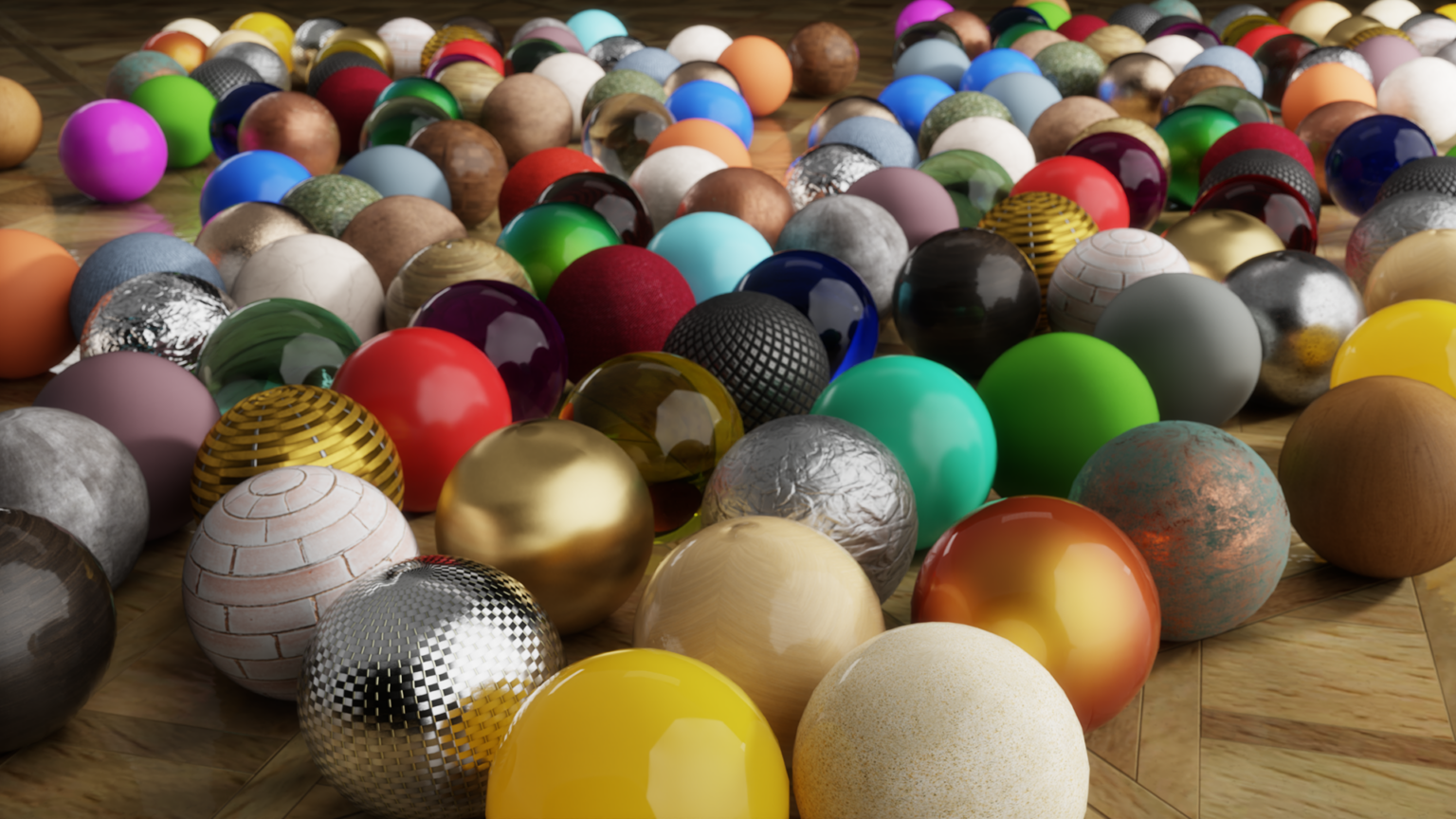}
    \end{minipage}
    \caption{Photo-realistic rendering in Isaac Lab using the Omniverse RTX renderer, demonstrating high-quality ray tracing with complex physically-based materials authored using NVIDIA’s MDL. The rendering showcases realistic effects such as reflections and refractions, resulting in visually rich and high-quality scenes.}
    \label{fig:rtx-combined}
\end{figure}

An important aspect of high-quality rendering is applying the right materials to USD prims. Material schemas for USD prims are authored using NVIDIA's \href{https://www.nvidia.com/en-us/design-visualization/technologies/material-definition-language/}{\ac{MDL}}. \ac{MDL} provides a flexible and powerful framework for describing complex, physically-based materials (e.g., OmniPBR, OmniGlass) with realistic details such as reflections, refractions, and surface patterns. These materials enable high-quality visual results in rendered scenes, as illustrated in~\cref{fig:rtx-combined}. Additionally, to support semantic segmentation, USD prims can also be annotated with semantic information, such as object class or instance identifiers. This metadata allows per-pixel instance and class labels to be rendered alongside RGB images, which is fully compatible with tiled rendering in large-scale, multi-camera setups.

\begin{figure}
\centering
    \includegraphics[width=0.95\linewidth]{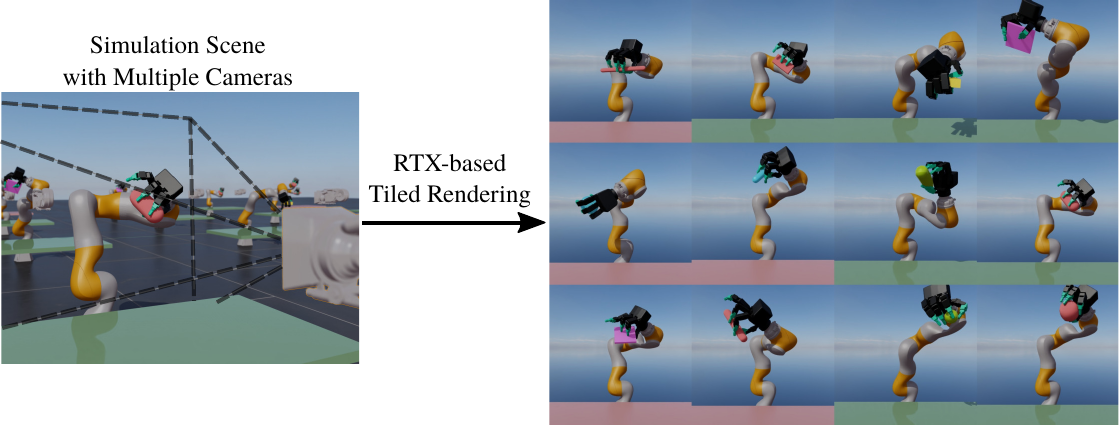}
    \caption{Tiled rendering of multiple simulated environments. Each environment has a separate camera, and their outputs are spatially tiled into a single GPU frame-buffer. The deterministic layout allows efficient reconstruction of per-environment observations without costly host–device transfers.} 
\label{fig:tiledrendering-figure}
\end{figure}

To support thousands of cameras in parallel simulation environments, Isaac Lab uses the tiled rendering pipeline of the RTX renderer. This method batches multiple cameras into a single render pass by spatially arranging them as tiles within the GPU framebuffer. Each camera preserves its own intrinsics and pose, and the deterministic tiling layout enables efficient reconstruction of per-environment tensors without incurring costly host–device transfers. This design is essential for large-scale data generation workloads, such as vision-in-the-loop \ac{RL}, as it scales sensor throughput linearly with GPU resources, minimizes latency, and synchronizes observations across environments for policy training (see \cref{fig:tiledrendering-figure}). Although active sensors such as LiDARs and radars are already supported by the Omniverse RTX renderer, their integration with tiled rendering is forthcoming. As an alternative, the \texttt{RayCaster} sensor uses NVIDIA Warp operations for ray-casting, as described in~\cref{subsubsec:sensor-warp-based}.

\begin{wrapfigure}{R}{0.45\textwidth}
    \centering
    \vspace{-10pt}
    \includegraphics[width=\linewidth, trim={100, 100, 100, 100}, clip]{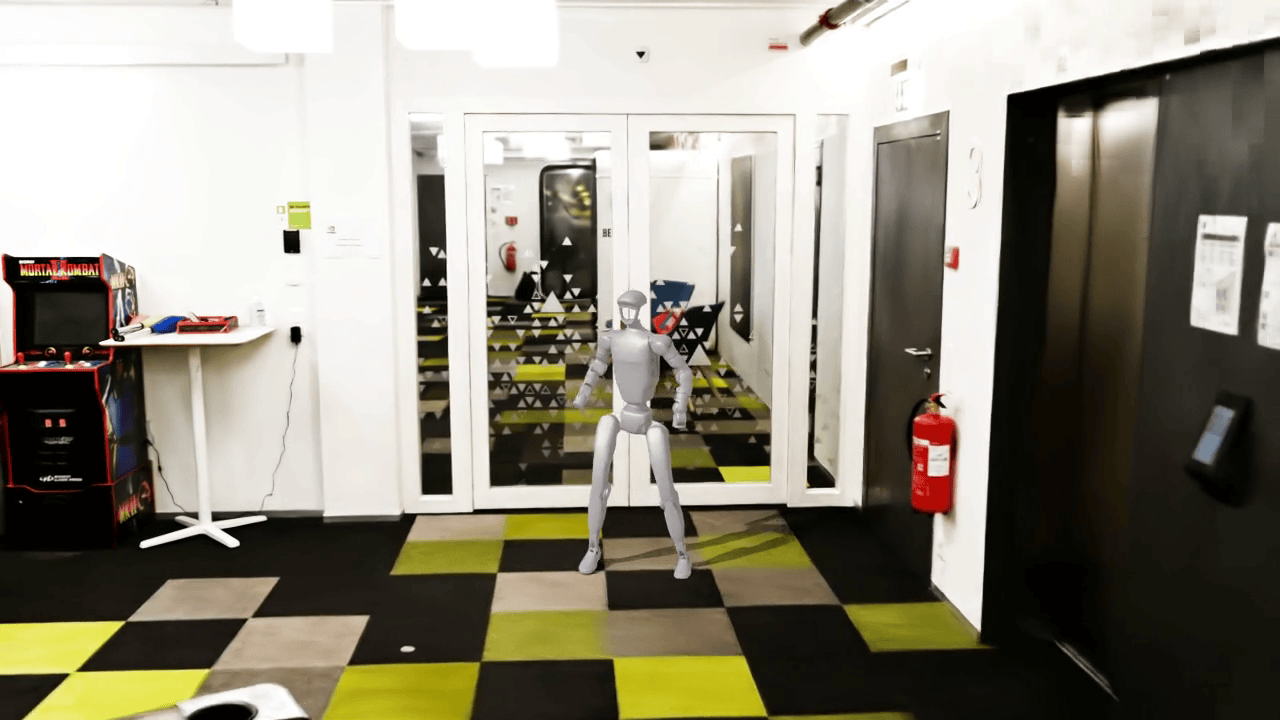}
    \caption{3D Gaussian rendering combined with mesh rendering, with shadows from the mesh affecting the Gaussian scene.} 
    \label{fig:nurec-figure}
\end{wrapfigure}
The RTX renderer can also perform 3D Gaussian rendering via \href{https://docs.omniverse.nvidia.com/materials-and-rendering/latest/neural-rendering.html}{Omniverse NuRec}, supporting realistic reconstructions of the real-world scenes using approaches such as 3D Gaussian Splats (3DGS) by~\cite{kerbl3Dgaussians} and 3D Gaussian Unscented Transforms (3DGUT) by~\cite{wu20253dgut}.
These Gaussian primitives integrate seamlessly with ray-traced geometry in the RTX renderer, allowing robots and synthetic objects to operate within photorealistic, reconstructed environments. This approach improves visual realism and facilitates policy transfer without requiring hand-crafted assets.
\cref{fig:nurec-figure} shows an early example of this integration from~\cite{liu2025compass}.

Isaac Lab uses the RTX renderer to implement the \texttt{TiledCamera} sensor class, which batches the rendering output for learning pipelines. It supports specifying and retrieving camera poses in multiple conventions, such as those used in ROS and computer graphics. The class also ensures that sensor data is updated at a specified frequency to match real-world sensors. Since the RTX renderer settings affect performance, Isaac Lab provides users with presets that trade off between quality and speed, and exposes rendering parameters through configuration objects for further customization. Furthermore, Isaac Lab employs \href{https://docs.omniverse.nvidia.com/extensions/latest/ext_replicator.html}{Replicator API} to randomize \ac{MDL} materials on prims and scene lighting. 
Together, these features enable scalable, high-fidelity rendering with multi-camera observations while supporting realistic variations in textures and illumination.

\section{Isaac Lab Design and Features}
\label{sec:features}

\begin{figure}
    \centering
    \includegraphics[width=\linewidth]{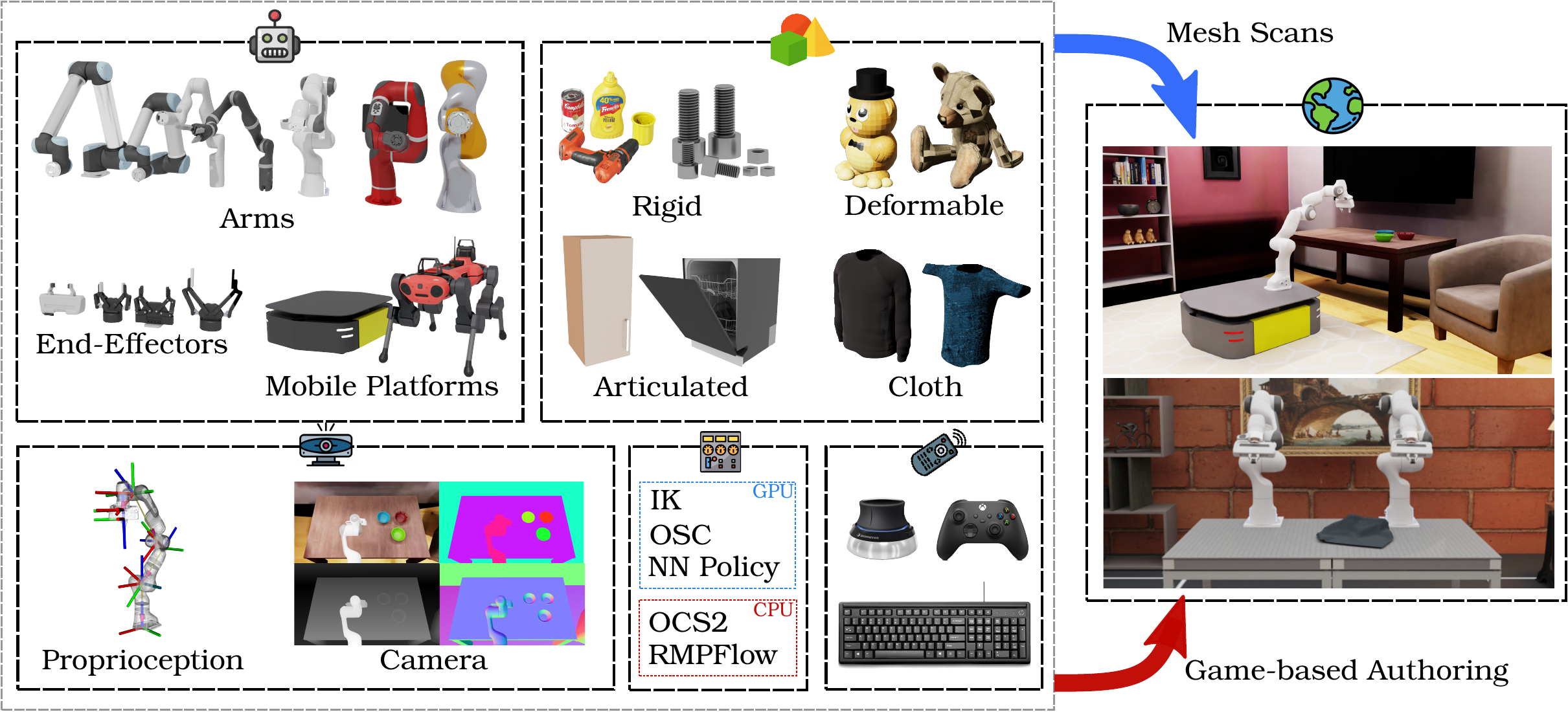}
    \caption{Isaac Lab supports diverse asset types (articulated robots, rigid and deformable objects), sensor modalities (proprioception, RGB/depth images, height scans), controllers (\acs{IK}, cuRoBo), and teleoperation devices (keyboard, spacemouse, and \acs{XR}). Environments can be created using USD scenes from custom mesh scans, game-based authoring, or programmatic generation. All these features are integrated into task definition frameworks to facilitate robot learning at varying levels of abstraction.}
    \label{fig:features-overview}
\end{figure}

Building on the core technologies introduced in~\cref{sec:sim-tech}, Isaac Lab brings state-of-the-art simulation capabilities to robot learning researchers.
While these general-purpose technologies expose numerous low-level states and properties, this flexibility can create a steep learning curve that alienates non-expert users.
Isaac Lab aims to lower this entry barrier through a modular, integrated framework that leverages the latest advances in physics and rendering.
Its interfaces are specialized for robot learning, simplifying environment design, and introducing features essential for facilitating deployment to physical robots. 
The framework adopts a bottom-up design philosophy, starting with modeling complex actuator dynamics, asynchronous sensing and control, realistic sensor noise, and environmental uncertainties, and building upward to high-level task abstractions and robot learning interfaces.
As highlighted in~\cref{fig:features-overview}, it supports different robotic platforms and object types, low-level actuators and sensor models to aid sim-to-real transfer, and integrates peripherals for data collection -- enabling both reinforcement learning and imitation learning.

\subsection{Assets}

Assets correspond to any physics-enabled object that can be added to the simulation. These include rigid objects, articulations, and deformable objects. Each asset provides a high-level interface that wraps around low-level USD and OmniPhysics View APIs. These interfaces manage the spawning of objects into the simulation scene and provide structured access to their physical states. Although the underlying TensorAPI views expose direct read and write access to simulation data, Isaac Lab introduces the additional abstraction layer to ensure a unified and reliable way of interacting with the simulation.

The asset interface consolidates attributes that are not directly managed by TensorAPIs but are essential for robotics. These include user-defined properties, such as the default initial state and collision filter group, which are specified through configuration instances, as well as derived quantities, such as the velocity of a robot’s torso in its local frame, which are stored as data parameters accessible during simulation.

Additionally, to minimize performance overhead, the interface employs a lazy-update mechanism for simulation state retrieval. While the TensorAPI views allow fetching data for a subset of objects, this operation requires copying the queried attributes into a contiguous tensor, which can introduce an overhead if performed repeatedly. With the lazy-update approach, data for each attribute is fetched only upon its first access after a simulation step, and subsequent accesses within the same step use the cached values.

\begin{figure}
    \centering
    \includegraphics[width=0.95\linewidth]{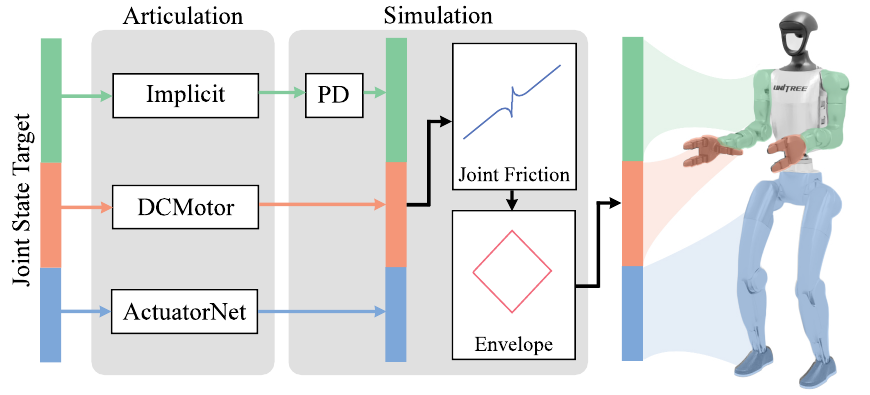}
    \caption{Custom actuator integration in Isaac Lab. Different robot joints can use different actuator models. Implicit actuators rely on the simulator’s PD controller, while explicit actuators (\eg neural networks) process commands directly to generate joint efforts that are set into the simulator.}
    \label{fig:actuators}
\end{figure}

For articulated systems, the same interface provides a mechanism for defining and integrating custom actuator models. While the physics engine includes an internal joint-level PD controller, real-world joint dynamics often exhibit characteristics that differ from this model, as described in~\cref{sec:actuators}. At every simulation step, the \texttt{Articulation} interface processes the user-specified joint targets through these custom models and applies the resulting commands to the physics engine, ensuring accurate and consistent simulation of the intended actuator behavior.
As shown in~\cref{fig:actuators}, the interface supports configuring different actuator models for distinct joint groups within the system to model heterogeneous actuation characteristics flexibly.

Extending the actuator concepts from articulated systems to thrusters in multi-rotor platforms allows consistent modeling of thrust and torque generating actuators for realistic flight dynamics. Each rotor is treated as an individual actuator that converts control inputs, such as RPM setpoints, into forces and moments that are applied to the vehicle’s body frame.
The \texttt{MultiRotor} interface supports multiple control layers, ranging from direct RPM commands to high-level acceleration, velocity, or position setpoints via  geometric controllers inspired by~\citet{lee2010control}. This multi-layered control architecture accommodates both low-level controller development and high-level planning research~\citep{kulkarni2025aerial}.

In addition to rigid objects and articulated systems, Isaac Lab supports deformable objects under the same asset interface.
However, unlike rigid bodies, the simulation state of a deformable object is represented through mesh point attributes rather than a single transform. The interface abstracts this complexity by caching a default initial state for the mesh at the start of the simulation and converting user-specified transforms into updates for individual mesh points. It also supports partial kinematic control, enabling targeted manipulation of specific nodes while allowing the remainder of the object to evolve naturally under full physical simulation. Furthermore, the interface provides access to derived physical quantities, including deformation gradients, stress tensors, and element-wise rotations, providing detailed information for robot learning.

\subsection{Actuators}
\label{sec:actuators}

In Isaac Lab, actuators are the interface between desired joint actions and articulation motion. Actuators provide a control loop over desired motion and joint intrinsic model definitions. All actuators provide interfaces for defining joint friction using two different models. The first model is a simple Coulomb friction model with a constant coefficient of friction. The second model provides a stiction model with a static slip threshold, dynamic friction, and viscous friction coefficients. Actuators also have an armature value that represents the motor inertia and is used to affect the dynamic response of the actuator. Limits for both velocity and effort can also be provided to better approximate real-world limitations. Actuators are separated into two different classes: implicit and explicit actuators.

\subsubsection{Implicit Actuators}

Implicit actuators utilize the NVIDIA PhysX joint PD controllers. These controllers can be used to track both position and velocity by specifying the stiffness and damping terms of the PD controller. A feedforward effort can also be provided from the desired control action. Implicit actuators can utilize the velocity and effort limits to apply inequality constraints to the simulation solver. This can introduce issues with solvers if they become too restrictive. These joint-level controllers are typically more stable than explicit actuators, as they are iteratively solved in step with physics. They also tend to be more accurate at low sampling rates than explicit actuators.

\subsubsection{Explicit Actuators}

Explicit actuators operate in a similar manner as the implicit actuators, but they better approximate the discrete implementation of joint controllers on hardware, especially at higher physics sampling rates. Given a desired joint action, explicit actuators compute the applied effort that is sent to the physics solver. This introduces the inherent numerical challenges of discrete-time control. Ways to mitigate this include traditional control stability analysis, as well as actuator armature. Explicit actuators can utilize the solver limits for velocity and effort, but can also apply limits that are used by the explicit effort calculations. A variety of different explicit actuators are available in Isaac Lab, including:

\begin{itemize}
    \item \textbf{Ideal PD} actuators are the simplest form of explicit actuators. Similar to the implicit actuator, the Ideal PD actuator implements a PD feedback controller with feedforward effort, using specified stiffness and damping parameters. However, the Ideal PD actuator operates in a discrete-time formulation. It can also apply an effort limit, clipping the computed torque before passing it to the solver.
    \item \textbf{DC Motor} actuators utilize the same control and intrinsic parameters as the Ideal PD actuator, but provide an additional velocity-dependent effort limit model that approximates a four-quadrant DC motor torque speed curve. They do not simulate the electro-mechanical dynamics of a DC motor.
    \item \textbf{Delayed PD} actuators build upon the Ideal PD actuator by providing a mechanism for simulating communication delay in the control system by buffering desired actions by a configurable number of simulation steps. These delays are important to simulate when approximating the communication delays on real distributed systems. The Spot application described in (\cref{subsec:app-locomotion}) is an example use case of the Delayed PD actuator. 
    \item \textbf{Remotized PD} actuators are used to simulate actuators with joint position-dependent effort limits. Some robots have remote actuators that utilize linkages to transfer power to distal joints. This type of actuation results in the nonlinear transformation ratios and position-dependent effort limits that are emulated by the Remotized Actuator. An example use of this actuator is the Spot knee actuator (\cref{subsec:app-locomotion}). 
    \item \textbf{Neural Net} actuators utilize a trained neural network for joint control rather than the PD control loop of the Ideal PD actuator. These models can be trained on hardware data to better emulate the dynamic response of the actuator in the real world. They are often coupled with the DC motor torque-speed curve effort limits. Isaac Lab has examples for both Long Short-Term Memory LSTM actuators ~\citep{rudin2022learning} and Multi-layer Perceptron (MLP) actuators ~\citep{hwango2019actuator} models.  
\end{itemize}

\subsection{Sensors}
\label{subsec:sensors}

\begin{figure}
    \centering
    \includegraphics[width=\linewidth]{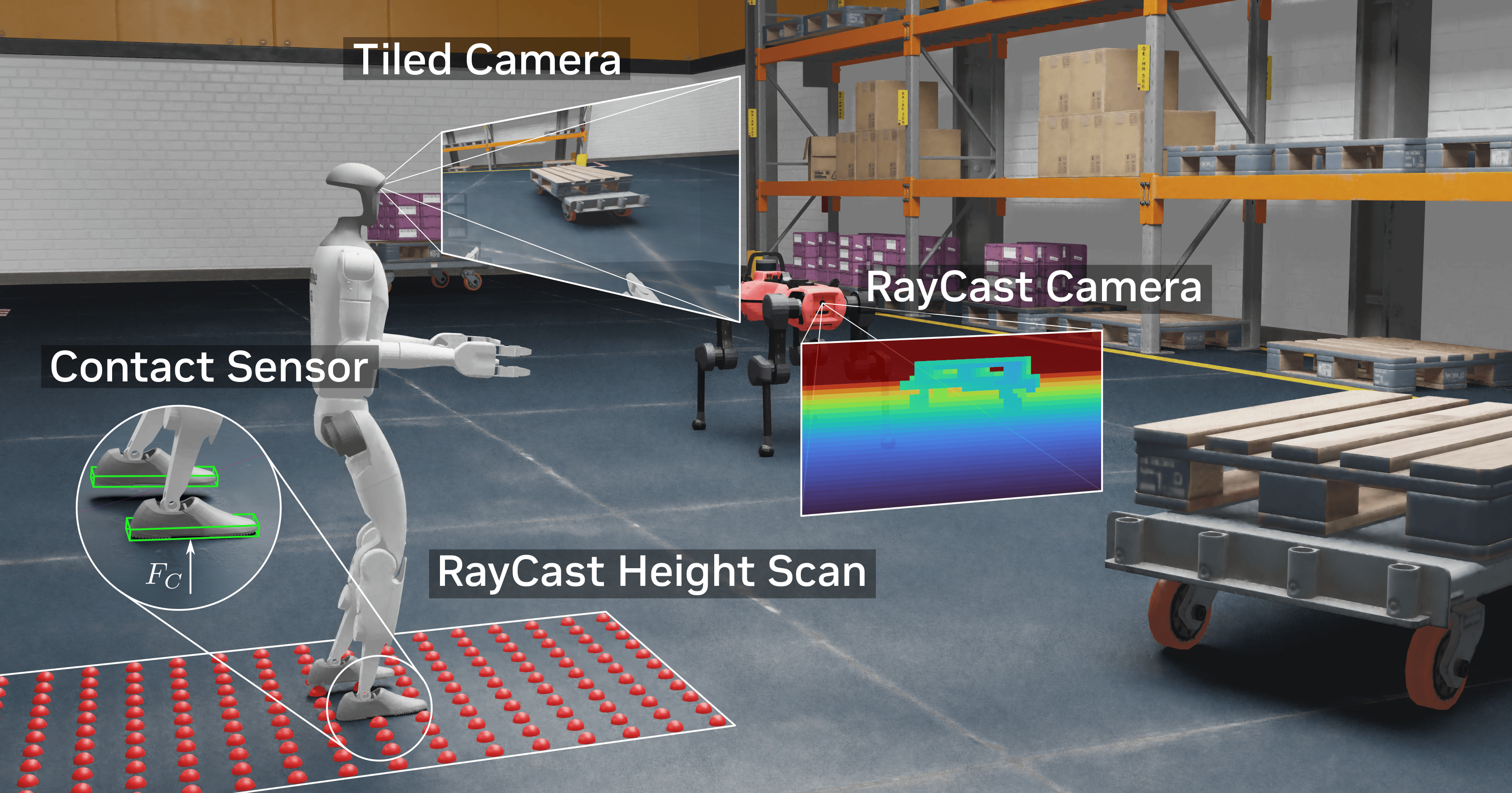}
    \caption{Overview of the Isaac Lab sensor suite, including physics-based sensors (IMUs, frame transformers, contact sensors), parallelized camera implementations, and warp-based raycasting for geometric perception. The figure shows example outputs in a photorealistic scene.}
    \label{fig:sensors}
\end{figure}

Sensors are fundamental for robot learning, as they enable agents to perceive both their own physical state and the surrounding environment.
Isaac Lab provides three major sensor classes: \emph{physics-based}, \emph{rendering-based}, and \emph{warp-based} sensors, as shown in~\cref{fig:sensors}.  
Similar to assets, all sensors are unified under a common interface, simplifying instantiation, configuration, and runtime integration. The interface also supports flexible sensor update frequencies in parallelized environment settings.  
This avoids the typical unrealistic assumption of synchronized sensing and control, and better approximates real-world conditions where sensors operate at different rates and experience communication delays.

\subsubsection{Physics-based Sensors}
\label{subsubsec:sensor-physics}
While many physical signals can be extracted from the articulation and rigid object data classes, certain signals require additional handling, and use of dedicated physical sensors. In particular, Frame Transformers, Inertial Measurement Units (IMUs), and Contact Sensors can be used to collect relevant physical data across these modalities.

The \textbf{Frame Transformer} sensor, while not a direct analog to a physical sensor, provides a convenient method for computing the poses of multiple target frames relative to a specified source frame. Notably, it batches transformation computations, significantly reducing computational overhead compared to performing transformations individually. Users can also define offsets for both the source and target frames, enabling precise specification of transformations relative to key reference points — such as a body’s center of mass or known positions of motion capture markers.

An \textbf{IMU} sensor provides rich state information about moving bodies, whether they are connected in an articulation or on their own as a single rigid body. In the real world, simple IMUs traditionally collect angular velocity via a gyroscope and linear acceleration via an accelerometer. Others can collect orientation information via a magnetometer or estimation of the direction of gravity. Advanced units may come with onboard sensor fusion to provide other components of the inertial state. The IMUs in Isaac Lab provide linear and angular components of pose, velocity, and acceleration. Accelerations are computed via finite difference, which can introduce noise — especially at low physics rates. Following real-world sensors, the IMUs in Isaac Lab provide measurements in the sensor’s local frame relative to the world and can be placed on any rigid body with an arbitrary offset. To simulate real-world behavior, modifiers such as observation noise and signal drift can be applied. Additionally, a built-in measurement provides the projection of world gravity in the sensor frame.

The \textbf{Contact Sensor} captures interactions between rigid bodies with colliders. In its basic form, it reports the net normal contact forces acting on the assigned bodies, which can be used to measure data such as ground reaction forces for locomotion or grasping forces for manipulation. For more fine-grained measurements, users can specify filter bodies to separate forces between different contact pairs. The sensor also offers optional outputs, including temporal information such as contact lengths and the intervals between them, as well as the average point of contact between the body the sensor is attached to and the filter bodies. Since contacts between rigid bodies are inherently discrete and depend on the observation rate, the sensor can maintain a short history of contact events. Exposing this history to policies can provide richer feedback and improve learning performance.

\subsubsection{Rendering-based Camera}
\label{subsubsec:sensor-rendering}

Rendering-based sensors emulate real-world cameras, which remain among the most informative sensors for robotic perception.  
Isaac Lab supports both \textbf{Pinhole} and \textbf{Fisheye} camera models, with outputs including RGB, depth, semantic labels, instance segmentation, surface normals, and motion vectors. Cameras can be configured using standard parameters (e.g., focal length, aperture) or intrinsic matrices. Since orientation conventions vary across frameworks, Isaac Lab supports multiple frame definitions: \textit{world} (x-forward, z-up), \textit{ROS} (z-forward, -y-up), and \textit{OpenGL} (-z-forward, y-up). For photorealistic rendering, Isaac Lab relies on RTX-based pipeline, as described in~\cref{subsec:rendering}.

Each camera sensor generates a \emph{render product}, which first uses path-tracing to generate a rough image and then undergoes post-processing steps such as denoising and anti-aliasing to produce high-quality outputs. 
The \textbf{USD-Camera} implementation assigns one render product per environment, enabling highly accurate images that capture illumination, shadows, and reflections. This is well-suited for data generation tasks where photorealism is critical and parallelization is less of a concern.
In contrast, learning tasks often require frequent image generation across thousands of environments, where individual render products become a computational bottleneck. To address this, the \textbf{Tiled-Camera} offers a parallelized implementation that aggregates all camera data into a single tiled render product. While this approach may slightly reduce photorealistic quality due to post-processing being optimized for single images rather than tiled layouts, it provides significant speedups. Ongoing improvements in tiled post-processing aim to further close this quality gap.  

\subsubsection{Warp-based Ray-Caster}
\label{subsubsec:sensor-warp-based}

Distance queries are a key component in many robotic tasks and policy learning pipelines. In real systems, such measurements are typically provided by LiDAR sensors or derived from environment height maps~\citep{miki2022learning}. In Isaac Lab, we implement these queries through the \textbf{RayCaster} sensor, which leverages \href{https://github.com/NVIDIA/warp}{NVIDIA Warp} to achieve lightweight and highly parallelized geometric computations on the GPU.  

The \texttt{RayCaster} can be configured with flexible raycast patterns to emulate a variety of sensors, including height scanners, solid-state LiDARs, and rotating LiDARs. Rays can be cast against arbitrary dynamic meshes, enabling users to include or exclude specific actors in the scene. Internally, scene geometry is stored as Warp meshes, and their poses are synchronized through PhysX views to support dynamic environments. The ray origins and directions, defined by the chosen pattern, together with the current mesh transforms, are then passed to Warp kernels for efficient GPU-based raycasting.
All raycast results are computed at a single time instance, eliminating temporal distortion effects that can occur in real-world rotating sensors where individual rays are captured at different timestamps.

Providing a pinhole raycast pattern enables a \textbf{RayCasterCamera}, which integrates with the standard camera interface to provide depth images. RGB and semantic outputs are currently under development. Compared to USD- or Tiled-Cameras, raycast-based sensors prioritize efficiency over photorealism by omitting rendering effects, making them particularly well-suited for large-scale geometric training scenarios. \\

\subsubsection{Visuo-Tactile Sensor}
\label{subsubsec:visuo-tactile}

\begin{wrapfigure}[20]{r}{0.37\textwidth}
  \centering
  \vspace{-20pt}
  \includegraphics[width=\linewidth]{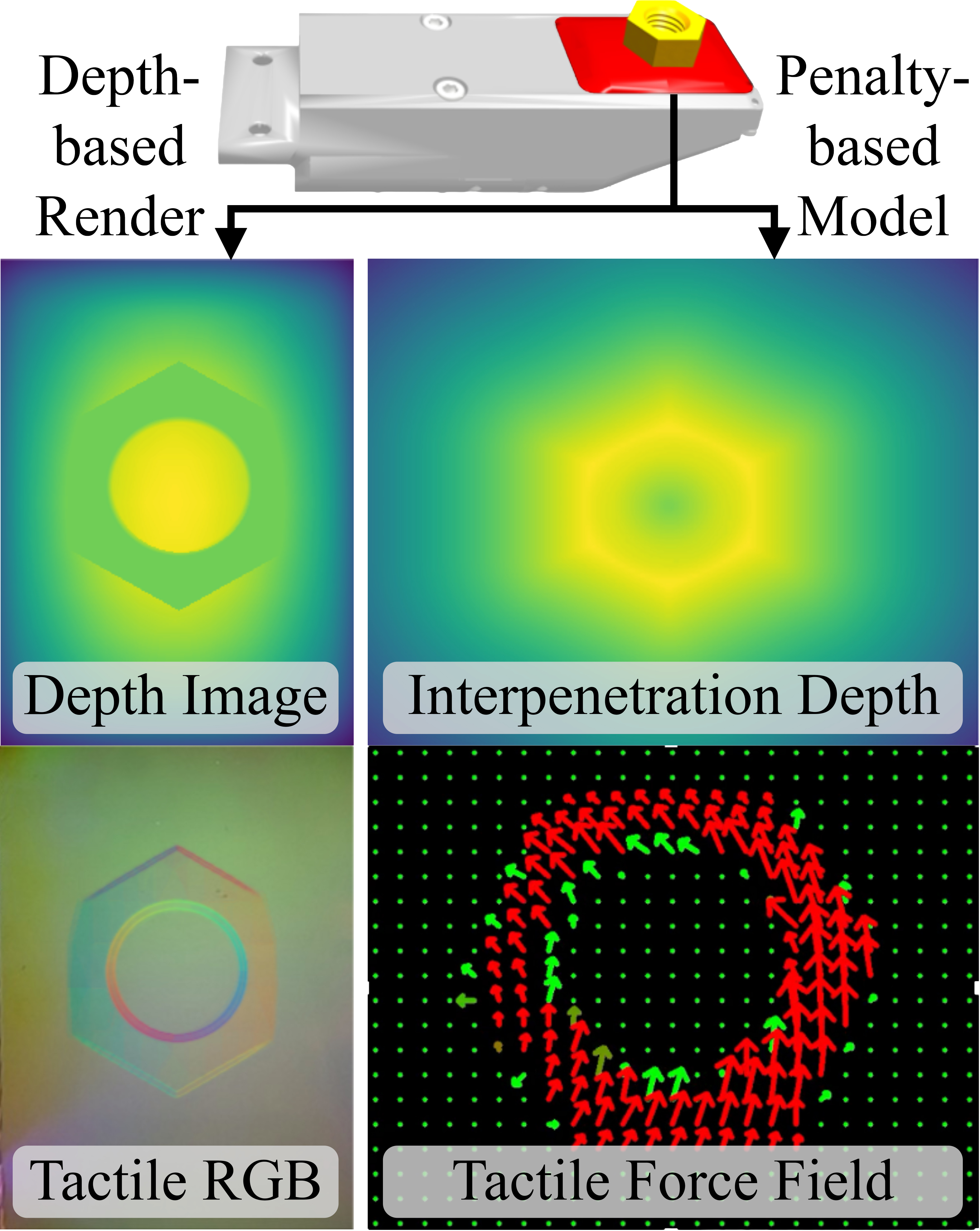}
  \caption{A fast visuo-tactile simulation module that generates tactile images and force fields, based on a soft-contact model between rigid bodies and the sensor.}
  \label{fig:tactile-sensor}
\end{wrapfigure}

Tactile sensor simulation consists of modeling (1) sensor–object contact interactions and (2) transduction of these interactions into measurable signals. These processes are implemented in Isaac Lab for vision-based tactile sensors, where contact interactions are transduced into camera images, as demonstrated in~\cref{fig:tactile-sensor} \citep{akinola:tro2025}. The implementation consists of two GPU-parallelized phases. In the first phase, soft contact dynamics are approximated with a compliant contact model. Deformation is not explicitly modeled, but resistive forces are captured using stiffness and damping parameters, which regulate softness and velocity decay during sensor-object interaction. In the second phase, the contact interaction is used to derive visuotactile RGB images and contact force fields, where the latter characterizes distributed normal and shear forces across the sensor. Depth images are rendered using tiled cameras and mapped to RGB space \citep{si2022taxim}, while a penalty-based model is applied to compute the force distributions \citep{xu:corl2022}. Together, this implementation enables substantial performance gains relative to existing simulation frameworks by combining GPU parallelization, approximate but efficient contact modeling, and tiled image rendering.

\subsection{Controllers} %

Controllers in Isaac Lab represent a class of robotic tools that generate desired joint-level commands (position, velocity, effort) from a higher-level input. These higher-level inputs typically involve task or joint space desired motion. The role of a controller is to determine the desired joint-level actions required to complete the higher-level command. Controllers in Isaac Lab can be organized into a few key categories: inverse kinematics, force control, and motion planners.  These controllers are typically integrated into the actions of an MDP formulation of the robot learning tasks.

\subsubsection{Inverse Kinematics}
\label{subsubsec:IK}

The \ac{IK} controllers in Isaac Lab convert desired task-space motion into the required joint space motion using two main methods. First is the differential IK controller that utilizes the kinematic Jacobian to compute changes in joint position from a desired change in end-effector pose. The second is an implementation of the Pink library that utilizes a quadratic programming method to determine desired joint velocities from a desired end-effector velocity. 

\label{subsubsec:diffik}
 
The \textbf{Differential IK} controller converts desired changes in task-space pose to changes in joint position by utilizing the notion of differential kinematics defined by the kinematic Jacobian, by using the inverted Jacobian matrix. The Jacobian inverse has a flaw that occurs when reaching kinematic singularities. The Differential IK controller allows for different ways of handling this singularity. Options include the Moore-Penrose pseudo-inverse, adaptive singular value decomposition, the Jacobian transpose approximation, and the damped pseudo-inverse. Configurable functionality for the Differential IK controller includes controlling the position or pose of the end-effector and handling the relative or absolute desired position or pose.

\label{subsubsec:pinkik}

Isaac Lab integrates the \textbf{Pink} library, a Python framework for task-based differential inverse kinematics that uses Pinocchio for forward kinematics and quadratic programming (QP) solvers for optimization. Unlike traditional analytical IK methods, Pink allows for computing motions that steer the robot toward achieving multiple simultaneous tasks through a weighted objective formulation.
Building on this foundation, we have implemented a reactive IK controller that achieves real-time performance.
A key advantage of Pink lies in its extensible task-based architecture, which allows users to easily introduce custom objectives into its underlying QP solver. Tasks are defined by residual functions that describe desired behaviors (e.g., end-effector positioning, collision avoidance, joint limits). While conflicts between competing tasks are resolved through normalized cost weighting, this approach trades off performance and accuracy of each task. Isaac Lab extends Pink's task library with the NullSpacePostureTask, designed for high degree-of-freedom robotic systems. This task regularizes redundant joints that do not contribute to the primary control objective, such as maintaining a preferred arm posture during end-effector positioning. The implementation provides controlled null-space behavior while preserving task-space accuracy, improving motion quality for articulated systems with kinematic redundancy.

\subsubsection{Force Control}

Force control is a useful way to enable robots to interact with environments and utilize proprioception to regulate how forces are imparted on the environment and objects. Isaac Lab provides implementation of two force control methods: the joint impedance controller and the operational space controller. 

The \textbf{Joint Impedance} controller provides an interface to control the joint position of an articulation given a desired impedance. This happens individually at the joint level using a similar PD control to the Ideal PD joint actuator, but it additionally provides interfaces for calculating feedforward efforts for both inertial and gravity compensation. This controller also provides the ability to have fixed impedance, variable stiffness, and variable impedance versions.

The \textbf{Operational Space} controller provides an interface for controlling the impedance of a robot at the operational or task space of the robot. This controller comes with a generalized interface for functionality like hybrid force-motion control, dynamics decoupling, gravity compensation, variable impedance, and null space control for redundant manipulators.

\subsubsection{Motion Planning}
\label{itm:curobo} %

The \textbf{cuRobo}~\citep{sundaralingam2023curobo} integration in Isaac Lab enables fast, GPU-parallelized collision-aware motion planning. At its core is cuRobo’s \texttt{MotionGen}, which combines inverse kinematics, collision checking, and trajectory optimization, with optional graph-based planning for global motion. This provides low-latency planning for dynamic scenes and manipulation tasks such as those in workflows like~\cref{subsubsec:skillgen}. Within Isaac Lab, the planner is initialized from the robot configuration and world description, running on a dedicated CUDA device. Tensor handling is managed internally to ensure consistency between cuRobo computation and Isaac Lab tensors. A brief warmup call primes kernels and caches to reduce first-plan delay.

\subsection{Teleoperation Support}

Teleoperation remains an integral part of robotics. Its uses span direct user control, controller and policy evaluation, and demonstrations for robot learning. Teleoperation in Isaac Lab can be performed via a range of hardware devices and applications in extended reality. 

\subsubsection{Classical Teleoperation Devices}

Isaac Lab provides support for various teleoperation devices that allow users to control a diverse set of robots. A keyboard device interface is provided and enables delta pose control over robotic arms and grippers. Users can map additional keys to custom callback functions to design specialized workflows, such as controlling the base height/velocity of a humanoid. For smoother human control, Isaac Lab also supports spacemouse teleoperation devices. The use of a spacemouse can enable higher quality human demonstrations for data collection for imitation learning tasks, as well as provide simultaneous multi-axis movement. 

\subsubsection{Extended Reality (XR) Device Teleoperation}

Teleoperation of bimanual or humanoid dexterous manipulation tasks can be challenging to nearly infeasible with traditional devices such as keyboards and spacemice. To enable these tasks, Isaac Lab includes support for extended reality (XR) devices which allow users to perform dexterous control of bimanual/humanoid robots using an Apple Vision Pro (AVP) headset. The user's hand joints detected by the AVP are mapped to the end-effector pose, which is then used in inverse kinematics (IK) to compute the corresponding joint angles and arm positions. To provide a natural and intuitive user experience, the IK formulation incorporates two tasks: (1) a waist task, which infers waist joint positions from hand motion, and (2) a null-space task to maintain the arm in nominal (neutral) positions as described in \cref{subsubsec:pinkik}.

In contrast to prior XR teleoperation systems that directly stream stereoscopic video from the robot’s perspective to the operator’s display \citep{zhang2025unleashing,cheng2024tv}, Isaac Lab uses \href{https://developer.nvidia.com/cloudxr-sdk}{NVIDIA CloudXR} framework to enhance user comfort through low-latency streaming, supported by GeForce Now technology.  CloudXR incorporates re-projection algorithms to mitigate the perceptual effects of residual streaming delays, thereby providing stable and visually comfortable image qualities. An additional benefit is the platform's support for augmented reality (AR) overlays, which allows operators to maintain contextual awareness of their environment during teleoperation.

\subsection{Simulation Scene Generation}

\begin{figure}[H]
    \centering
    \includegraphics[width=\linewidth]{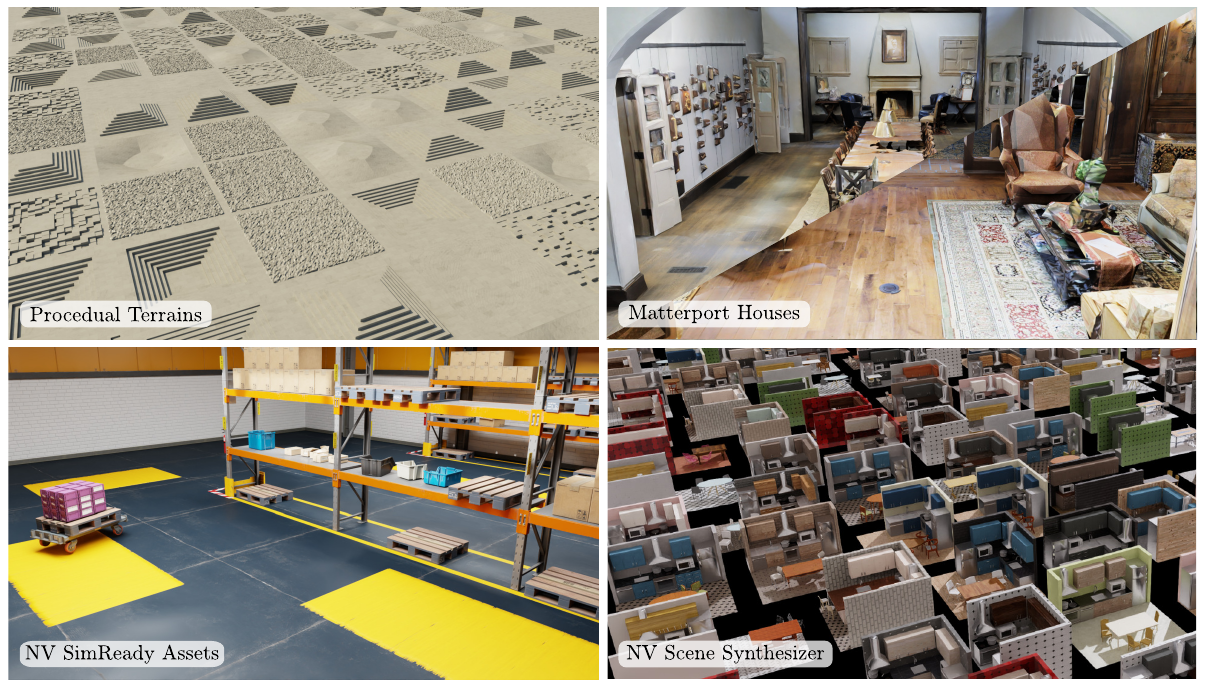}
    \caption{Different procedurally generated scenes in Isaac Lab. Large randomized terrains can be generated using Trimesh library. At the same time, USD terrains from Matterport~\citep{chang2017matterport3d}, NVIDIA SimReady Assets, or the Scene Synthesizer~\citep{Eppner2024} can be imported into Isaac Lab.}
    \label{fig:terrain}
\end{figure}

\subsubsection{Terrain Generation}

Isaac Lab supports multiple approaches for creating simulation environments: (i) \emph{procedural generation} through scripts, (ii) importing scanned meshes~\citep{straub2019replica}, (iii) interactive editing via the Isaac Sim game-like GUI, and (iv) hybrid combinations of them.  
The script-based interface provides fine-grained control over environment difficulty and enables systematic terrain randomization for benchmarking robustness. This interface leverages Trimesh~\citep{trimesh} to generate primitive shapes and arbitrary meshes, which can be flexibly combined into complex structures (e.g., pyramids or composite terrains). It also supports heightfields, which are automatically converted into meshes to represent rough terrain surfaces.  
To ensure efficient simulation, all generated structures are optimized to minimize mesh complexity while preserving geometric fidelity, thereby improving performance in the underlying PhysX engine. \cref{fig:terrain} showcases various types of terrain assets available in Isaac Lab.  

\subsubsection{Integration with External Asset Generation Frameworks}

Scene graph generation frameworks such as Scene Synthesizer \citep{Eppner2024} and Infinigen \citep{infinigen2024indoors} can be used to programmatically design cluttered 3D environments (such as homes and kitchens) and exported to USD for use in Isaac Lab. These frameworks have a library of predefined objects and seemlessly combine them with artist-generated datasets like Objaverse \citep{deitke2023objaverse} to generate USD scenes. Training loco-manipulation policies on such procedural scenes in Isaac Lab results in better robustness and generalization. For instance, \citep{sleiman2024guided} showed sim-to-real transfer of a door opening policy by a quadrupedal mobile manipulator, while \citep{luo2025emergent} demonstrated generalizable pick-and-place skills for a humanoid robot trained in a large number of procedural kitchen environments.

\subsubsection{Multi-Asset Spawning and Randomization}

After defining a terrain, Isaac Lab allows assets to be instantiated in the scene and cloned across parallel environments.  
Within each environment, multiple assets can be spawned simultaneously, and both geometric and visual properties can be randomized.  
Isaac Lab further supports swapping assets across environments, provided they belong to the same asset class.  
For example, in a manipulation task, the target object may vary between a cup, a glass, or cutlery, enabling the policy to generalize across different shapes while pursuing the same objective.  
For vision-based policies, Isaac Lab integrates domain randomization techniques to improve sim-to-real transfer.  
Here, visual properties such as textures, materials, and colors can be procedurally varied across assets and environments.  
This combination of geometric variation and visual randomization promotes robustness by exposing policies to a broader distribution of task-relevant scenarios.

\subsection{Task Framework and Environments}

Isaac Lab supports two primary paradigms for constructing and executing robot learning workflows: the manager-based workflow and the direct workflow. These workflows differ in their level of abstraction, modularity, and intended use cases. This provides flexibility for users ranging from those building minimal, high-performance environments to those developing complex, structured robotic systems. 

The manager-based workflow provides a more structured and modular design, encapsulating robot components into reusable subsystems (e.g., actions, observations, rewards, commands). In contrast, the direct workflow offers an interface that allows users to interact directly with the simulation, physics, and learning components without enforcing a specific organizational structure. It is particularly well-suited for performance-sensitive training pipelines, where minimal overhead and tight integration with GPU resources are critical. \cref{alg:mdp_env_step} outlines the \texttt{step} function logic for both workflows, demonstrating the sequence of computations to step an environment.

\subsubsection{Manager-Based Workflow}
\label{subsec:manager-based-worflow}

The manager-based workflow in Isaac Lab is a structured way to build environments by decomposing the MDP into reusable managers, such as for observations, actions, rewards, terminations, commands, curricula, events, and recording. Each unit has a single responsibility and a clear interface. This design targets ML/RL practitioners: you interact with MDP concepts directly while the framework handles simulation plumbing (scene updates, decimation), vectorization, per-env resets, and Gym-compatible spaces derived from the active terms. As a result, environments remain readable and extensible, and switching on/off terms or swapping configurations becomes a configuration change rather than a refactor.

Beyond enabling flexible MDP definitions, Isaac Lab ships a catalog of tested, independent term functions with explicit inputs/outputs. Practitioners can add or ablate terms without touching unrelated code paths, accelerating idea testing, ablation studies, and continuous development. The manager API formalizes where each computation lives and exposes logging hooks to attribute signals at term-level granularity (e.g., reward contributions, termination causes), improving training analysis and reproducibility.
While a manager layer can introduce modest overhead compared with a hand-tuned direct workflow, most of that overhead is CPU orchestration and kernel-launch latency rather than physics itself — precisely the kind of cost that CUDA Graphs are designed to reduce by bundling many GPU operations into a single launch. We foresee manager term graphed executions as a plausible roadmap to bootstrap the MDP calculation runtime, thus narrowing any remaining gap.

\begin{algorithm}[t]
    \small
    \caption{Environment Step Function}
    \label{alg:mdp_env_step}
    \begin{algorithmic}[1]
      \State \textbf{input:} action
      \State \textbf{output:} observations, rewards, terminations, timeouts, extras
      \Procedure{Step}{action}
        
        \BlockComment{// pre-physics step}
        \State Process actions (\eg clipping, affine transformation)

        \BlockComment{// physics step}
        \For{each substep in environment decimation}
          \State Apply processed actions to simulation buffers (via actuator models)
          \State Apply pre-simulation events (if any)
          \State Advance simulation (without rendering)
          \If{render interval reached \textbf{and} \texttt{is\_rendering}}
              \State Render simulation
          \EndIf
          \State Update scene buffers
        \EndFor

        \BlockComment{// post-physics step}
        \State Update episode counters
        \State Compute terminations and timeouts
        \State Compute rewards

        \BlockComment{// environment reset}
        \State \texttt{reset\_env\_ids} $\gets$ environments to reset
        \If{\texttt{reset\_env\_ids} not empty}
            \State Update curriculum
            \State Apply reset events to reset the environments
            \State Reset episode counter and internal buffers (\eg state history)
            \State Update simulator kinematic state
            \If{sensors present \textbf{and} rerender configured}
                \State Render simulation
            \EndIf
        \EndIf

        \BlockComment{// command and observation update}
        \State Update commands
        \State Apply interval events (if any)
        \State Compute observations
        \State \Return observations, rewards, terminations, timeouts, extras
      \EndProcedure
    \end{algorithmic}
\end{algorithm}

\subsubsection{Direct Workflow}
\label{subsec:direct-worflow}

The direct workflow in Isaac Lab is designed to expose fine-grained control over simulation and learning pipelines with minimal abstraction overhead. It allows users to instantiate simulation environments, robots, and tasks directly through procedural APIs, providing immediate access to low-level data structures such as joint states, contact forces, and sensor outputs. 

This workflow is particularly aligned with the end-to-end training paradigm introduced in Isaac Gym~\citep{makoviychuk2021isaac}, where simulation and learning are tightly coupled on the GPU. Users can execute physics steps, compute rewards, apply control actions, and collect observations within a single, optimized environment class. The simplicity and performance of the direct workflow make it ideal for benchmarking algorithms and performance-sensitive workflows.

Despite its simplicity, the direct workflow remains extensible. Users can define custom tasks, observations, and reward functions, and integrate third-party learning frameworks or data logging tools as needed. It serves as a strong foundation for users who require maximum performance and flexibility, especially in scenarios where control and learning are prioritized over system modularity.

\subsubsection{Environments}

\begin{figure}[htp]
    \centering
    \includegraphics[width=\linewidth]{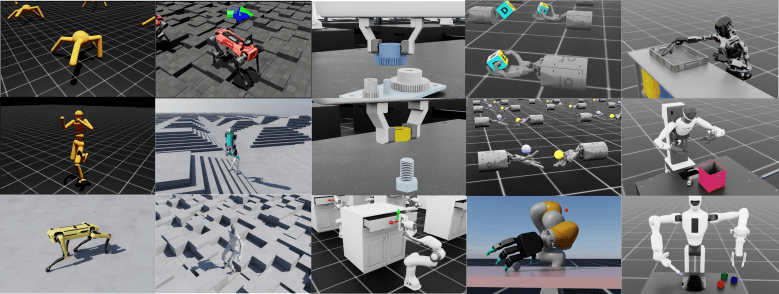}
    \caption{Suite of environments in Isaac Lab. They illustrate a variety of simulation capabilities, including contact-rich interactions, multiple sensor modalities for observations, and support for diverse robotic morphologies, such as humanoids, quadrupeds, fixed-arm robots, and dexterous hands.}
    \label{fig:lab_envs}
\end{figure}

Isaac Lab provides robot learning environments across four families, namely classic, locomotion, manipulation, and navigation, with many tasks offered in both direct and manager-based variants. \cref{fig:lab_envs} showcases some examples of environments available in Isaac Lab.

Across categories, the environments were chosen for their sim-to-real relevance, their advanced simulation features (e.g., SDF, contact sensing, tiled rendering), and their clear skill structure spanning dexterity, locomotion, and hierarchical control. Re-implemented benchmarks share unified APIs and data layouts to minimize glue code and maximize comparability and accessibility for the community to recreate, study, or expand.

\textbf{Classic.} “Hello World” tasks (Ant, Humanoid, Cartpole) are quick-start baselines for bring-up, regression, and throughput testing. Their simplicity allows for rapid prototyping before integration into more complex environments.

\textbf{Locomotion.} Quadruped velocity-tracking and terrain curricula follow massively parallel training recipes and evaluation protocols inspired by prior work ~\citep{rudin2022learning}, which we expanded to 11 different robot morphologies, including A1, G1, H1, Go1/2, Anymal-B/C/D, Cassie, Digit, and Spot. Notably, this implementation has shown robust sim to real for Anymal and Spot.

\textbf{Manipulation.} Dexterous hand–arm tasks (e.g., reorientation, grasping) ~\citep{DexPBT_RSS23, DextrAH-RGB24} are re-created, alongside contact-rich assembly benchmarks that leverage SDF-based contact modeling and force-aware strategies ~\citep{narang2022factory, noseworthy2025forge, tang2024automate}. The suite covers both single-object dexterity and multi-part assembly, with options for tactile/force sensing.

\textbf{Navigation.} Goal-conditioned navigation tasks target hierarchical RL setups where we show how to compose a high-level command policy with a low-level locomotion policy into one environment.

\textbf{Multi-Agent Environments}
Isaac Lab includes and supports the creation of custom environments for solving general physical-based \ac{MARL} tasks, in which multiple learning agents coexist and interact within a shared environment. Depending on the nature of the problem and the interests of the agents, tasks can be defined with different settings, such as competition (agents play against each other), cooperation (agents work together to achieve a common goal), or a combination of the two. 
The exposed API is available for direct workflows and is based on PettingZoo~\citep{terry2021pettingzoo} Parallel API.

\section{Simulation Performance}

\subsection{Environment Throughput}
\label{sec:perf-benchmark}

We evaluate throughput for complex learning tasks with different sensor setups. All benchmarks are executed in headless mode on three GPU platforms: \textit{L40} (48 GB), \textit{RTX Pro 6000} (96 GB), and \textit{GeForce 5090} (32 GB). These cover a range of numbers and generations of RTX cores and VRAM sizes. In addition, we study the scaling behavior under distributed training across multiple GPUs on the \textit{RTX Pro 6000}. As a performance metric, we report frames per second (FPS), defined for environment learning throughput as:  

\begin{equation}
    FPS = \frac{\# \text{ of environment steps}}{\text{simulation time} + \text{learning time}}
\end{equation}

Apart from the GPU, environment throughput can also be dependent on single-core CPU performance due to bottlenecks in some parts of PhysX simulation and the main training loop. The \textit{L40} server uses two AMD EPYC 7763 64-Core Processors, the \textit{RTX Pro 6000} server uses two AMD EPYC 9554 64-Core Processors, and finally the \textit{GeForce 5090} workstation uses a single 8-core AMD 9800X3D processor with single-core performance more than double that of the \textit{L40} server. 
Configurations for all evaluations are provided in the IsaacLab repository.

\subsubsection{State-based Environments}

\begin{figure}[ht!]
    \centering
    \includegraphics[width=\linewidth]{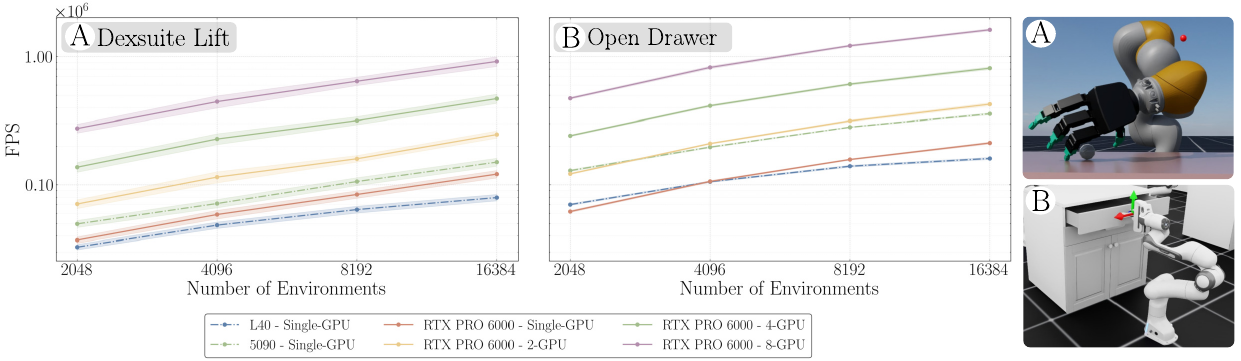}
    \caption{Log-scale throughput comparison for state-based manipulation tasks on three GPU platforms, including distributed training with two, four, and eight GPUs. These are shown for the Dexsuite task to grasp and lift an object and the Franka arm opening a cabinet drawer task.}
    \label{fig:benchmark-state-based}
\end{figure}

For state-based training relying on proprioceptive information only, we benchmark two manipulation tasks - the DextrAH \citep{DextrAH-G24} task to grasp and lift an object with no perception, and the Franka robotic arm opening a cabinet drawer task.
As shown in~\cref{fig:benchmark-state-based}, the systems with newer Blackwell GPUs have improved performance across both environments compared to the previous generation \textit{L40}. Distributed training provides further gains, scaling almost perfectly linearly as the number of GPUs increases. With eight GPUs and 16384 environments, the DextrAH teacher task reaches over 900,000 frames per second in training, while the Franka cabinet task reaches over 1.6 million frames per second. \\

It is notable that the workstation \textit{GeForce 5090} system comes close in performance to the two GPU \textit{RTX Pro 6000} server in the Franka cabinet drawer task, but is only about 25\% faster in the DextrAH task. These differences are primarily due to the combination of CPU bottlenecks and extremely fast single-core CPU performance in the 9800X3D CPU compared to the throughput-oriented server CPUs. Eliminating these bottlenecks is an important goal for future releases of Isaac Lab and the Newton physics engine.

\subsubsection{Perceptive Environments}

In perceptive tasks, the simulation speed depends on the combination of PhysX and rendering speeds, where the latter typically dominates computational demands. 
We benchmark three perceptive tasks: two rough terrain locomotion tasks with the Unitree G1 humanoids and Agility Robotics Digit humanoid, and the end-to-end perception-based version of DextrAH for dexterous manipulation~\citep{Singh2025_E2E}. The two locomotion tasks use a height-scanner of size $1.6m \times 1.2m$ implemented using the Warp-based \emph{RayCasterCamera}.
For the DextrAH environment, we compare two rendering approaches: the \emph{Tiled-Camera} and the \emph{RayCasterCamera}. 
The Digit locomotion environment also showcases the simulation of closed-loop kinematic chains which requires higher solver iteration count for stabler simulation. Benchmarks are run on the same GPU platforms as for the state-based learning tasks.

\begin{figure}[h!]
  \centering
    \includegraphics[width=\linewidth]{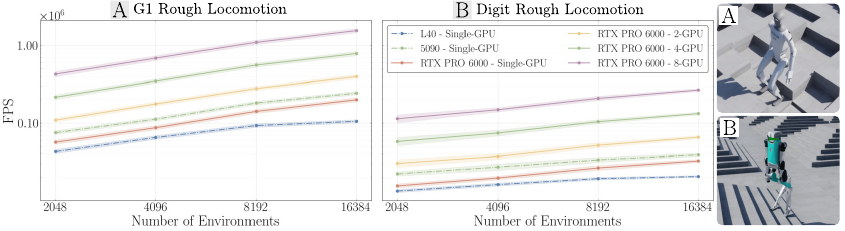}

  \caption{Log-scale throughput comparison for perceptive locomotion tasks across GPU models and distributed training setups. The task for both Unitree G1 and Agility Digit use a RayCaster-based height scanner ($1.6\,\text{m} \times 1.2\,\text{m}$, resolution $0.1\,\text{m}$).
  The Digit robot is simulated with closed-loop kinematic chains, that requires a higher solver iteration count for stable simulation.
  }
  \label{fig:perceptive_raycaster_benchmark}
\end{figure}

\begin{figure}[h!]
  \centering
    \includegraphics[width=\linewidth]{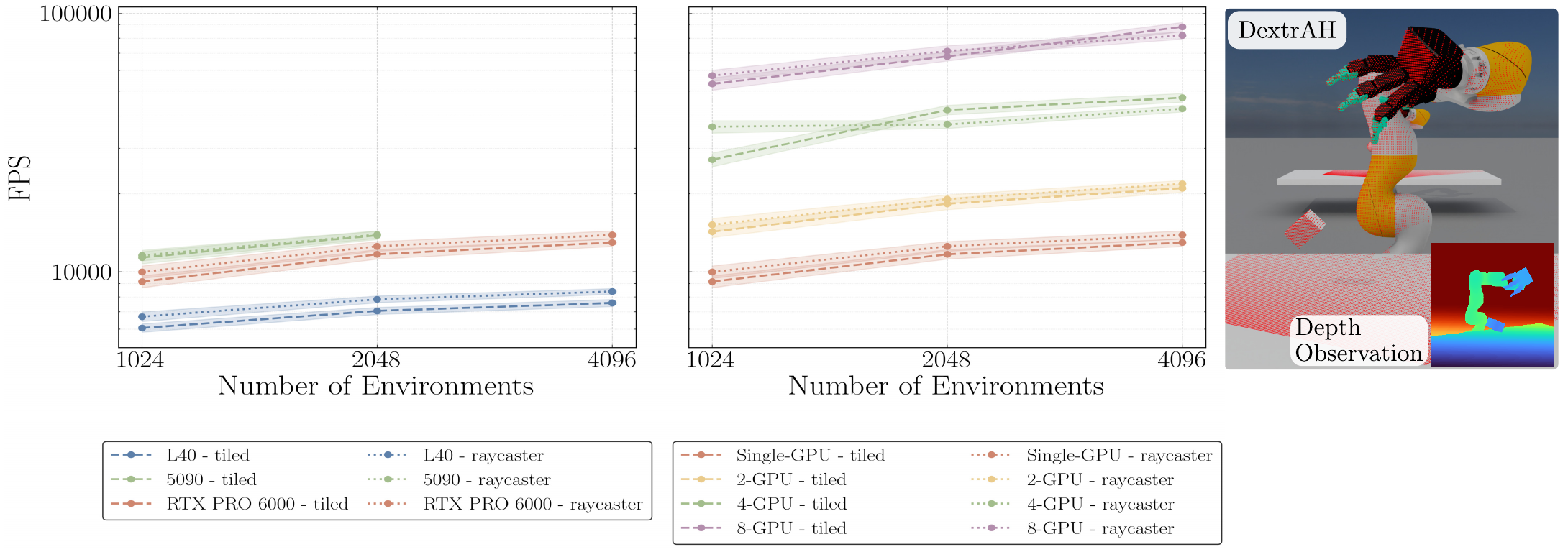}

  \caption{Log-scale throughput comparison for perceptive learning task in dexterous manipulation (DextrAH). Results are shown for the Tiled and Raycaster-based camera at 64x64 resolution across GPU models (left) and distributed training setups (right). All multi-GPU training is tested on \textit{RTX Pro 6000} systems.}
  \label{fig:perceptive_tiled_benchmark}
\end{figure}

As shown in~\cref{fig:perceptive_raycaster_benchmark} and~\cref{fig:perceptive_tiled_benchmark}, throughput improves substantially with newer GPUs (\textit{RTX Pro 6000} and \textit{GeForce 5090}) and with distributed training, mirroring the trends observed in state-based environments. However, throughput remains lower overall compared to state-based tasks due to the additional rendering cost. 

The \textit{GeForce 5090} system, with its high single-core performance CPU, achieves higher throughput at smaller environment counts, while the \textit{RTX Pro 6000} system narrows the gap as the number of environments increases. Owing to its larger VRAM capacity, the RTX Pro 6000 also supports a greater degree of parallelization, enabling more environments to be simulated concurrently.
When comparing sensor implementations, the RayCasterCamera demonstrates superior performance on a single GPU. In contrast, with multiple GPUs and large environment counts, the Tiled-Camera achieves better parallelization, eventually surpassing the RayCasterCamera.

\subsubsection{Workflow Comparison}

\begin{figure}[h!]
  \centering
    \includegraphics[width=\linewidth]{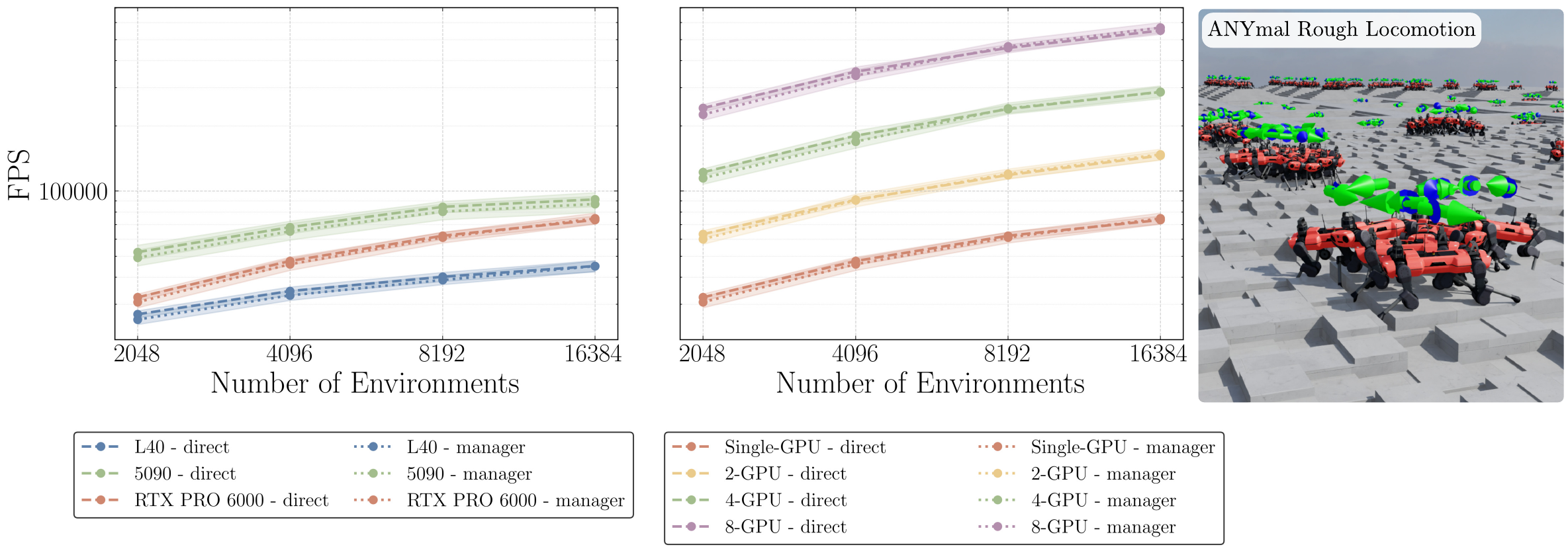}

  \caption{Log-scale throughput comparison of the rough terrain locomotion task for ANYmal-D robot implemented using the manager-based and direct workflows. Right: The plot over single GPU setups. Left: The plot over multi-GPU setups. The task uses a height-scanner for observations and an actuator network for the robot.}
  \label{fig:workflow_benchmark}
\end{figure}

We further evaluate the performance of the manager-based and direct workflows introduced in \cref{subsec:manager-based-worflow} and~\cref{subsec:direct-worflow}. For the comparison, we use the ANYmal rough terrain locomotion task, detailed configuration accessible in the IsaacLab repo under the task names \textit{Isaac-Velocity-Rough-Anymal-C-Direct-v0} and \textit{Isaac-Velocity-Rough-Anymal-C-v0} respectively. As shown in~\cref{fig:workflow_benchmark}, the direct workflow generally achieves slightly higher throughput, on average 3.53\% on a single \textit{RTX PRO 6000}. However, the performance gap is small and becomes negligible for larger environment counts and in tasks where perception dominates the computational cost. This indicates that the manager-based workflow offers comparable efficiency while providing additional structure and flexibility for complex training setups.

\subsection{Sensor Throughput Numbers}

While environment throughput benchmarks combine both simulation and learning time, here we isolate sensor performance to analyze simulation costs in detail. Specifically, we measure the update times of different sensor implementations. Consequently, FPS for this section can be simplified to:

\begin{equation}
    FPS = \frac{\# \text{ of rendering steps}}{\text{simulation time}}
\end{equation}

We first benchmark various camera implementations and then investigate the scaling behavior of the Warp-based RayCaster sensor (\cref{subsubsec:sensor-warp-based}) with respect to mesh complexity, environment count, and other parameters.

\subsubsection{Camera}

\begin{figure}[h!]
  \centering
  \includegraphics[width=\linewidth]{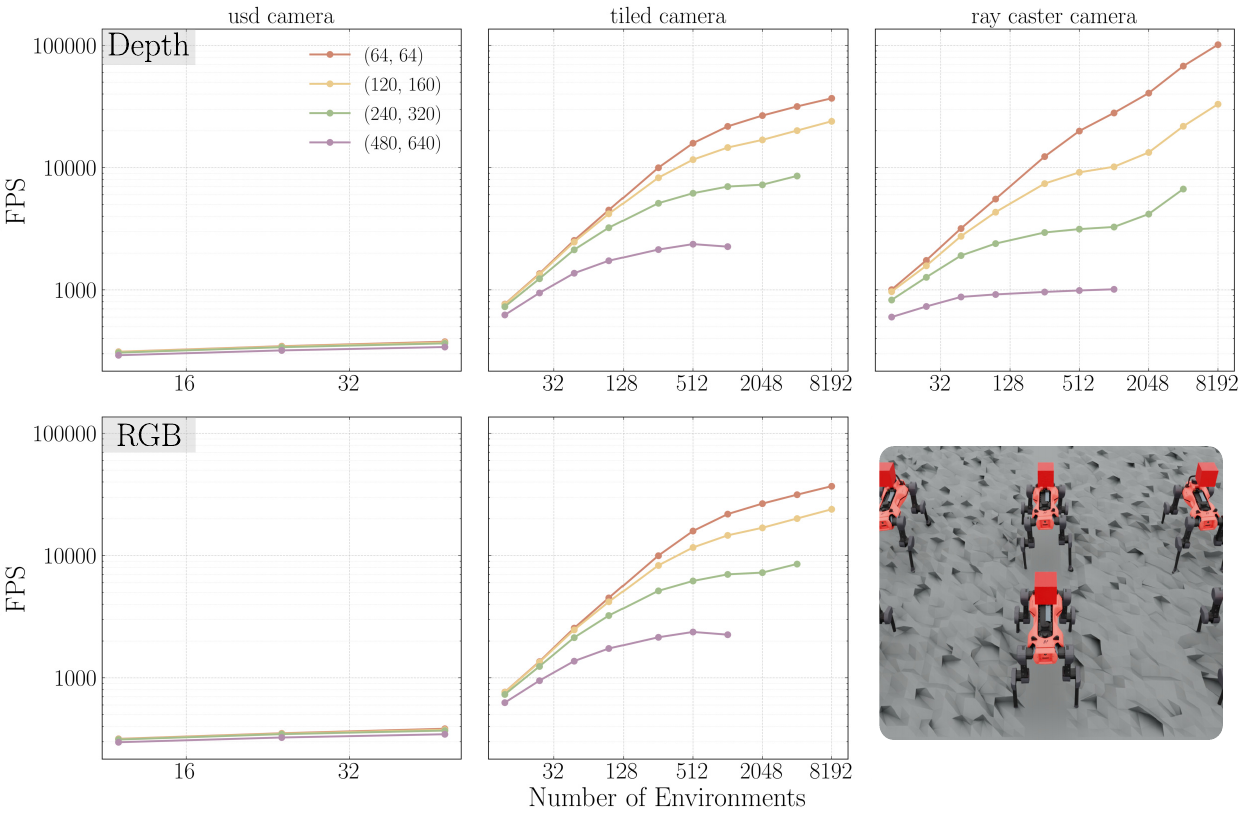}
  \caption{Throughput comparison of different camera sensor implementations in Isaac Lab-- naive USD camera rendering, tiled rendering, and Warp raycast based rendering. The plots shows throughput for different imgage sizes on a single GPU. The curves terminate when the GPU runs out of memory.}
  \label{fig:cam_benchmark}
\end{figure}

Isaac Lab provides multiple rendering-based and raycasting-based camera implementations, introduced in~\cref{subsubsec:sensor-rendering} and~\cref{subsubsec:sensor-warp-based}. To benchmark their performance and highlight the trade-offs between photorealism, efficiency, and scalability, we evaluate throughput for different numbers of environments and image resolutions. To mimic a typical robot learning setup, each sensor is mounted on an ANYmal-D robot, and we scale scene complexity with the number of parallel environments by adding an additional mesh per environment (see~\cref{fig:cam_benchmark}). All experiments are executed on a single \textit{RTX Pro 6000}.  

The USD-Camera provides the highest visual fidelity but incurs substantial memory demand. As seen in~\cref{fig:cam_benchmark}, Out-of-Memory errors occur when simulating more than 48 cameras in parallel, for both RGB and depth modalities. In contrast, the Tiled-Camera and RayCasterCamera have significantly lower memory footprints, enabling scaling to several thousand environments. For the Tiled-Camera, throughput is not affected when rendering RGB images compared to depth images.

When comparing the Tiled-Camera and RayCasterCamera, the latter exhibits a slightly smaller memory footprint. Throughput trends differ by resolution: at lower resolutions, the RayCasterCamera is more efficient, whereas at higher resolutions, the Tiled-Camera achieves superior performance. Ongoing development efforts aim to reduce the memory footprint of the TiledCamera and to further improve the performance of both the TiledCamera and RayCasterCamera in subsequent releases.

\subsubsection{Warp Ray-Caster Sensor}

\begin{figure}[h!]
    \centering
    \begin{subfigure}[c]{0.30\linewidth}
        \centering
        \includegraphics[width=\linewidth]{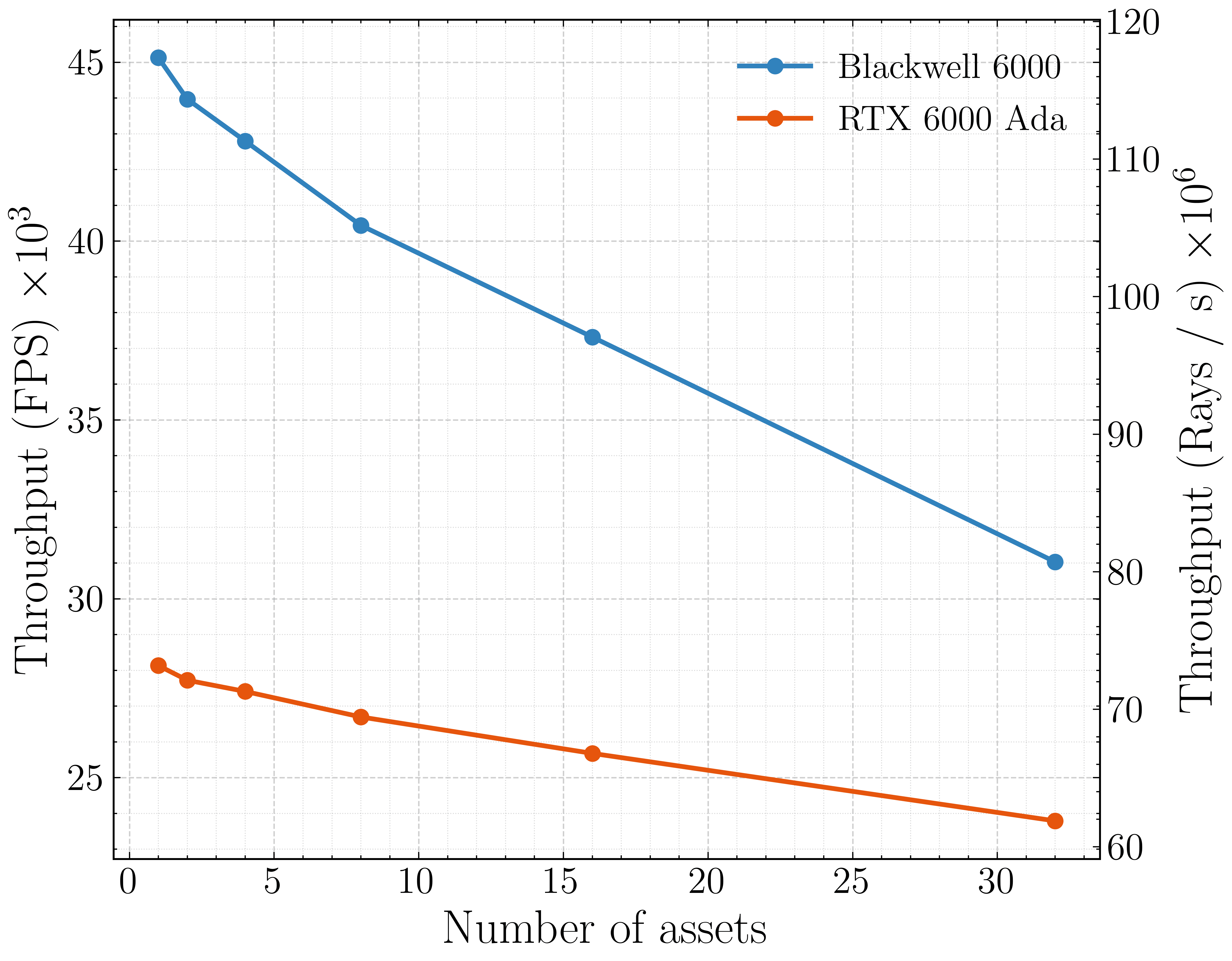}
        \caption{Number of Assets}
        \label{fig:raycaster_assets}
    \end{subfigure}
    \hfill
    \begin{subfigure}[c]{0.30\linewidth}
        \centering
        \includegraphics[width=\linewidth]{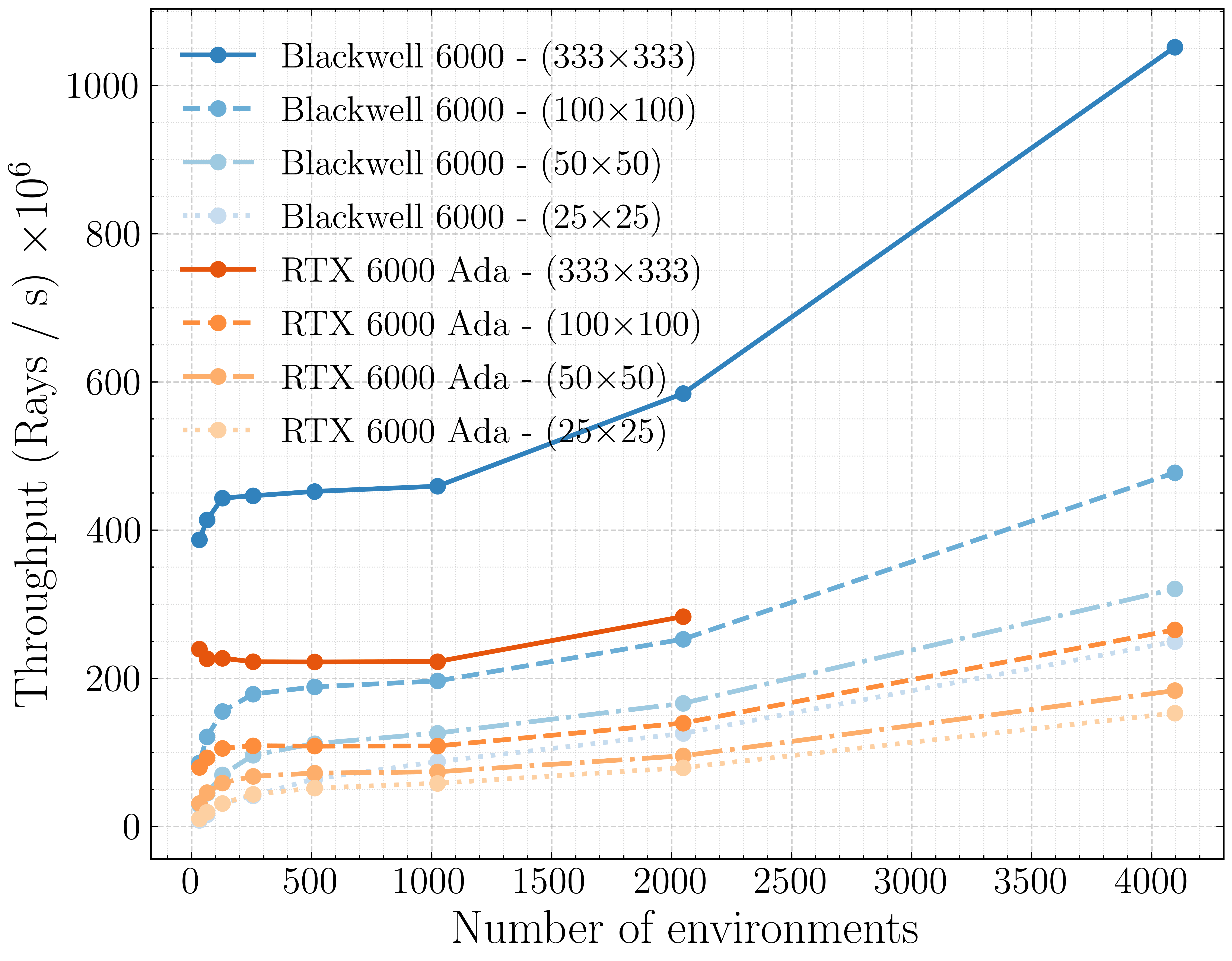}
        \caption{Image Resolution}
        \label{fig:raycaster_resolution}
    \end{subfigure}
    \hfill
    \begin{subfigure}[c]{0.30\linewidth}
        \centering
        \includegraphics[width=\linewidth]{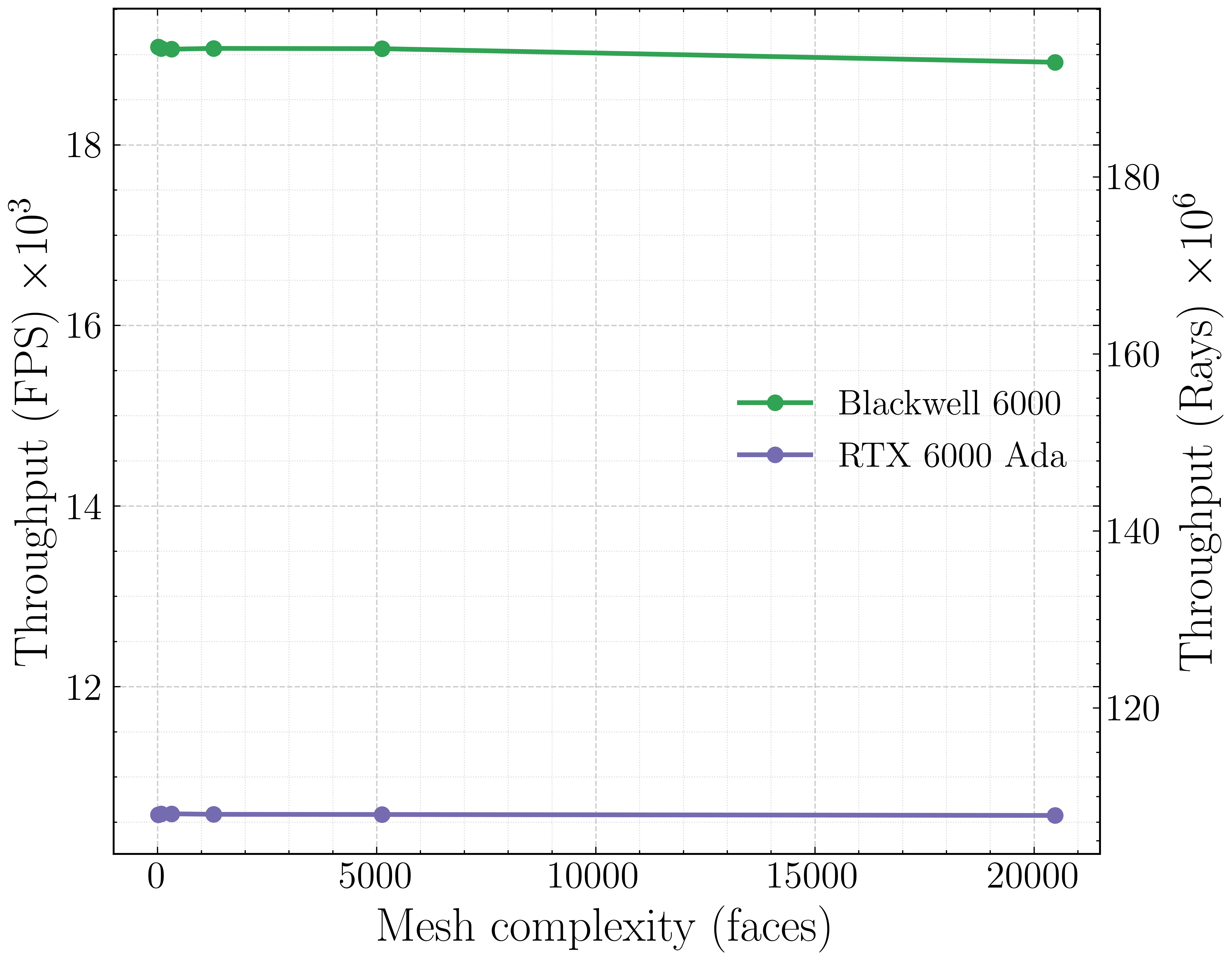}
        \caption{Mesh Complexity (faces)}
        \label{fig:raycaster_mesh}
    \end{subfigure}
     \caption{Benchmarking of the Warp RayCaster sensor. 
Throughput is evaluated under different scaling factors: (a) varying the number of target assets, showing sub-linear performance degradation as assets increase; (b) resolution scaling, where throughput increases more strongly at higher resolutions and larger environment counts; and (c) mesh complexity, indicating minimal impact within the tested range of 20–200k faces.
}
    \label{fig:raycaster_results}
\end{figure}

To evaluate the scalability of the Warp RayCaster sensor, we construct a benchmark environment consisting of a rough terrain mesh and varying numbers of moving spheres. The results, shown in~\cref{fig:raycaster_results}, highlight the effect of different scaling factors on throughput and memory usage.  
In~\cref{fig:raycaster_assets}, we vary the number of tracked assets and observe a sub-linear decrease in throughput as additional meshes are added to the scene.
Resolution scaling, presented in~\cref{fig:raycaster_resolution}, demonstrates that throughput (here defined as number of rays per second) increases with higher image resolutions, with scaling effects becoming more pronounced as the number of environments increases.  Finally,~\cref{fig:raycaster_mesh} shows that mesh complexity has only a marginal impact within the tested range of 20k–200k faces.  

Overall, these benchmarks indicate that raycasting performance is primarily influenced by GPU capability, the number of meshes to raycast against, ray density, and the number of parallel environments. Mesh complexity, by contrast, plays only a minor role. 
These results suggest that tuning ray density provides a more effective lever for balancing accuracy and efficiency than mesh simplification, with the choice of GPU remaining the dominant factor for maximum throughput in large-scale simulations.

\section{Learning Techniques and Workflows}

\subsection{Reinforcement Learning}

Isaac Lab adheres to the Gymnasium API~\citep{towers2024gymnasium}, a widely adopted interface for defining \ac{RL} environments. By conforming to the \texttt{gymnasium.Env} specification, Isaac Lab environments can be used directly with any library that supports Gym-compatible environments.
At the same time, many RL frameworks implement custom specifications for handling batched data from parallelized environments. To address this, Isaac Lab is designed with extensibility in mind to integrate custom solutions with minimal engineering effort. 
Out-of-the-box, Isaac Lab provides built-in support for SKRL~\citep{serrano2022skrl}, {RSL-RL}~\citep{schwarke2025rslrl}, {RL-Games}~\citep{rl-games2022}, {Stable-Baselines3 (SB3)}~\citep{stable-baselines3}, and {Ray}~\citep{moritz2018ray}. These libraries complement different aspects of the RL workflow and highlight Isaac Lab’s interoperability.

Learning effective control policies directly from visual inputs is a central challenge in modern robotics and reinforcement learning. While traditional approaches rely on low-dimensional state representations, many real-world applications require policies capable of interpreting and acting upon high-dimensional sensory observations, such as RGB or depth images. Isaac Lab supports this line of research by providing fast high-fidelity rendering and privileged state access.
It supports visual domain randomization, exposing policies to diverse lighting conditions, textures, colors and backgrounds, and provides examples using pre-trained visual backbones such as Theia~\citep{shang2024theia} and ResNet~\citep{he2016deep}.

In the following, we describe two complementary paradigms for incorporating perception into reinforcement learning: \textit{teacher-student distillation}, where a perception-based student imitates a privileged state-based teacher, and \textit{end-to-end perception-in-the-loop training}, where policies are learned directly from raw sensory inputs.

\subsubsection{Teacher-Student Distillation}

 \begin{figure}[t]
    \centering
    \begin{subfigure}[t]{0.55\textwidth}
        \centering
        \includegraphics[width=\linewidth]{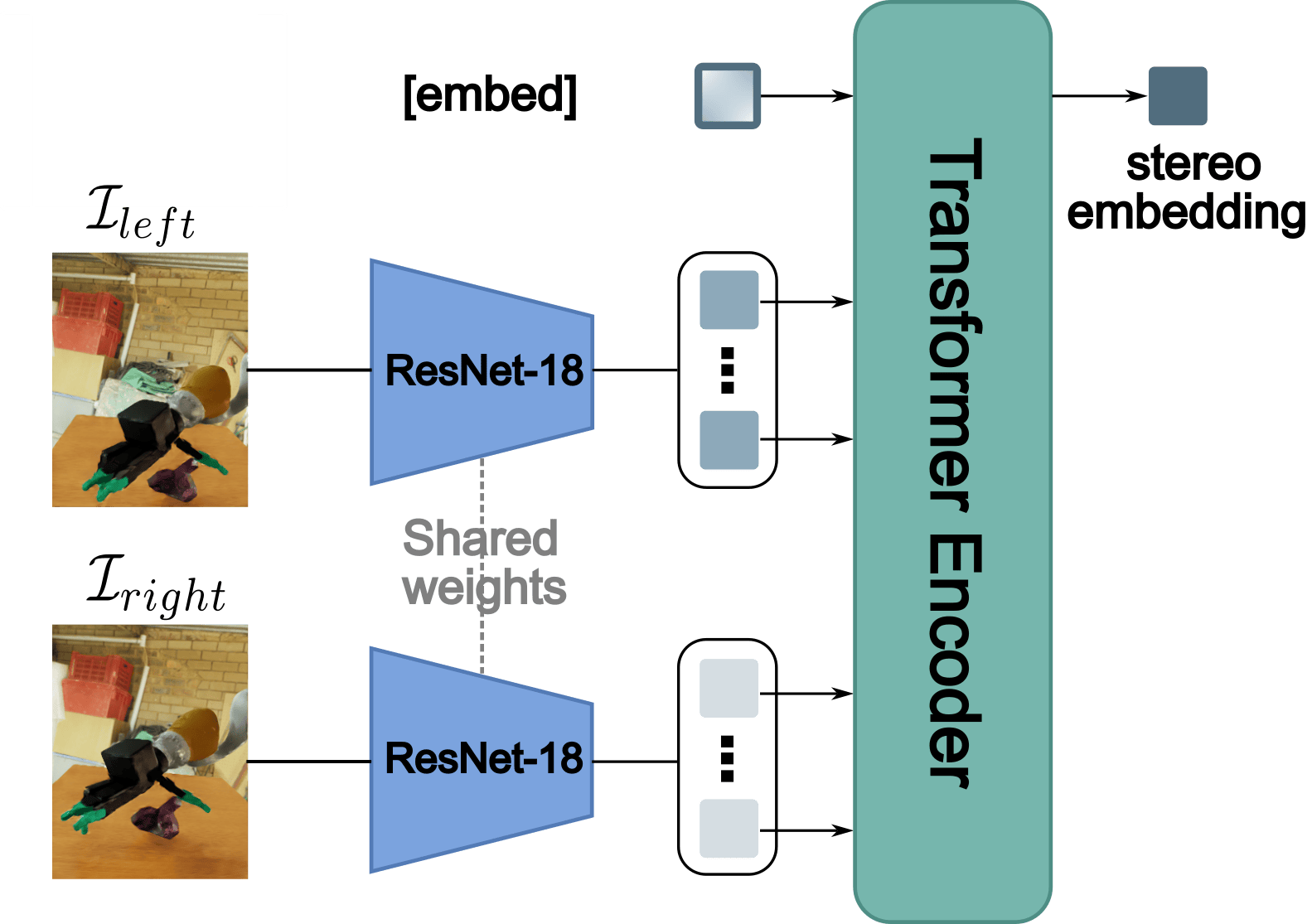}
        \caption{\centering Stereo Encoder}
        \label{fig:stereo_encoder_a}
    \end{subfigure}%
    \begin{subfigure}[t]{0.35\textwidth}
        \centering
        \includegraphics[width=\linewidth]{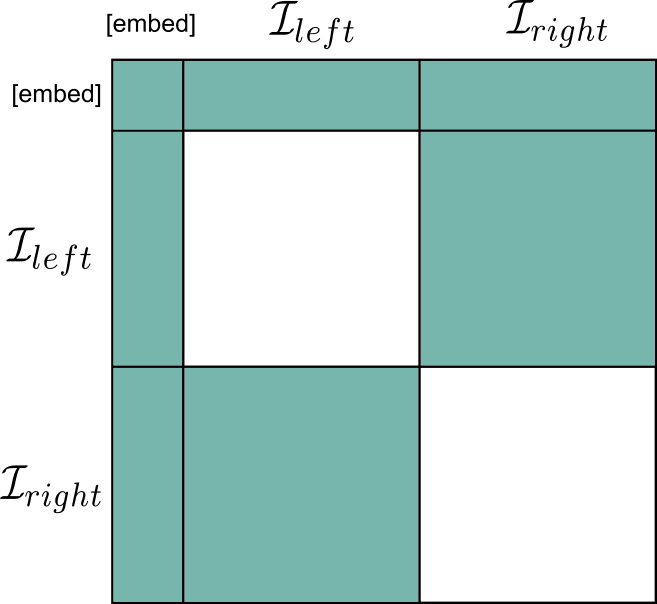}
        \caption{\centering Cross-Attention Mask}
        \label{fig:stereo_encoder_b}
    \end{subfigure}

    \caption{%
        \citet{DextrAH-RGB24} train a student policy with stereo RGB images. \textbf{(a)} The stereo encoder uses a pre-trained ResNet-18 (with the last two layers removed) to encode each image independently into a high-dimensional vector. Each vector is projected and split into 128 tokens. Tokens from both images, along with a learnable \texttt{[embed]} token, are passed into a two-layer transformer that performs cross-attention. The output from the \texttt{[embed]} token is processed through an MLP, producing the final stereo embedding vector.  
        \textbf{(b)} The turquoise regions illustrate cross-attention between the tokens. Each image's tokens attend to the other image's tokens and the shared \texttt{[embed]} token, which attends to all tokens.
    }
    \label{fig:stereo_encoder}
\end{figure}

Training end-to-end pixels-to-action policies directly with \ac{RL} presents unique challenges, as the agent must simultaneously learn perception and control.
A widely adopted solution is the teacher-student distillation approach \citep{chen2020learning, lee2020learning}, in which a teacher policy, trained via RL on state-based (privileged) observations, guides a student policy that receives partial information (such as rendered RGB images) corresponding to the teacher’s states. The student is trained using an online imitation learning method, such as DAGGER \citep{ross2011reduction}, which is particularly effective as it incrementally aggregates data from the student’s own trajectories while leveraging the teacher for corrective actions.

The distillation framework can be applied to students receiving different types of observations. For example, \citet{lee2020learning} trains a student with proprioception-based observations, while \citet{DextrAH-G24} uses depth-based observations. More recently, \citet{DextrAH-RGB24} trains a student policy with RGB images, using high-fidelity rendering in Isaac Lab.
Student architectures tend to be simpler for proprioception- and depth-based observations. In contrast, training an RGB-based student is more challenging, as the network must learn invariance to textures and lighting, infer 3D structure, and identify objects of interest directly from RGB inputs. To facilitate this, it is often helpful to initialize the student with a pre-trained encoder, as shown in \cref{fig:stereo_encoder}.

\subsubsection{End-to-End Perception-in-the-Loop}

While distillation enables training an RGB-driven student policy by imitating a state-based teacher, there are compelling reasons to also train policies end-to-end directly from images. First, end-to-end training can exploit rich spatial and shape cues that low-dimensional state vectors cannot capture. Second, imitation-based distillation introduces an information gap due to input mismatch: the teacher observes privileged state information, while the student sees only partial observations. This can lead to a performance drop, as the student must reconstruct unobserved states from images, a challenge that becomes particularly pronounced under high camera occlusions.

\cite{Singh2025_E2E} shows the first sim-to-real system trained end-to-end for multi-fingered hands using Isaac Lab. Training end-to-end \ac{RL} policies can also give rise to emergent active perception behaviors.~\cite{luo2025emergent} uses Isaac Lab to train a humanoid agent relying solely on egocentric vision and proprioception to perform various loco-manipulation tasks. Similarly,~\cite{yang2025improving} trains an attention-based RL policy for long-range navigation. The learned agents in these works exhibit search and exploratory behaviors, actively moving around to gather visual information relevant to the task, demonstrating active perception strategies that emerge naturally from end-to-end training.

\subsection{Population-Based Training}

\begin{figure}[H]
    \centering
    \includegraphics[width=0.98\linewidth]{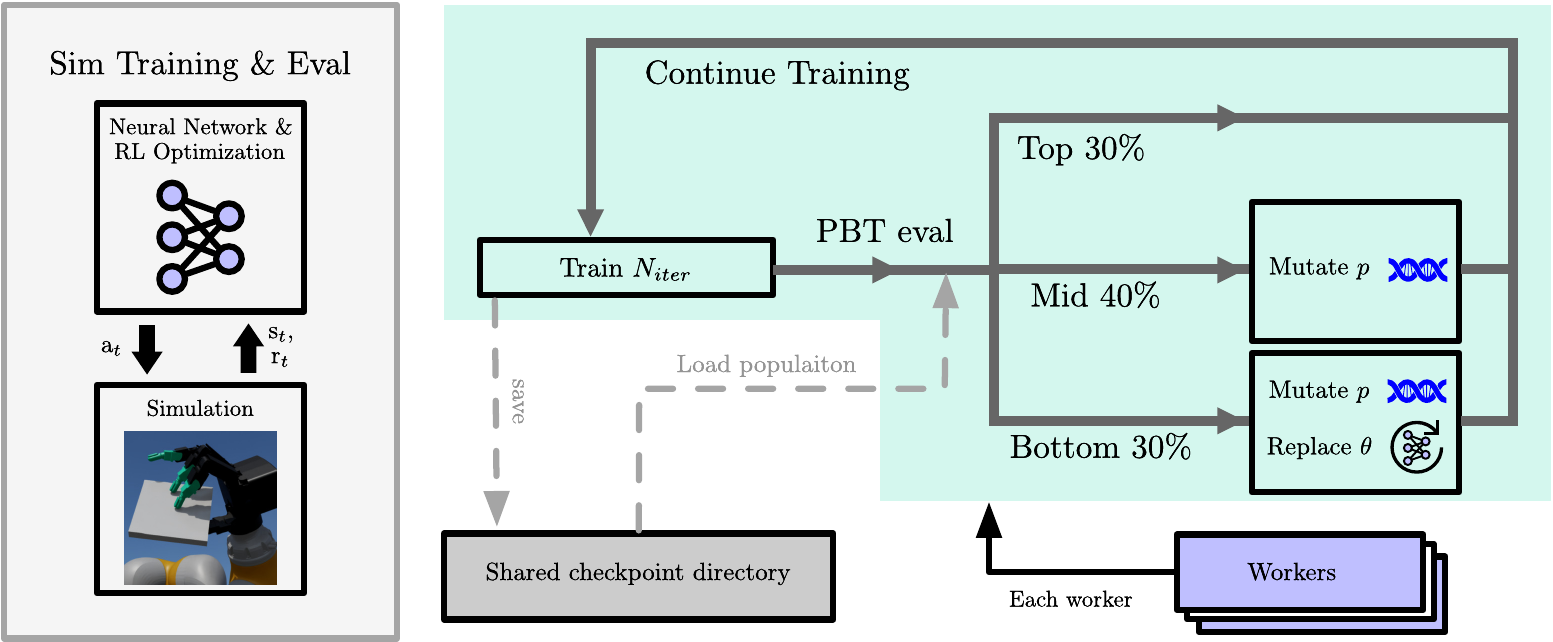}
    \caption{An overview of the \acs{PBT} framework for \ac{RL} in Isaac Lab. Each worker runs an independent \ac{RL} training process with its own hyperparameters. Periodically, top-performing workers share their weights and hyperparameters to replace those of bottom-performing workers, while genetic mutations introduce diversity in the hyperparameter space. This iterative process improves exploration and stability in RL training.}
    \label{fig:pbt-figure}
\end{figure}

While \ac{RL} remains the predominant approach for training policies in simulation, the stochastic nature of the training process and the large hyperparameter space collectively introduce high variance in training results, particularly for systems with high \ac{DoF} systems and tasks with sparse rewards. As a result, some runs may progress rapidly, while others stagnate with little or no improvement.~\citet{DexPBT_RSS23} addresses this problem by using genetic algorithms to mutate the hyperparameter space, promoting diversity and exploration in the learning agent. Built on top of \ac{PBT}~\citep{PBT_2017}, DexPBT assigns each worker an independent RL training process with its own set of hyperparameters. Periodically, the best-performing workers, with their weights and hyperparameters, replace the worst-performing ones, as illustrated in \cref{fig:pbt-figure}. %

Isaac Lab includes an implementation of DexPBT with {RL-Games} for the dexterous manipulation environments in DextrAH~\citep{DextrAH-G24, DextrAH-RGB24}.%
It reproduces the 6D reposing task from the original DexPBT work using 8 workers, each with 1–2 GPUs, and converges in approximately 16 hours on NVIDIA OVX L40 hardware.
The PBT framework in Isaac Lab is configurable and generalizable to other environments, enabling scalable multi-node RL training.

\subsection{Domain Randomization}

\ac{DR} is a critical component of sim-to-real transfer, where the system parameters are randomized in simulation to promote policy generalization and robustness on deployment. For physics, these include friction, armature, gravity, and mass~\citep{physicsrandomisation_jasonpengICRA2018}, while for rendering they include texture, material, and lighting~\citep{DomainRandomisationTobinFRSZA17,sadeghi2016cad2rl}. Such parameters may be difficult to estimate accurately or may vary over time in the real world. By randomizing them over broad distributions during training, policies acquire invariance to these parameters, improving transfer to reality. A key requirement for \ac{DR} is the ability to modify parameters on the fly, providing diverse training scenarios to the learning agent.

\begin{figure}[H]
\centering
\includegraphics[width=0.19\textwidth, angle=270]{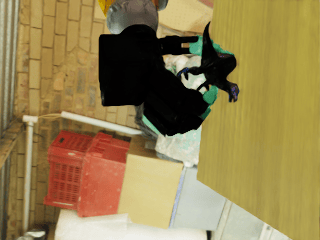}%
\includegraphics[width=0.19\textwidth, angle=270]{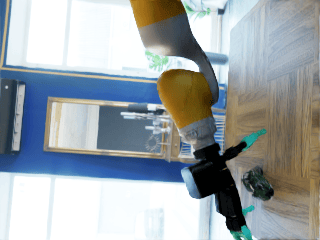}%
\includegraphics[width=0.19\textwidth, angle=270]{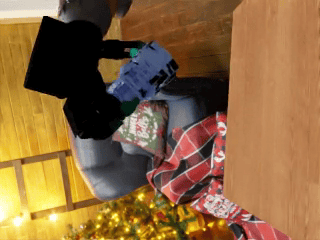}%
\includegraphics[width=0.19\textwidth, angle=270]{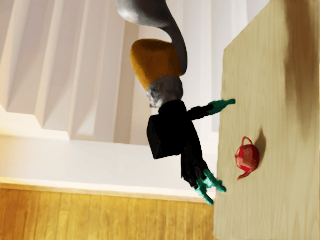}%
\includegraphics[width=0.19\textwidth, angle=270]{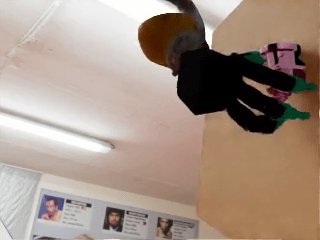}%
\includegraphics[width=0.19\textwidth, angle=270]{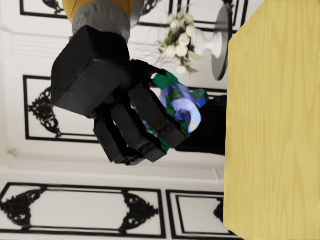}%
\includegraphics[width=0.19\textwidth, angle=270]{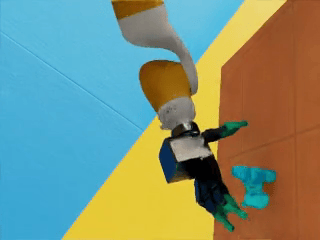}

\includegraphics[width=0.19\textwidth, angle=270]{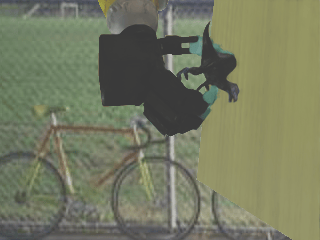}%
\includegraphics[width=0.19\textwidth, angle=270]{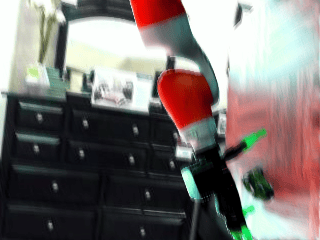}%
\includegraphics[width=0.19\textwidth, angle=270]{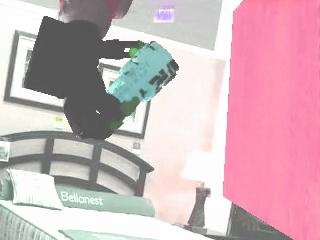}%
\includegraphics[width=0.19\textwidth, angle=270]{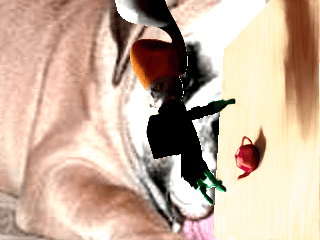}%
\includegraphics[width=0.19\textwidth, angle=270]{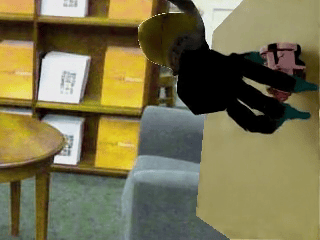}%
\includegraphics[width=0.19\textwidth, angle=270]{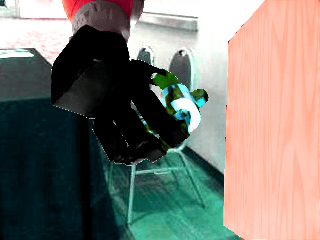}%
\includegraphics[width=0.19\textwidth, angle=270]{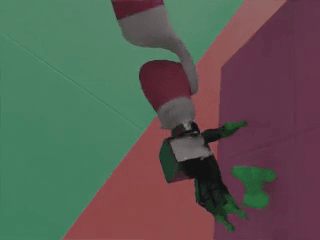}

    \caption{Top: Camera renderings from different environment instances in simulation, adapted from~\cite{DextrAH-RGB24}. Bottom: Examples of data augmentations (such as modifying brightness, contrast and saturation) applied to these renderings before passing them to the learning agent.}
\label{fig:dr_sim_montage}
\end{figure}

Isaac Lab supports randomization of both physics and rendering parameters. Mesh-related attributes, such as scale and collider type, can only be randomized before the simulation begins playing (\cref{alg:omniphysics-workflow}). Most other physics parameters can be randomized at runtime. As discussed in~\cref{sec:sim-tech-physx}, due to current PhysX design limitations, only simulation state is accessible directly on the GPU, while simulation parameters (\eg masses, friction, contact offsets, and joint armature) must be modified through the CPU API.
Rendering parameters such as visual materials, lighting intensities, light source locations, and background textures can also be randomized at runtime, though they currently rely on CPU-based USD APIs. \cref{fig:dr_sim_montage} illustrates RGB renderings in Isaac Lab with randomized textures and lighting, alongside standard computer vision augmentation techniques.

Furthermore, Isaac Lab supports \textbf{Automatic Domain Randomization (ADR)}~\citep{openai2019solvingrubikscuberobot} through a configurable curriculum framework that adaptively adjusts the difficulty of the environment based on the performance of the agent. The \texttt{dexsuite} examples in Isaac Lab provide reference ADR configurations for RL training.
The framework automatically manages difficulty progression, continuously challenging the learning agent without overwhelming it, thereby promoting more robust and generalizable policies.

\subsection{Imitation Learning}

While \ac{RL} has demonstrated clear benefits when scaling with large amounts of simulation data, designing reward functions for many robotic tasks remains a challenge. \ac{IL} offers an alternative by allowing agents to learn directly from expert demonstrations rather than relying solely on trial-and-error exploration. Simulation provides a safer, more scalable, and cost-effective environment for collecting training data for \ac{IL} than the real world. Additionally, recent advances in sim-to-real techniques, including the use of generative models such as \href{https://www.nvidia.com/en-us/ai/cosmos/}{NVIDIA
 Cosmos}, offer promising approaches for training policies entirely on synthetic data and transferring them effectively to real-world environments.

Isaac Lab supports \ac{IL} through integration with the widely used \textbf{RoboMimic} framework by \citet{mandlekar2021robomimic}. Demonstration data can be collected through human teleoperation or synthetically generated using \textbf{Isaac Lab Mimic} (described in \cref{sec:Synthetic-Data-Generation}) . All demonstrations are stored in the standardized HDF5 format based on the RoboMimic schema.
Isaac Lab includes a suite of reference training and evaluation scripts that work out of the box with RoboMimic, offering a streamlined starting point for experimentation.
It also provides a dedicated conversion utility allowing users to transform existing HDF5 datasets into the \textbf{LeRobot} format~\citep{cadene2024lerobot}, which utilizes columnar Parquet for time-series data and MP4 encoding for efficient visual data handling, accelerating integration with the models hosted on Hugging Face Hub.

\subsection{Synthetic Data Generation}
\label{sec:Synthetic-Data-Generation}

\textbf{Isaac Lab Mimic} focuses on synthetically generating a large number of robot demonstrations from a limited set of human demonstrations.
The workflow segments a human demonstration into object-centric subtasks, applies rigid transformations to each segment, and recombines them into new demonstrations~\citep{mandlekar2023mimicgen, jiang2024dexmimicen}. 
By transforming and stitching trajectories, Mimic adapts these demonstrations so that the robot can successfully execute tasks even when the robot or objects occupy poses different from those in the original demonstration, significantly expanding the diversity and coverage of the training dataset.
\begin{figure}[t]
  \centerline{
  \includegraphics[width=0.8\linewidth]{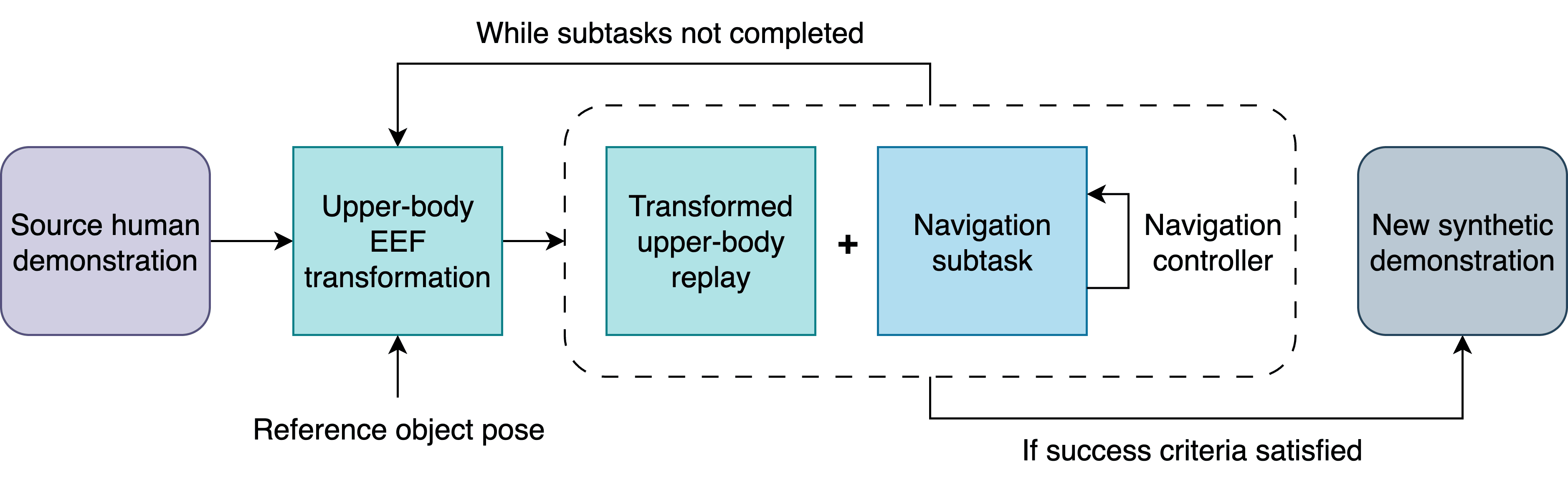}
  }
  \centerline{
    \includegraphics[width=0.325\linewidth]{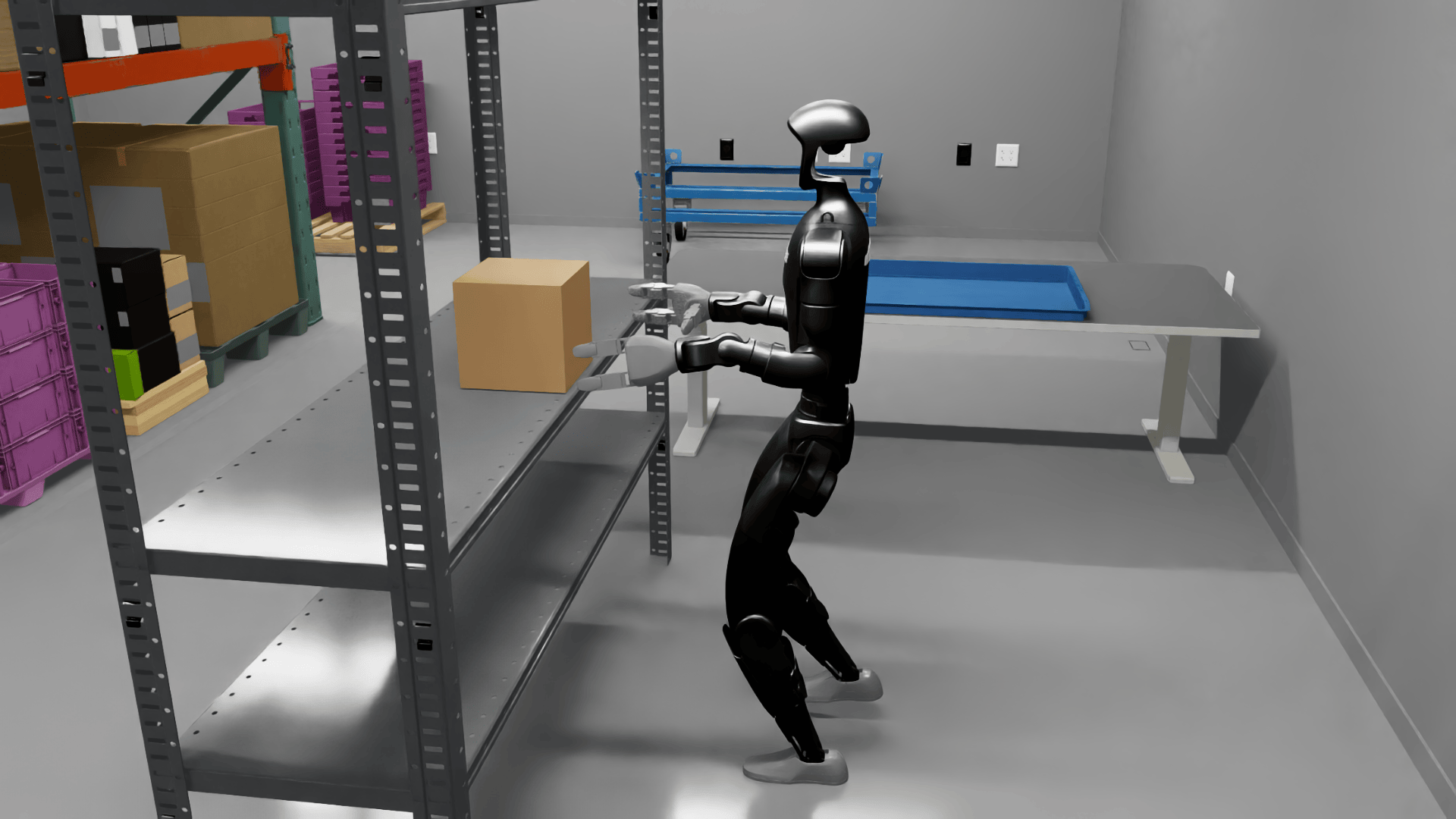}
    \includegraphics[width=0.325\linewidth]{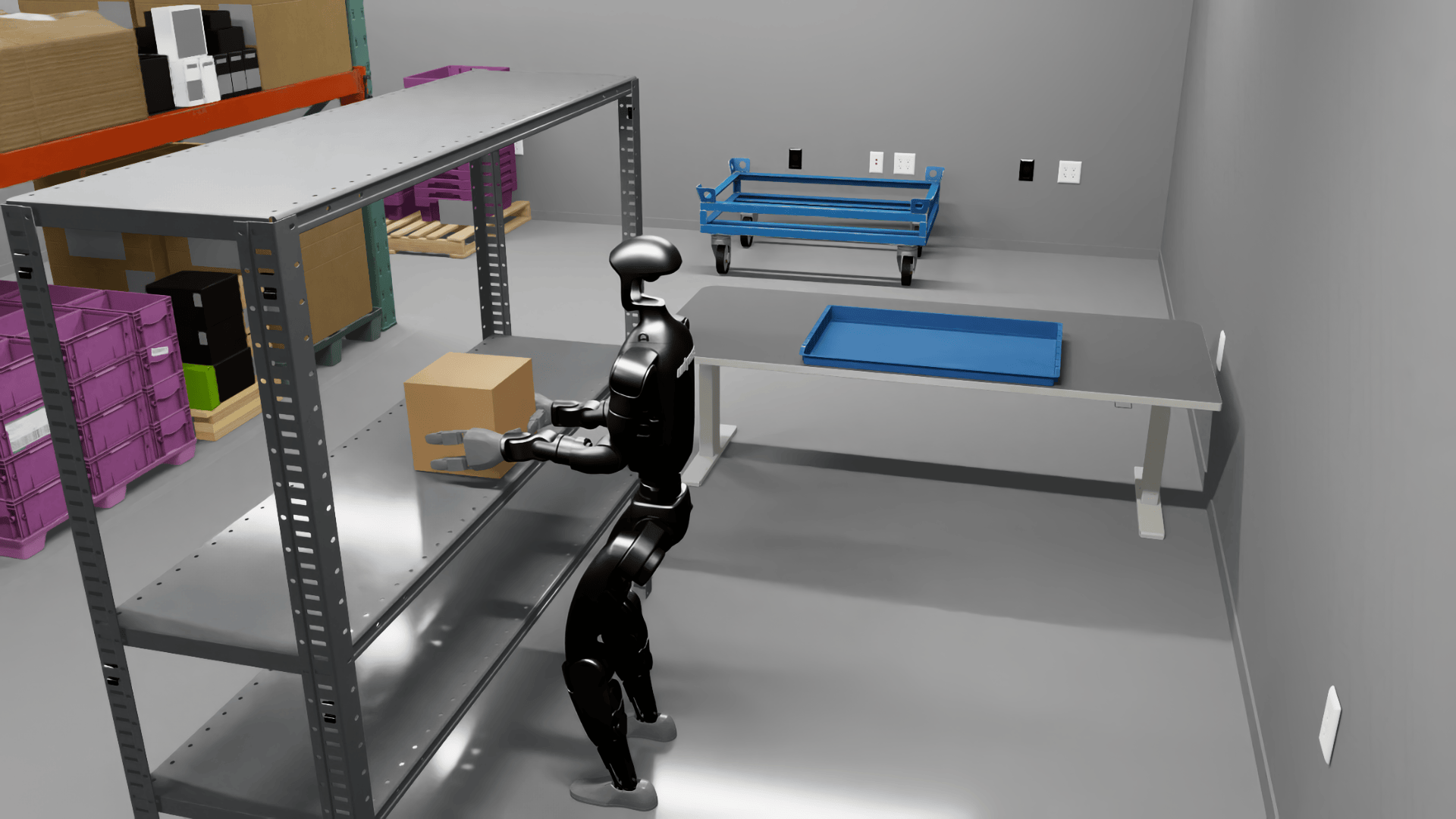}
    \includegraphics[width=0.325\linewidth]{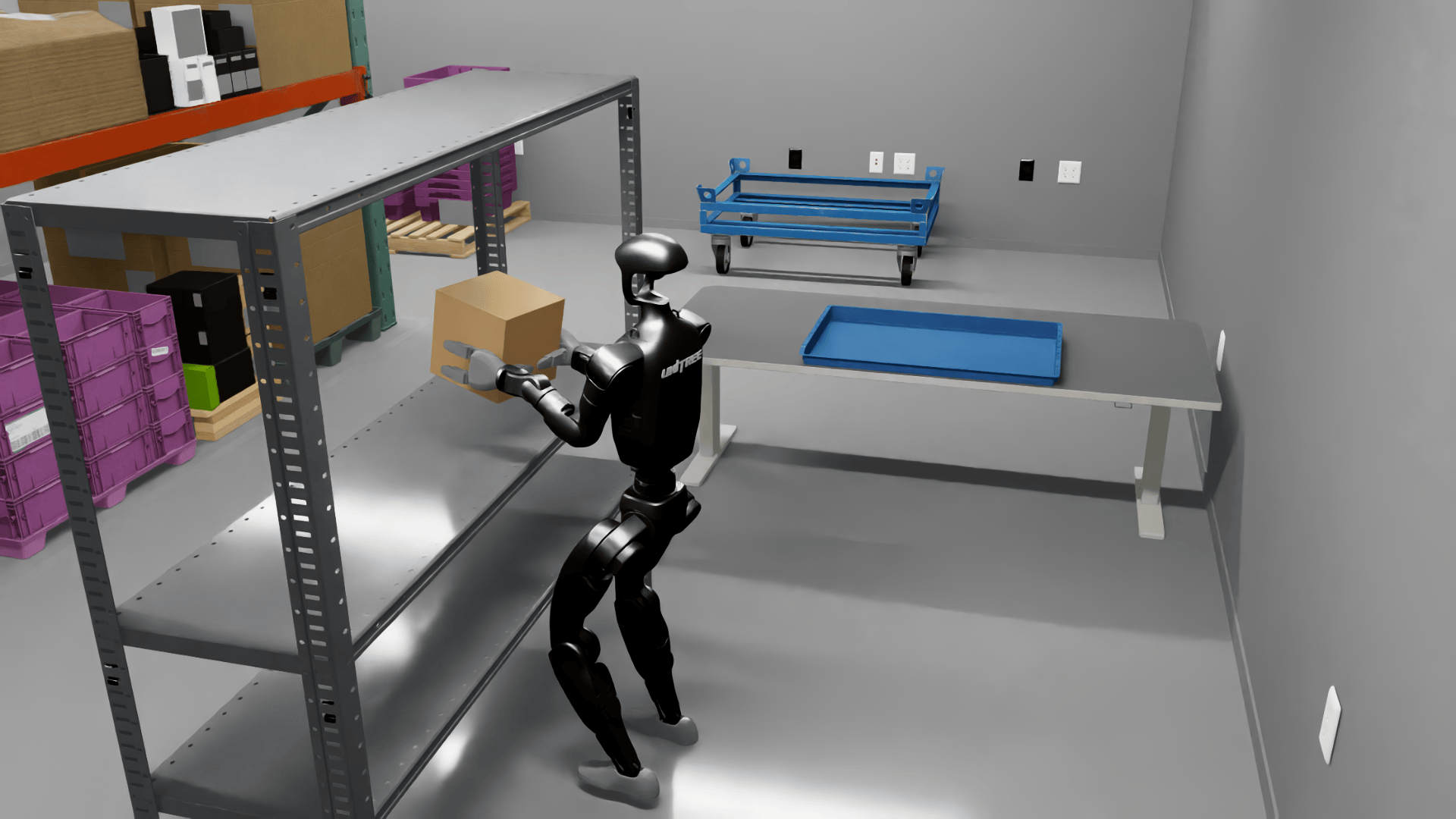}
  }
  \vspace{0.5mm}
  \centerline{
   \includegraphics[width=0.325\linewidth]{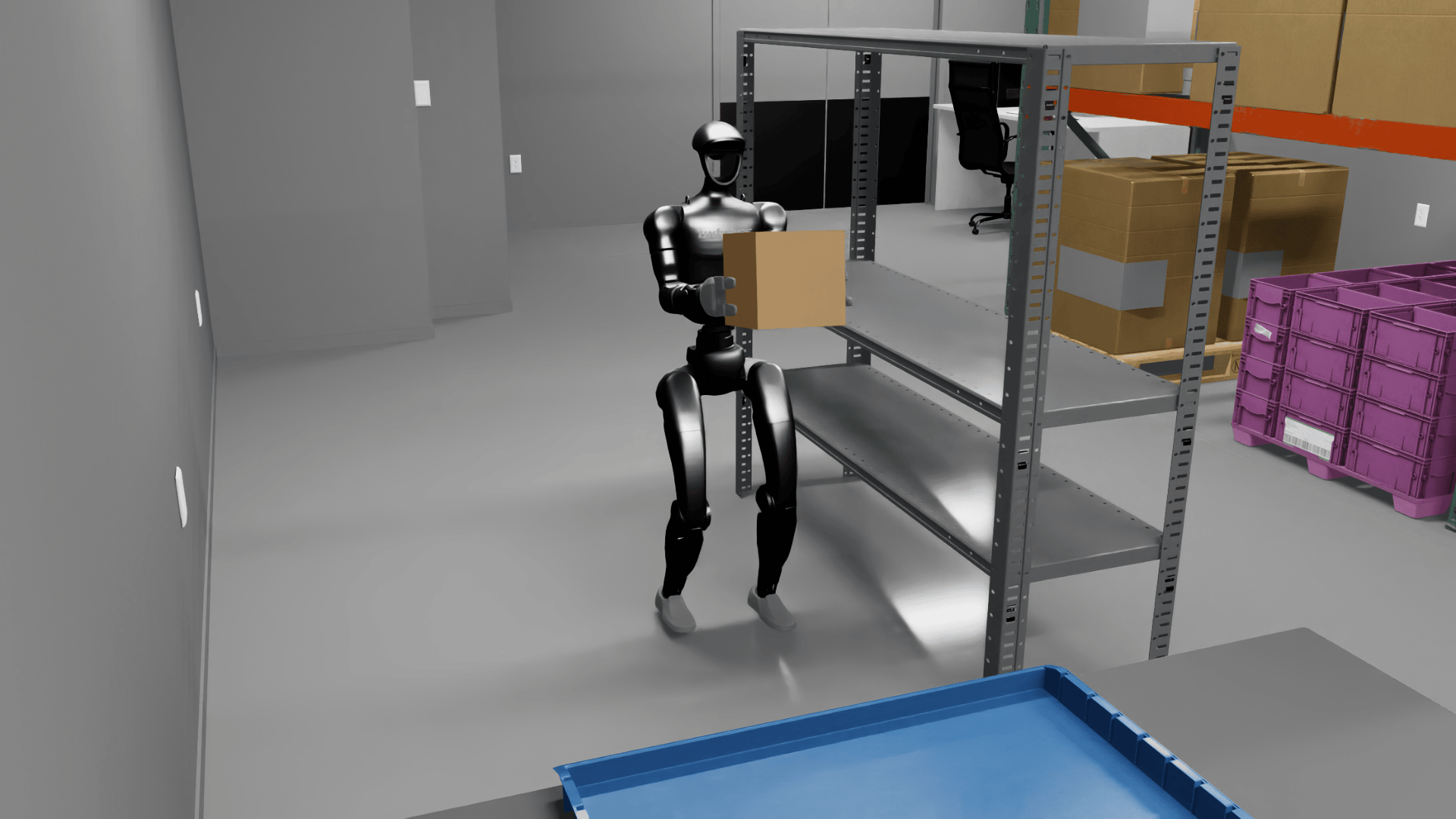}
    \includegraphics[width=0.325\linewidth]{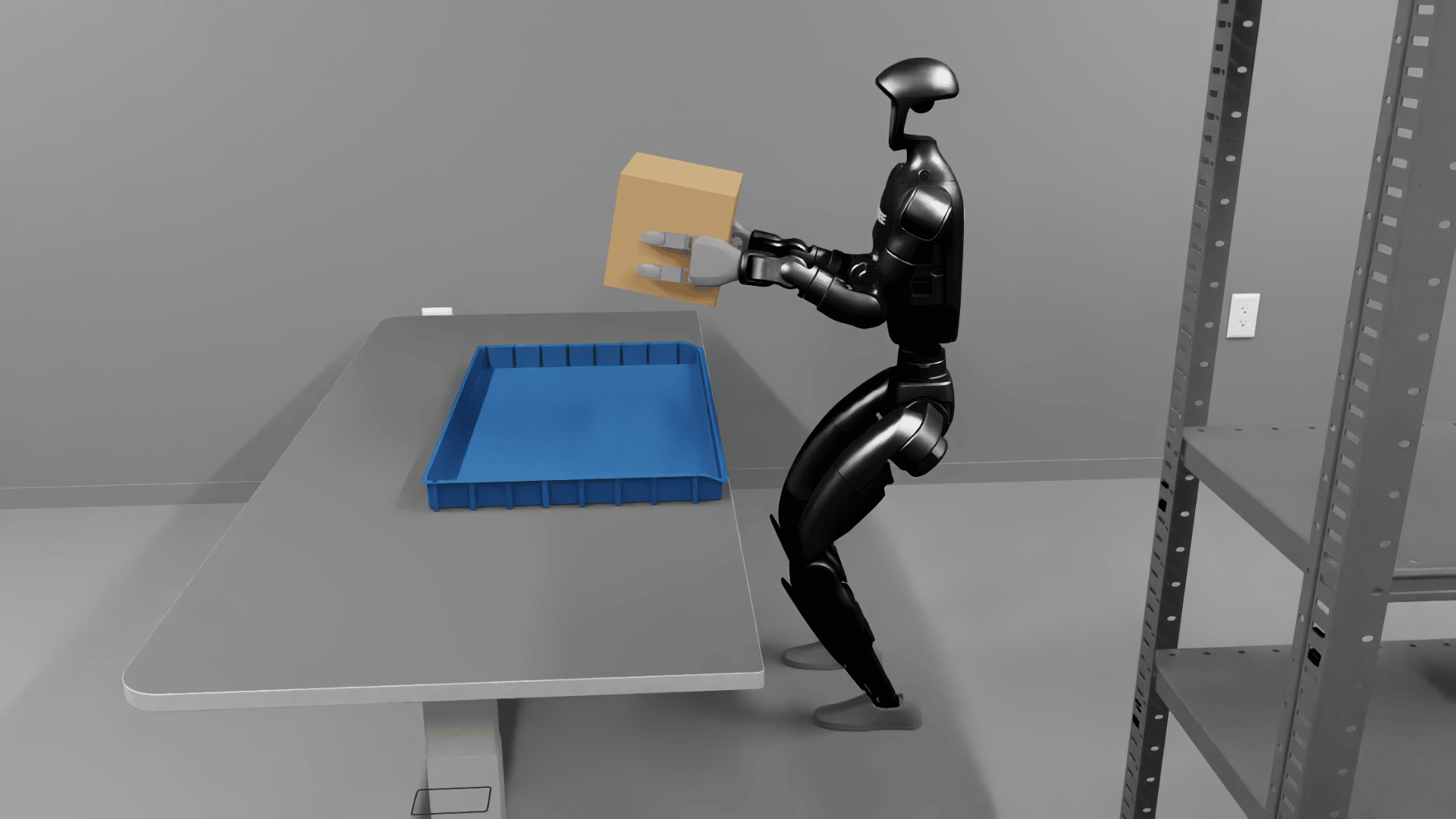}    \includegraphics[width=0.325\linewidth]{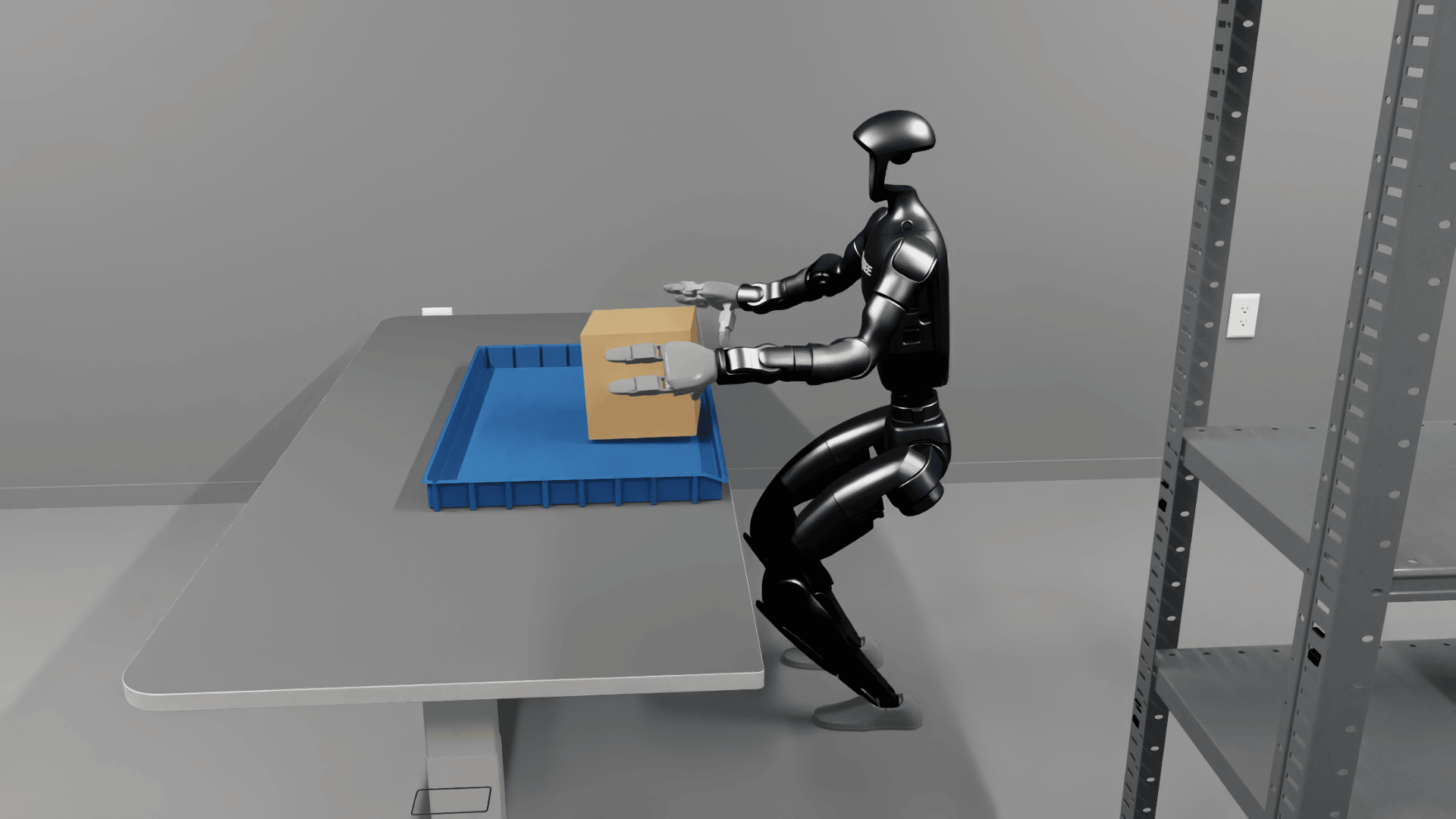}
  }
\caption{Top: The generation pipeline for loco-manipulation synthetic data using Isaac Lab Mimic. Bottom: A loco-manipulation task consisting of multiple subtasks for manipulation and navigation, such as walking to the shelf, picking up the box, turning towards the table, walking to the table, squatting down, and dropping the box. This example is included as part of Isaac Lab Mimic workflow.}
\label{fig:locomanip_mimic_datagen}
\end{figure}

To make a manager-based environment design compatible with the Mimic workflow, users implement the required integration interfaces, including functions to convert robot poses to actions (and vice versa), retrieve the robot end-effector pose, and extract the gripper state from environment actions. Once an environment is Mimic-compatible, it can generate an effectively unbounded number of synthetic demonstrations from as little as a single human demonstration. Furthermore, Isaac Lab Mimic supports parallelized environment execution for data generation, substantially increasing throughput and reducing overall demonstration generation time.

\subsubsection{Loco-Manipulation Data Generation}

An example of the Isaac Lab Mimic workflow is learning loco-manipulation tasks through imitation learning. \cref{fig:locomanip_mimic_datagen} provides an overview of the data generation pipeline that employs a decoupled whole-body controller for a humanoid, similar to~\cite{ben2025homie}.
By combining a whole-body controller with navigation, we synthesize task demonstrations for loco-manipulation, where coordinated locomotion and manipulation enable robots to move (\eg walk or squat) while simultaneously interacting with objects (\eg grasping, pushing, pulling). This coupling supports complex sequences such as picking up an object from a table, traversing space, and placing the object elsewhere.

The system augments demonstrations with randomized pickup and drop-off locations for boxes, and varied positioning of obstacles.
It enhances the data collection pipeline by segmenting manipulation into pick-and-place phases interleaved with locomotion.
Using this pipeline, we generate abundant augmented loco-manipulation data from manipulation-only human demonstrations, allowing humanoid robots to learn integrated loco–manipulation skills. The provided interface in Isaac Lab remains flexible, allowing users to apply different embodiments, including humanoids and mobile manipulators, with their choice of controllers.

\subsubsection{SkillGen-based Dataset Augmentation}\label{subsubsec:skillgen}

SkillGen \citep{garrett2024skillmimicgen} is an automated demonstration generation system in Isaac Lab Mimic that produces high-quality, collision-aware robot demonstrations at scale.
It combines human-provided subtask segments with GPU-accelerated motion planning (described in ~\cref{itm:curobo}) to create diverse feasible trajectories that adapt to new object placements and scene layouts.
By automating transit motions between annotated skills, it reduces manual data collection while improving dataset consistency and validity.

The Isaac Lab integration of SkillGen offers simple controls for scalable dataset creation. Users can select planner-backed generation with a single flag, configure the number of trials, number of parallel environments, and compute devices, and adjust the planner's parameters to balance speed, success rates, and motion quality.
The resulting datasets support \ac{IL} workflows in Isaac Lab.

\section{Application to Robotics Research}

In the following, we provide an overview of the various robotic research areas that have benefited from the capabilities of Isaac Lab. As the framework continues to mature, this section will be expanded to include the latest advancements and applications. We focus this survey primarily on work conducted by closely collaborating academic and industrial partners, including NVIDIA, ETH Z\"{u}rich, and the Robotics and AI Institute (RAI).

\subsection{Locomotion}
\label{subsec:app-locomotion}

\begin{figure}
    \centering
    \includegraphics[width=\linewidth]{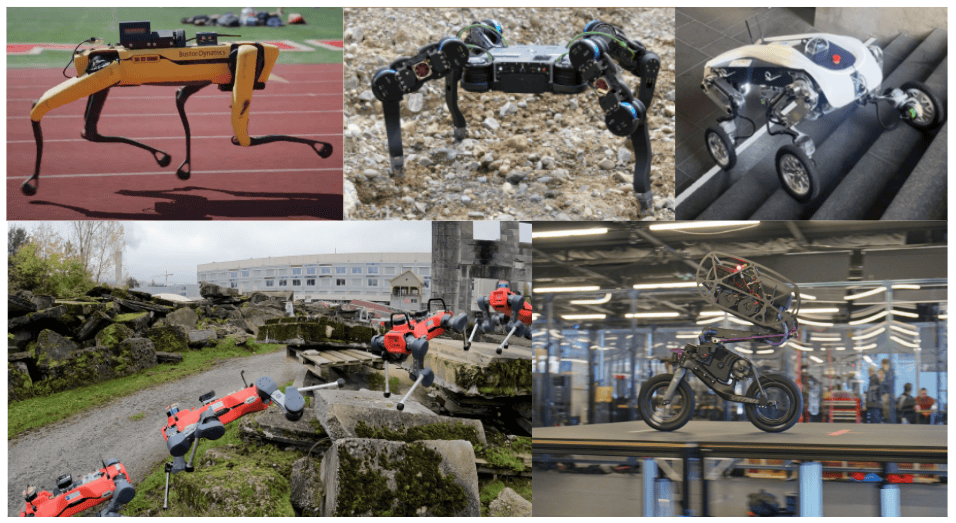}
    \caption{Platforms using Isaac Lab for learning robust and agile locomotion. Top to right: (a) Boston Dynamics Spot~\citep{miller2025high}, (b) Magnecko~\citep{arm2025efficient}, (c) LEVA~\citep{arnold2025leva}, (d) ANYmal~\citep{rudin2025parkour}, and (e) RAI UMV~\citep{rai2025umv}.%
    }
    \label{fig:locomotion}
\end{figure}

Legged locomotion has long posed a central challenge in robotics, due to the difficulty of maintaining balance, coordinating complex movements, and adapting to uncertain or dynamic environments. %
Reinforcement learning enabled a significant breakthrough~\citep{hwango2019actuator}, allowing robots to learn locomotion policies with a level of robustness that was difficult to achieve with traditional, model-based techniques~\citep{lee2020learning, miki2022learning}. Because this learning process requires vast amounts of data, simulation has become an indispensable component for developing controllers safely and efficiently~\citep{rudin2022learning}, with successful deployment across several robotic platforms, as depicted in \cref{fig:locomotion}.

The RAI Institute released the first open-source end-to-end training pipeline for Boston Dynamics' Spot using Isaac Lab, along with a method to close the sim-to-real gap via evolutionary strategies~\citep{miller2025high}, achieving zero-shot transfer from simulation to hardware with running speeds up to 5.2 m/s, nearly three times the default limit. Their approach closely models hardware-specific dynamics, including actuator delays and joint torque limits, using custom actuator classes in Isaac Lab to enable robust and dynamic real-world performance.
~\cite{wen2025constrained} optimizes a style-imitation objective with constraints to learn locomotion for quadruped and humanoid from imperfect demonstrations.~\cite{arm2025efficient} explores how to develop and validate a locomotion controller and a base pose controller in gravity environments from lunar gravity to a hypothetical super-Earth. \cite{cathomen2025divide} proposes an unsupervised skill discovery method to learn various locomotion skills for a quadruped without explicit task-specific rewards.~\cite{li2025marladona} creates an open-source multi-agent soccer environment and uses multi-agent reinforcement learning (MARL) to train decentralized policies for sophisticated team play behavior.~\cite{arnold2025leva} uses Isaac Lab to train an RL controller for a wheeled quadruped capable of transportation of materials over challenging terrains.

Although locomotion can be learned with purely proprioceptive information, incorporating exteroceptive feedback from sensors such as LiDARs or cameras can greatly increase the capabilities of learned controllers. For example, \citet{rudin2025parkour} trains a quadruped robot to climb obstacles and traverse unstructured terrain using depth images as input to an end-to-end visuomotor policy. It first applies reinforcement learning on a simplified terrain representation, then distills the policy with depth-image observations, and finally performs RL fine-tuning. The required algorithms are tightly integrated with Isaac Lab through the RSL-RL library~\citep{schwarke2025rslrl}. \cite{bjelonic2025towards} investigates bridging the sim-to-real gap for multiple legged robots by employing evolutionary algorithms for system identification, tuning simulation parameters to better match real-world robot trajectories.

Beyond legged locomotion, Isaac Lab has also been applied to wheeled platforms, such as the Ultra Mobility Vehicle (UMV), developed by the RAI Institute (\cref{fig:bd}). UMV enhances the bicycle form with a custom jumping mechanism, enabling both efficient wheeled locomotion and dynamic legged behaviors, such as hopping, flipping, and obstacle clearance. Achieving these behaviors requires accurate modeling of wheel–ground dynamics and the use of domain randomization to capture the uncertainties of high-energy impacts. Isaac Lab supports both aspects, allowing UMV to deploy RL policies with robust sim-to-real performance. This demonstrates how Isaac Lab facilitates high-performance mobility research on robotic systems that extend beyond conventional legged designs~\citep{rai2025umv}.

\subsection{Whole-body control}

Whole-Body Control (WBC) considers the robot as a single, integrated system, enabling simultaneous coordination of locomotion, manipulation, and environmental interaction across all available degrees of freedom. This holistic approach allows robots to perform complex, multi-objective tasks—such as walking while carrying objects, stabilizing against external forces, or adjusting posture in real time to maintain balance—making WBC essential for deployment in unstructured and dynamic environments. However, achieving effective WBC in practice presents significant challenges. It requires resolving multiple, often competing objectives (\eg balance vs. manipulation accuracy), handling kinematic and dynamic constraints, and managing the interactions between subsystems such as arms, legs, and torso. Additionally, WBC must be robust to modeling inaccuracies, sensor noise, and latency in perception and actuation. Despite these complexities, recent advances have accelerated progress in the field. Learning-based methods, motion retargeting from human demonstrations, and task prioritization strategies are increasingly integrated into modern WBC pipelines. Isaac Lab has further enabled scalable development and evaluation of WBC policies, offering high-fidelity simulation, physics realism, and seamless deployment on real-world hardware.

\begin{figure}[H]
  \centerline{
  \includegraphics[width=0.52\linewidth, trim={300, 0, 0, 0}, clip]{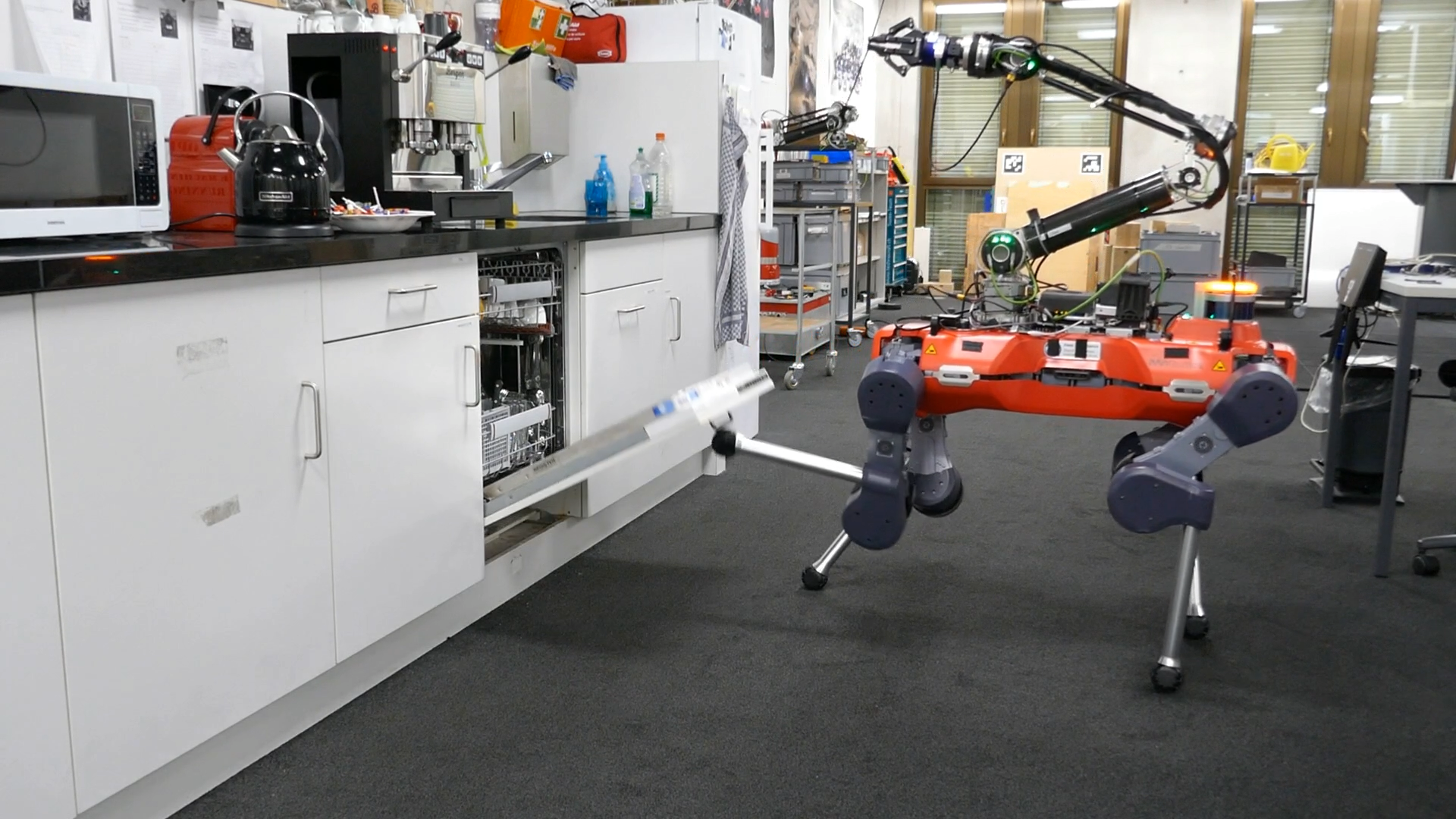}
  \includegraphics[width=0.477\linewidth]{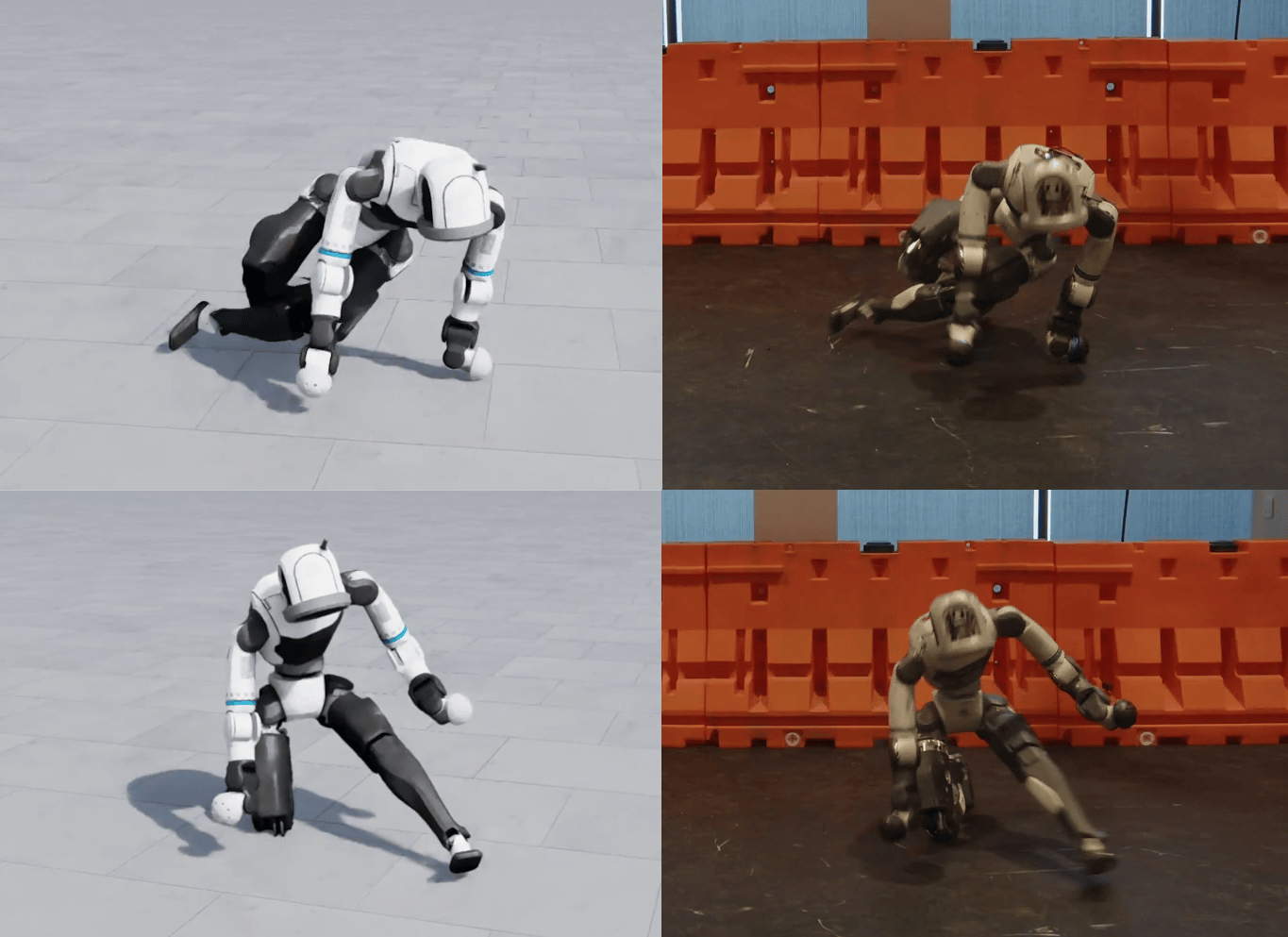}
  }
\caption{{Left:} The ANYmal robot with an arm interacting with a dishwasher~\citep{sleiman2024guided}. {Right:} Boston Dynamics Atlas robot to perform a wide array
of athletic skills, sourced from diverse human motion data, including video and motion capture~\citep{eatlas2025rai}.}
\label{fig:bd}
\end{figure}

Building on the idea of leveraging humanoid motion datasets for general-purpose control, \cite{he2024hover} proposes a policy distillation framework that unifies different control modes under a single WBC policy. A re-implementation of the \href{https://github.com/NVlabs/HOVER}{HOVER framework with Isaac Lab} enhances the original implementation with extensive sim-to-sim and sim-to-real evaluations, demonstrating the effectiveness of Isaac Lab training in comparison to the initial implementation in Isaac Gym. \cite{sleiman2024guided} applies DeepMimic-style training to long-horizon loco-manipulation tasks, where a quadrupedal robot with an arm learns to interact with articulated objects like dishwashers and spring-loaded doors. \cite{stolle2024perceptive} introduces a perceptive pedipulation policy that uses elevation maps to perform local collision avoidance while tracking foot positions, allowing the robot to safely navigate around dynamic obstacles using contact filtering in Isaac Lab. 

On the control side, \cite{portela2024whole} develops methods for robust end-effector pose tracking on rough terrains, while \cite{vijayan2025multi} proposes a multi-critic setup for smoother whole-body control through robust end-effector twist tracking. ~\cite{dadiotis2025dynamic} develops a learning-based controller for a mobile manipulator to move an unknown object to a desired position and yaw orientation through a sequence of pushing actions. The proposed controller for the robotic arm and the mobile base motion is trained using a constrained RL formulation. 

Isaac Lab has also been used to develop RL policies for humanoid platforms such as Boston Dynamics' Atlas~\citep{eatlas2025rai, eatlas2025bd}. Leveraging human motion capture and animation data, the trained policies enable a diverse set of whole-body movements, including army crawling, shoulder rolls, hand-stand forward rolls, and breakdance motions (\cref{fig:bd}). This work showcases rich multi-contact behaviors and demonstrates Isaac Lab's ability to support RL pipelines that extend humanoid control beyond conventional locomotion. Together, these works push humanoid capabilities beyond locomotion into rich loco-manipulation and robust interaction in unstructured environments.

\subsection{Navigation}

Navigation remains a central challenge in robotics, which requires an integration of perception, planning, and control across diverse embodiments and environments.  
Isaac Lab has rapidly become a preferred framework for learned navigation research, as demonstrated by a growing body of recent works spanning wheeled, legged, aerial, and even aquatic and space robots, as well as different learning paradigms.

\begin{figure}[H]
    \centering
    \includegraphics[width=\linewidth]{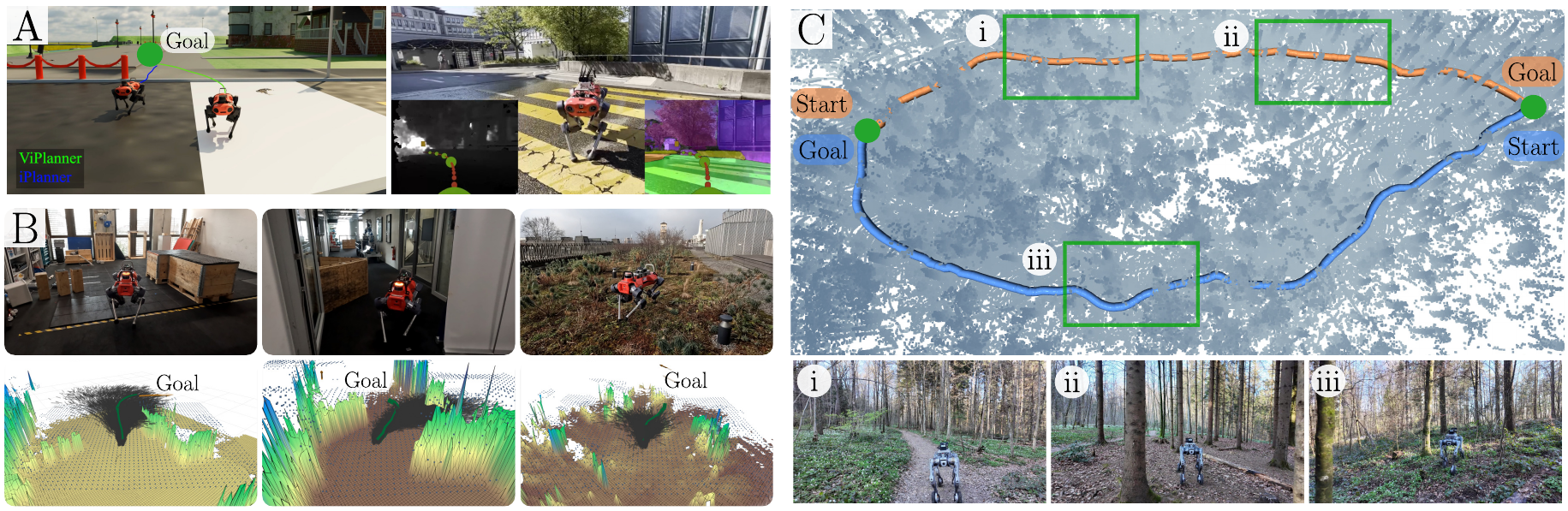}
    \caption{Sim-to-real navigation with Isaac Lab: (a) ViPlaner~\citep{roth2024viplanner} learns an end-to-end policy from depth and semantic images. (b) A perceptive Forward Dynamics Model~\citep{roth2025learned} trained in simulation and deployed with a sampling-based planner. (c) An RL navigation policy with a novel memory unit for spatio-temporal reasoning~\citep{yang2025improving}.}
    \label{fig:applications_navigation}
\end{figure}

Navigation requires policies to have a precise understanding of the environment and the skills of the embodiment. For the former, Isaac Lab’s sensor ecosystem (\cref{subsec:sensors}) provides physically and photometrically realistic modalities at scale, making it well-suited for vision-based navigation.  
This has enabled works such as ViPlanner~\citep{roth2024viplanner}, which trains an end-to-end semantic local planner purely with data generated in Isaac Lab (\cref{fig:applications_navigation}), and NaVILA~\citep{cheng2024navila}, which extends navigation to vision-language-action models for legged robots. 
By combining sensors with the accurate PhysX simulation of Isaac Lab, researchers at ETH have developed \ac{RL} based navigation policies with a novel recurrent memory architecture for long-range navigation~\citep{yang2025improving}.  
Another line of navigation research focuses on learning forward-dynamics models for safe motion planning, which can be established using the accurate dynamics provided within Isaac Lab~\citep{roth2025learned}.

\begin{figure}[H]
\centering
\includegraphics[width=\linewidth] {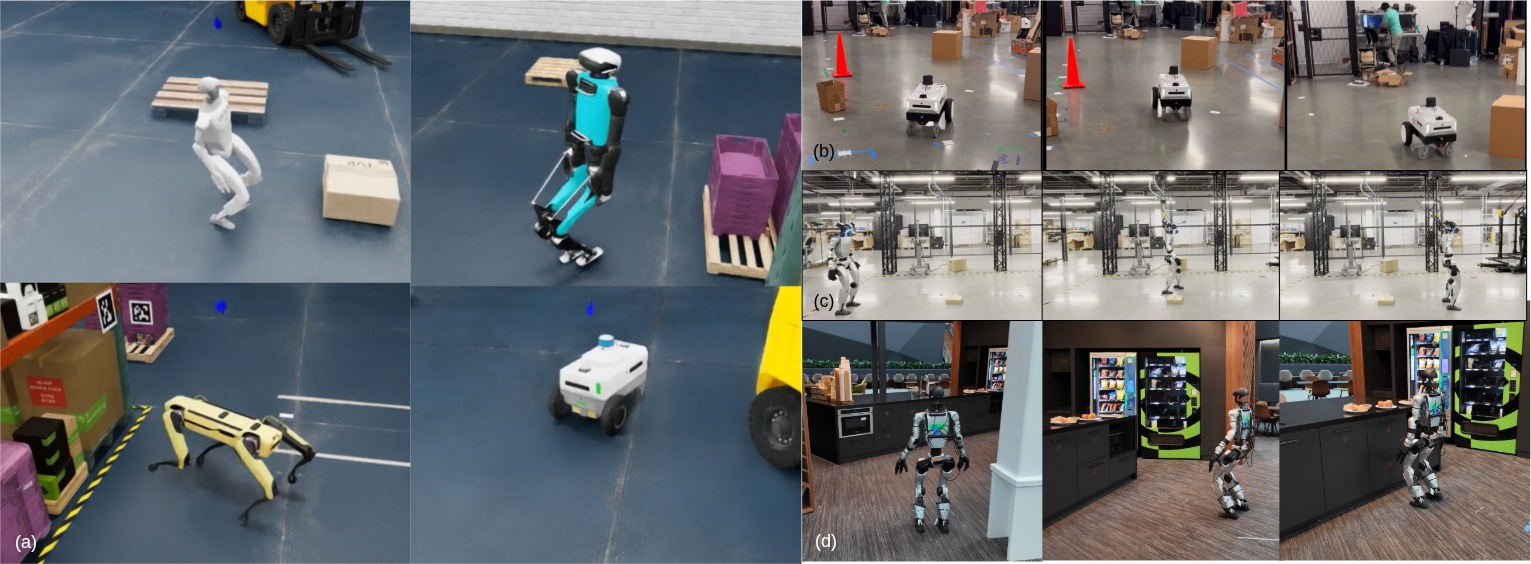}
\caption{COMPASS~\citep{liu2025compass} trains residual RL policies in diverse simulated scenes for cross-embodiment mobility. It enables sim-to-real transfer to a wheeled platform, a quadruped and a humanoid. Fine-tuning GR00T N1.5 on COMPASS distillation datasets further demonstrates zero-shot sim-to-real navigation.}
 \label{fig:compass}
\end{figure}

A unifying factor across these contributions is Isaac Lab’s ability to support \emph{multi-platform training} through the manager-based workflow (\cref{subsec:manager-based-worflow}), where switching between environment scenes and robot embodiments requires only minimal reconfiguration while preserving realistic physical behavior.
Building on this foundation, a central goal in robot learning is to develop generalizable navigation policies that transfer across embodiments and adapt efficiently to real-world conditions.
COMPASS~\citep{liu2025compass}, a novel workflow for developing cross-embodiment mobility policies by integrating imitation learning, residual RL, and policy distillation, has demonstrated the effectiveness of RL fine-tuning and strong zero-shot sim-to-real performance using Isaac Lab (see~\cref{fig:compass}). COMPASS can also provide navigation specialist policies for generating large-scale synthetic datasets in Isaac Lab to train advanced VLA foundation models (\eg Gr00t N1.5). By leveraging COMPASS, the USD scenes generated from the real-to-sim NuRec pipeline, such as \href{https://huggingface.co/datasets/nvidia/PhysicalAI-Robotics-NuRec}{these example assets}, can be used to further reduce the sim-to-real gap. 

\subsection{Industrial Manipulation}

Industrial manipulation frequently involves contact-rich interactions, where robots must maintain sustained physical contact with objects and precisely regulate forces and motions under uncertainty. In contrast to simple pick-and-place operations, such tasks demand careful handling of friction, compliance, and alignment, and are essential for assembly processes like peg insertion, gear meshing, threaded-fastener mating, and snap-fit assembly. Isaac Lab provides simulation capabilities, robotic assembly environments, and algorithms for simulating and learning skills for precise contact-rich manipulation.

\begin{wrapfigure}{R}{0.5\textwidth}
\vspace{-3mm}
  \centerline{
 \includegraphics[width=0.48\linewidth]{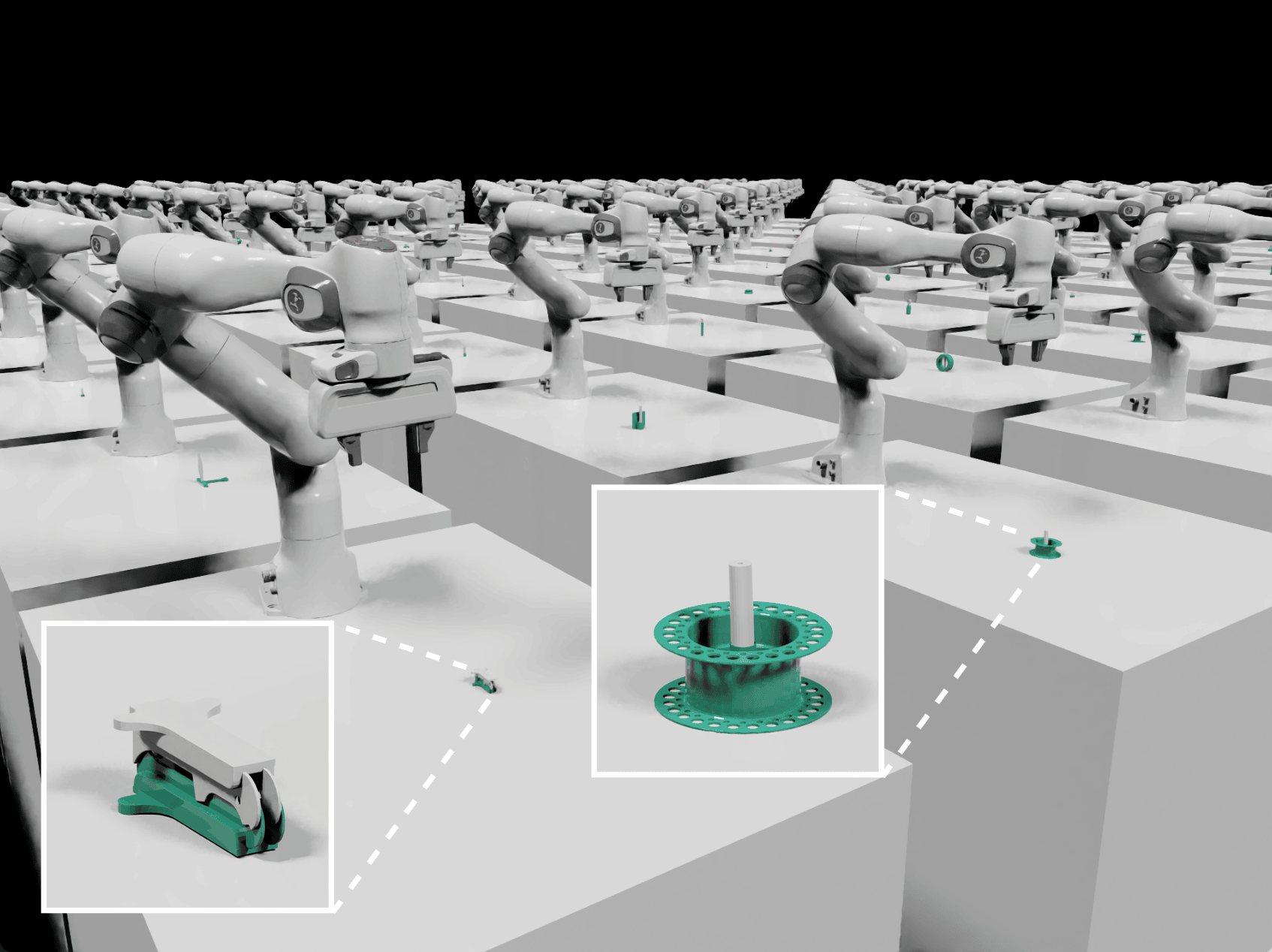}
    \includegraphics[width=0.48\linewidth]{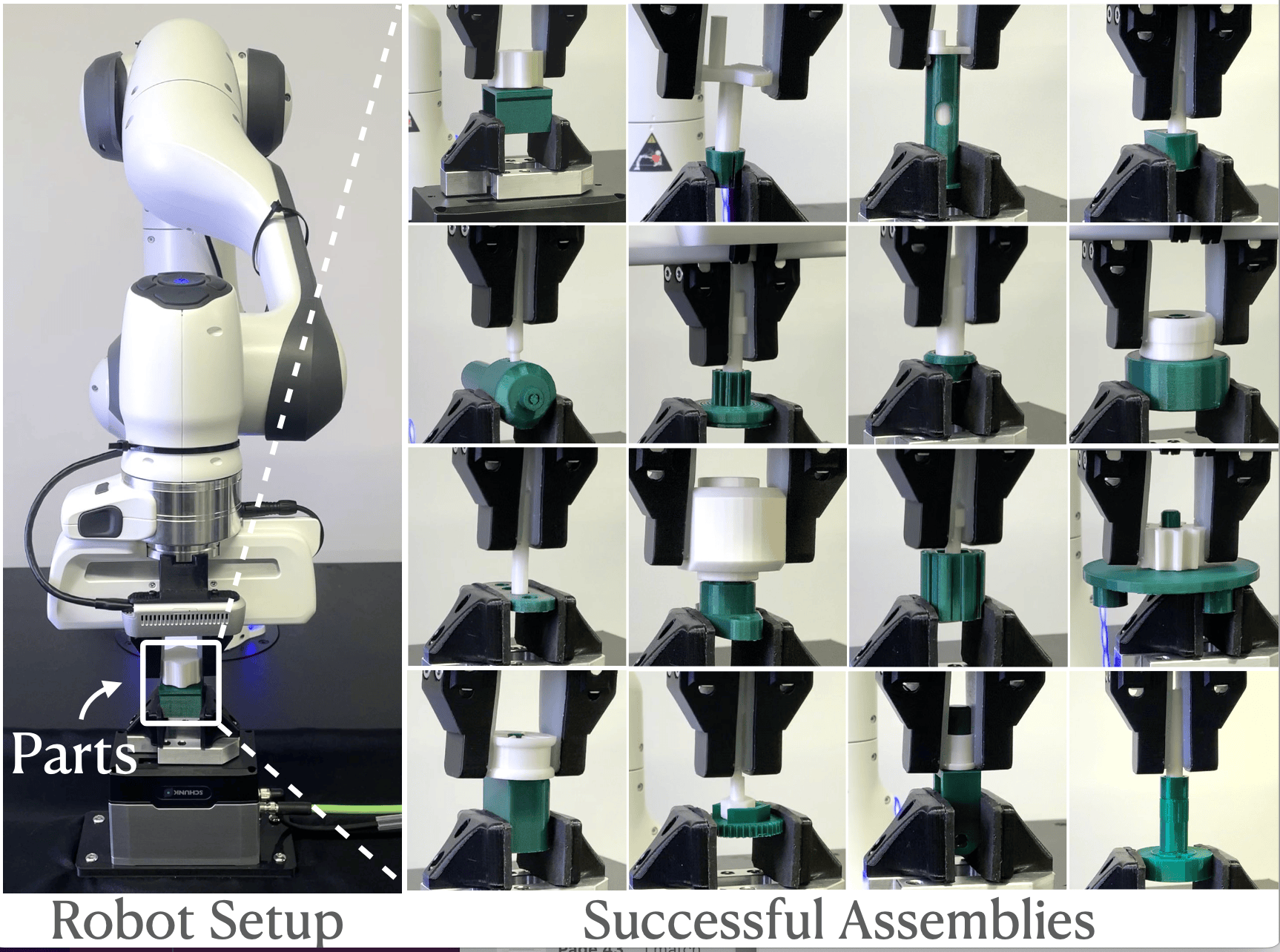}
  }
\caption{Assembly Environments in Isaac Lab~\citep{tang2023industreal}. Left: Simulation. Right: Real World.}
\label{fig:assembly-envs}
\end{wrapfigure}

The \texttt{Factory} environments combine SDF-based contact generation, a contact reduction technique, and a Gauss–Seidel solver for efficient contact simulation~\citep{narang2022factory}. Policies trained in these environments can achieve zero-shot sim-to-real transfer with 83–99\% success rates~\citep{tang2023industreal}. The AutoMate environments extend these capabilities towards more challenging geometries and generalizable assembly by combining reinforcement and imitation learning and train multi-task policies~\citep{tang2024automate}. The environments include 100 simulation-compatible assembly assets, specialist policies for $\sim$80 tasks, and a distilled generalist policy for 20 tasks, all achieving around 80\% success rates both in simulation and in the real world. See~\cref{fig:assembly-envs}.

SRSA~\citep{guo2025srsa} enables continual learning by selecting and fine-tuning pre-trained skills from a task library based on task similarity, while MatchMaker~\citep{wang2025matchmaker} procedurally expands this library for scalable training. FORGE~\citep{noseworthy2025forge} introduces force-aware environments with features such as adaptive force regulation, thresholding, and randomized dynamics to enable robust execution of precision tasks like snap-fit insertion and gear meshing under uncertainty. Complementing these approaches, TacSL~\citep{akinola:tro2025} uses visuotactile sensing to handle occlusion and lighting variation, enabling precise contact-rich manipulation with reliable sim-to-real transfer, supported by Isaac Lab’s tactile simulation tools.

\begin{figure}[H]
\centering
\includegraphics[width=0.49\linewidth] {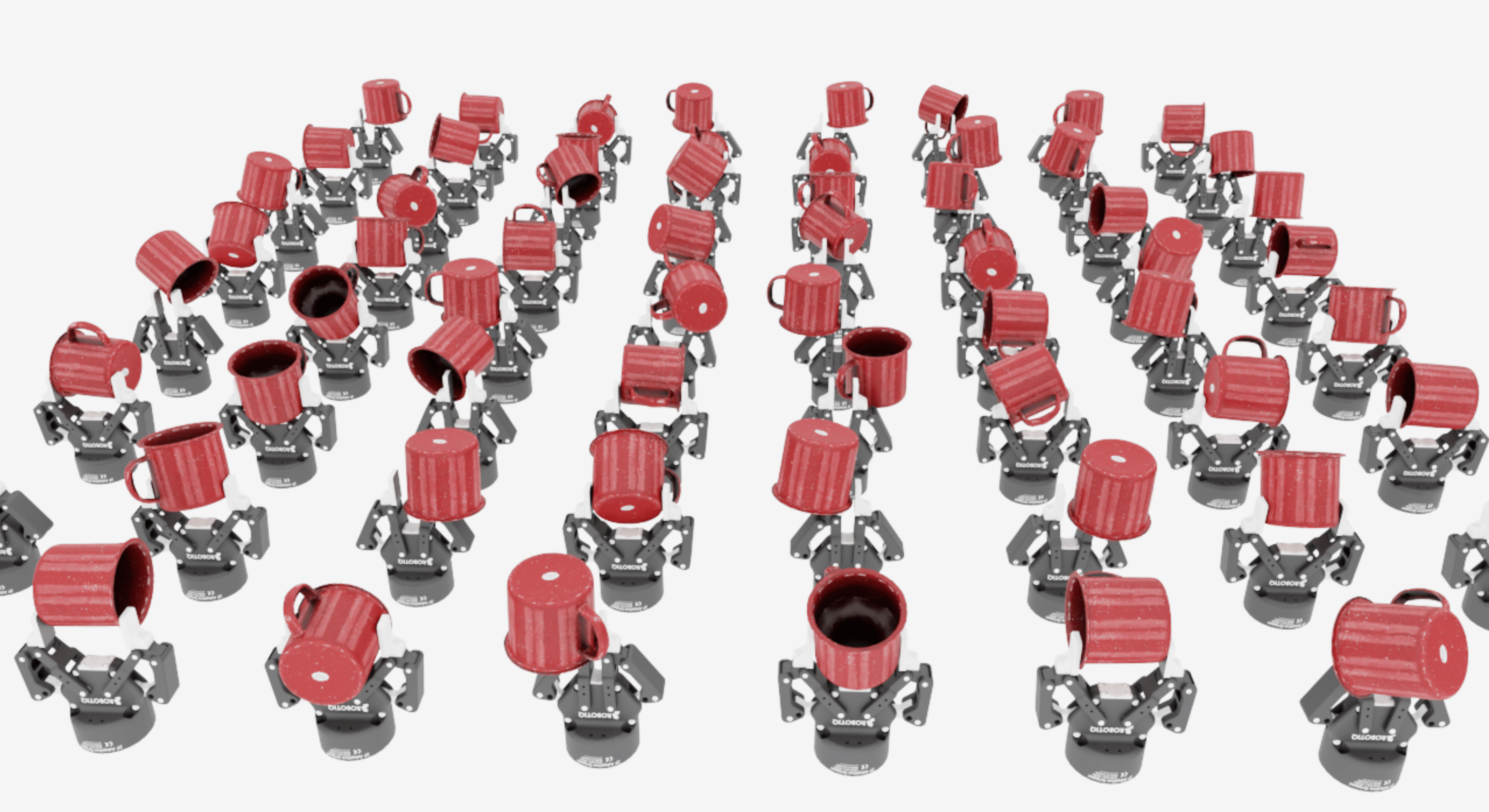}
\hfill
\includegraphics[width=0.47\linewidth, trim={0 100 0 0}, clip] {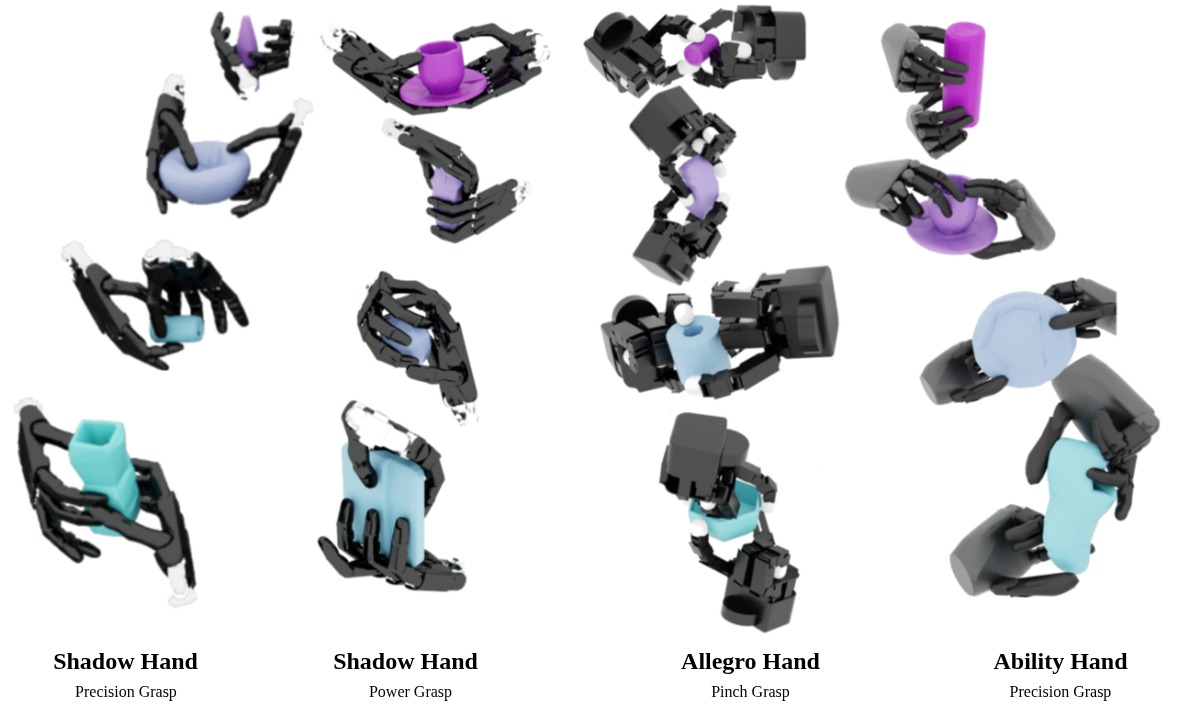}
 \caption{Left: GraspDataGen~\citep{carlson2025graspdatagen} exhaustively evaluates 6 DoF grasps given an object and gripper asset. Right: GraspQP~\citep{zurbrugg2025graspqp} evaluates grasp candidates, synthesized using an analytical formulation, on various multi-DoF grippers in Isaac Lab.}
 \label{fig:issaclabgrasp}
\end{figure}

In addition to assembly tasks, grasping is also a foundational task in robotic manipulation, requiring precise coordination between sensing, planning, and control to enable robots to reliably interact with objects in diverse and often unstructured environments. Isaac Lab advances this capability by providing high-fidelity simulation, accurate contact modeling, and GPU-accelerated data generation pipelines. 
vMF-Contact~\citep{shi2024vmf} introduces a novel probabilistic architecture for learning hierarchical contact grasp representations and uses Isaac Lab for both data collection and task-level grasp evaluation. 
GraspDataGen ~\citep{carlson2025graspdatagen} (see~\cref{fig:issaclabgrasp}) recently introduces a new Isaac Lab-based pipeline for modular 6 DoF grasp sampling and evaluation conditioned on object meshes. Several common industrial pinch grippers (Robotiq 2F-85, 2F-140, RG6, Franka-Panda) are supported out of the box and can be used to sample a dense set of Isaac Lab verified grasp poses for custom objects. 
Additionally, GraspQP~\citep{zurbrugg2025graspqp} introduces modular Isaac Lab environments that extends grasp evaluation to multi-DoF grippers, including ShadowHand, AbilityHand, Allegro, and both two- and three-fingered Robotiq grippers. 

\subsection{Dexterous Manipulation}

\begin{wrapfigure}{R}{0.6\textwidth}
\vspace{-3mm}
  \centerline{
  \includegraphics[width=.605\linewidth]{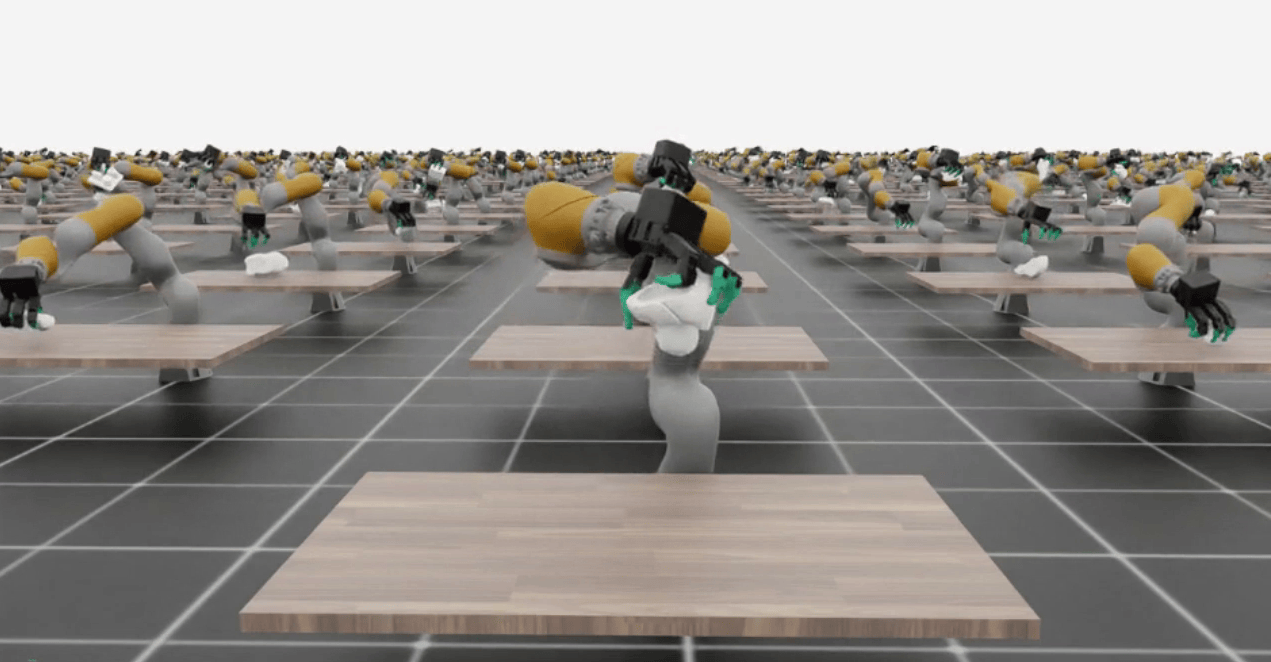}
  \includegraphics[width=.38\linewidth]{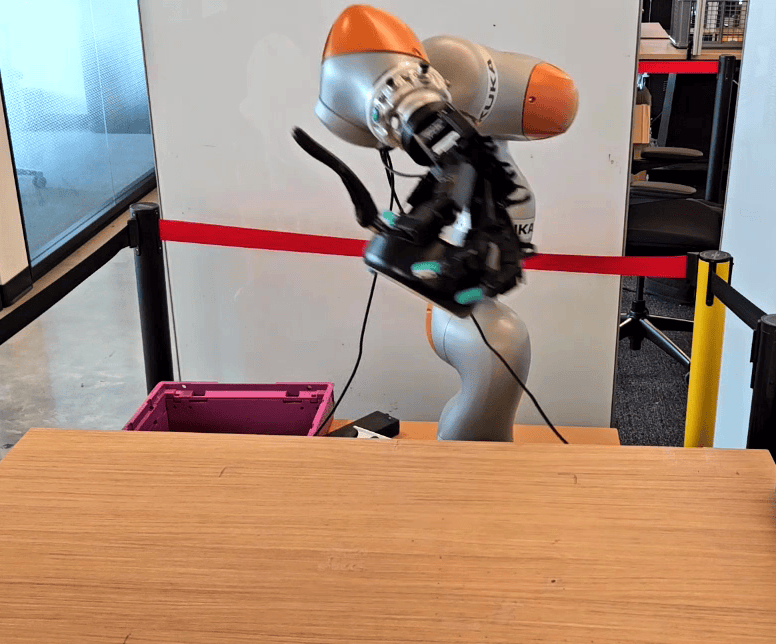}
  }
  \centerline{
    \makebox[0.61\linewidth][c] { \footnotesize{(a) Simulation} }
    \hfill
    \makebox[0.38\linewidth][c] { \footnotesize{(b) Real} }
  }
\caption{Left: DextrAH~\citep{DextrAH-RGB24} environment training in simulation. Right: The trained policy deployed in the real world.}
\label{fig:dextrah_sim_real}
\vspace{-2mm}
\end{wrapfigure}

Dexterous manipulation with multi-fingered hands remains challenging compared to standard parallel-jaw grasping due to the high-dimensional action space and fine-grained control required for coordinated finger movements and in-hand object manipulation. While most grasping systems operate open-loop and predict gripper poses for simple contact, they often fall short in dynamic tasks or tasks with high precision requirements. Isaac Lab enables the development of advanced dexterous manipulation policies by offering high-fidelity simulation of multi-DOF hands, accurate contact modeling, and support for rich sensory inputs, including vision and proprioception. It allows for scalable policy training and deployment of learned dexterous manipulation policies on real-world robotic platforms through features such as domain randomization and tiled rendering.

DextrAH-RGB \citep{DextrAH-RGB24} trains an RL policy on the KUKA arm with an Allegro hand in simulation that uses privileged state information and then distills this into a network that takes stereo RGB pairs as input. This is the first system to have shown that an end-to-end network directly operating on raw RGB streams can control arm and multi-fingered hands, all leveraging the high-quality rendering from Isaac Lab (\cref{fig:dextrah_sim_real}). 
More recently, ~\cite{Singh2025_E2E} leverages Isaac Lab to train depth-based end-to-end policies from scratch for the DextrAH \citep{DextrAH-G24, DextrAH-RGB24} task.

\begin{figure}[H]
    \centering
    \includegraphics[width=1\textwidth]{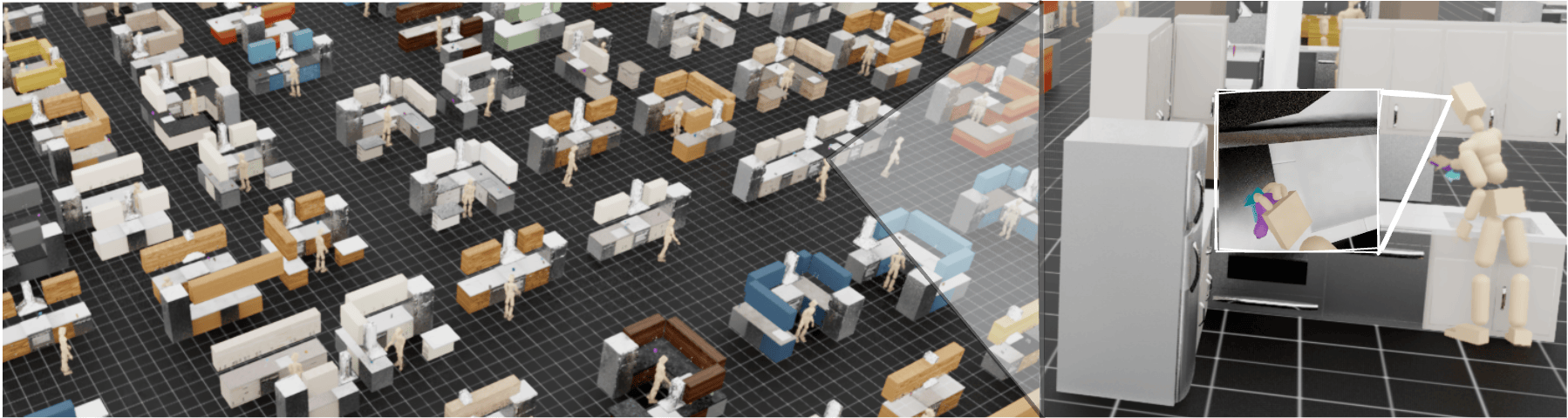}
    
    \captionof{figure}{{Perceptive Dexterous Control (PDC)~\citep{luo2025emergent} enables a humanoid equipped with egocentric vision to search, reach, grasp, and manipulate objects in cluttered kitchen scenes. PDC uses visual perception as the sole indicator of which hand to use, which object to grasp, where to move, and which drawer to open.}}
    \label{fig:pdc_teaser}
\end{figure}

Perceptive Dexterous Humanoid Control (PDC) \citep{luo2025emergent} demonstrates vision-driven whole-body control of simulated humanoids (shown in \Cref{fig:pdc_teaser}), closing the perception–action loop through egocentric visual input. 
Prior approaches often rely on privileged object states from simulation, limiting the emergence of human-like behaviors such as active search. The PDC framework addresses this by using egocentric vision and proprioception only, introducing a perception-as-interface paradigm with visual cues (\eg masks, 3D markers, hand indicators) for tasks like search, grasp, placement, and drawer manipulation.
Leveraging Isaac Lab’s large-scale GPU-accelerated simulation, policies are trained via \ac{RL} with motion priors \citep{luo2023universal}, scaling across procedurally generated kitchens and tabletop tasks. Results show that training directly from pixels, rather than distilling from state-based policies, led to better generalization and more natural behaviors. This highlights the power of Isaac Lab as a platform for scalable visual RL in complex loco-manipulation tasks. 

\begin{wrapfigure}{R}{0.6\textwidth}
\vspace{-5mm}
  \centerline{
  \includegraphics[width=.45\linewidth]{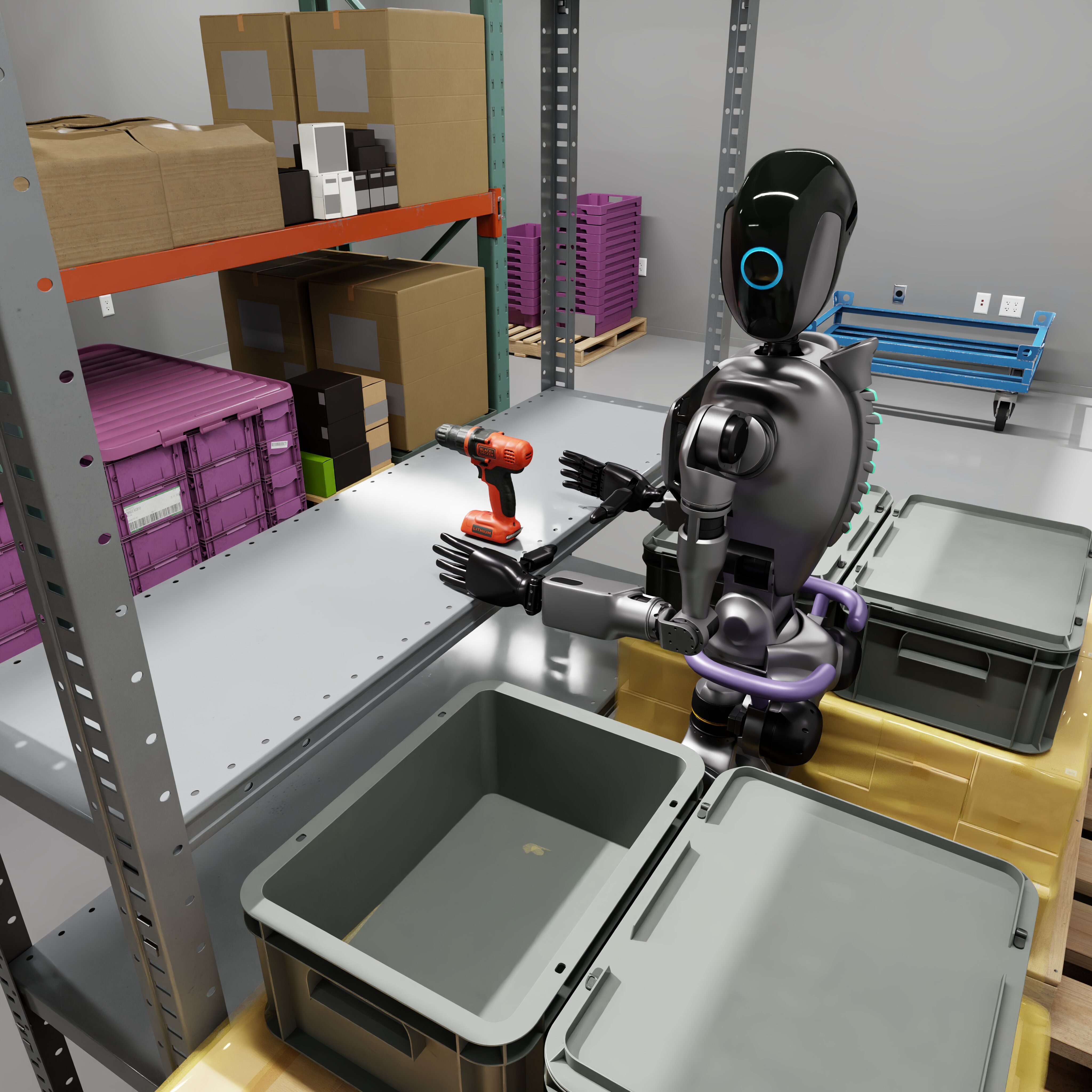}
  \hfill
  \includegraphics[width=.45\linewidth]{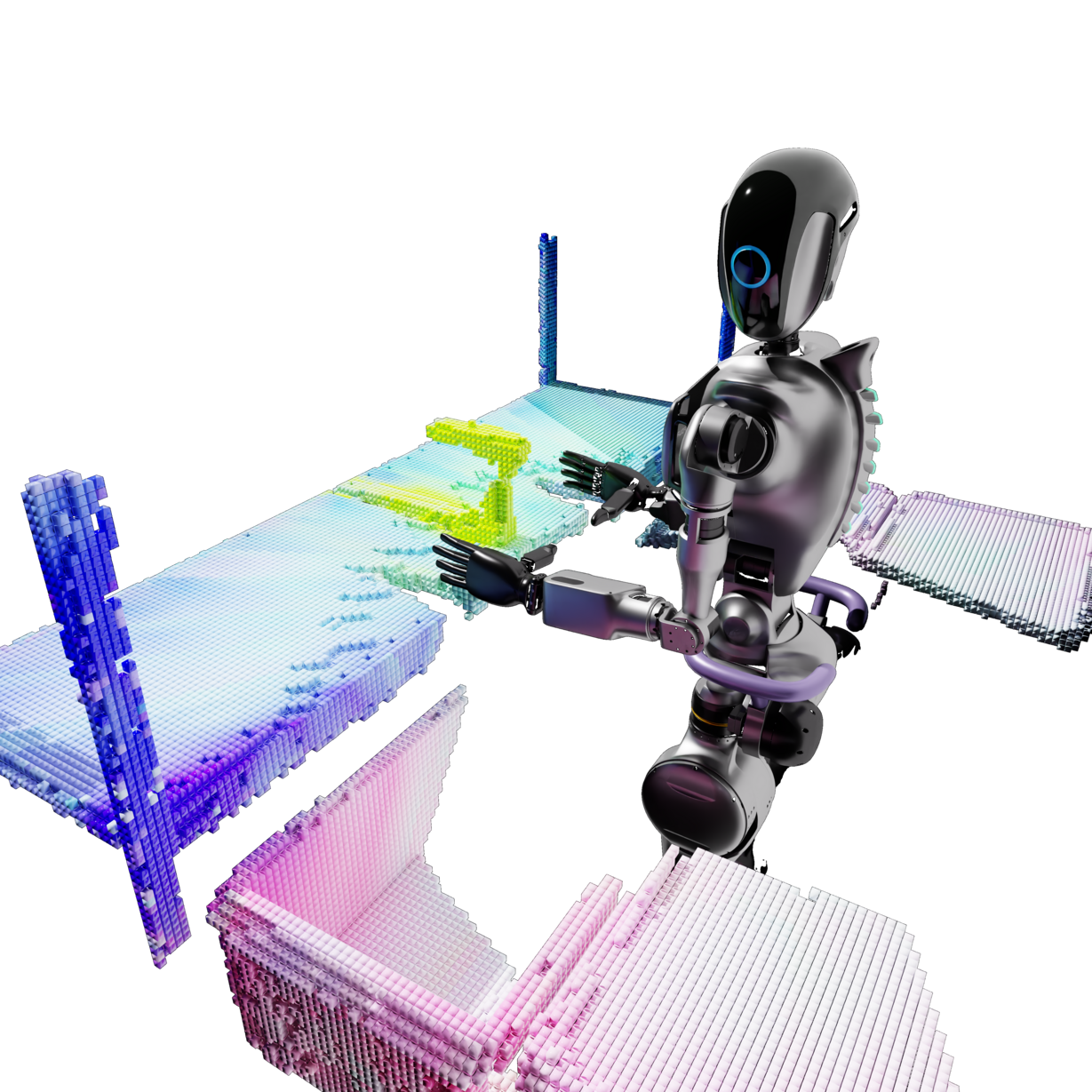}
  }
\caption{Spatial Memory Task~\citep{remo2025}: Humanoid in a simulated industrial space with a metric-semantic reconstruction built by Mindmap (colored by \acs{PCA}).}
\label{fig:mindmap}
\vspace{-2mm}
\end{wrapfigure}

Mindmap \citep{remo2025}, shown in~\cref{fig:mindmap}, contributes tools for extending 3D manipulation policies with spatial memory, including an imitation learning pipeline in Isaac Lab that utilizes a metric-semantic map built using \href{https://nvidia-isaac.github.io/nvblox/pages/torch_examples_deep_features}{\nvbloxtorch}~\citep{millane2024}. The authors demonstrate the efficacy of these tools by extending a state-of-the-art 3D diffusion policy~\citep{3d_diffuser_actor}, and demonstrate significantly improved success rates on challenging manipulation tasks that require spatial memory.

\subsection{Healthcare}

Autonomy in healthcare robotics presents unique challenges compared to other domains, such as manipulation, locomotion, and autonomous driving, due to the need for high-precision, contact-rich interaction with soft tissue and delicate instruments. Progress in this domain has been limited by the absence of domain-specific simulation tools capable of capturing the complexities of clinical environments. Simulation has become central to addressing these gaps, enabling scalable data generation, safe policy training, and sim-to-real transfer. Recent advances have focused on building high-fidelity digital twins tailored for healthcare workflows, accelerating the development of intelligent robotic systems across surgical, diagnostic, and teleoperated applications.

Isaac Lab powers the simulation backbone of Isaac for Healthcare \citep{isaacforhealthcare}, a domain-specific developer framework designed to accelerate the development of intelligent healthcare robotic systems. It addresses key challenges such as simulating anatomical and procedural complexity, integrating clinical data, and supporting robust sim-to-real transfer. High-fidelity digital twins with contact modeling and photorealistic rendering enable scalable training, synthetic data generation, and teleoperation pipelines essential for surgical autonomy.
As part of a broader stack spanning AI model training, simulation, and clinical deployment, Isaac Lab plays a central role in healthcare robotics as it supports digital prototyping, hardware-in-the-loop training, and teleoperation-based imitation learning via XR and peripheral interfaces. %

\begin{figure}
    \centering
    \includegraphics[width=\textwidth]{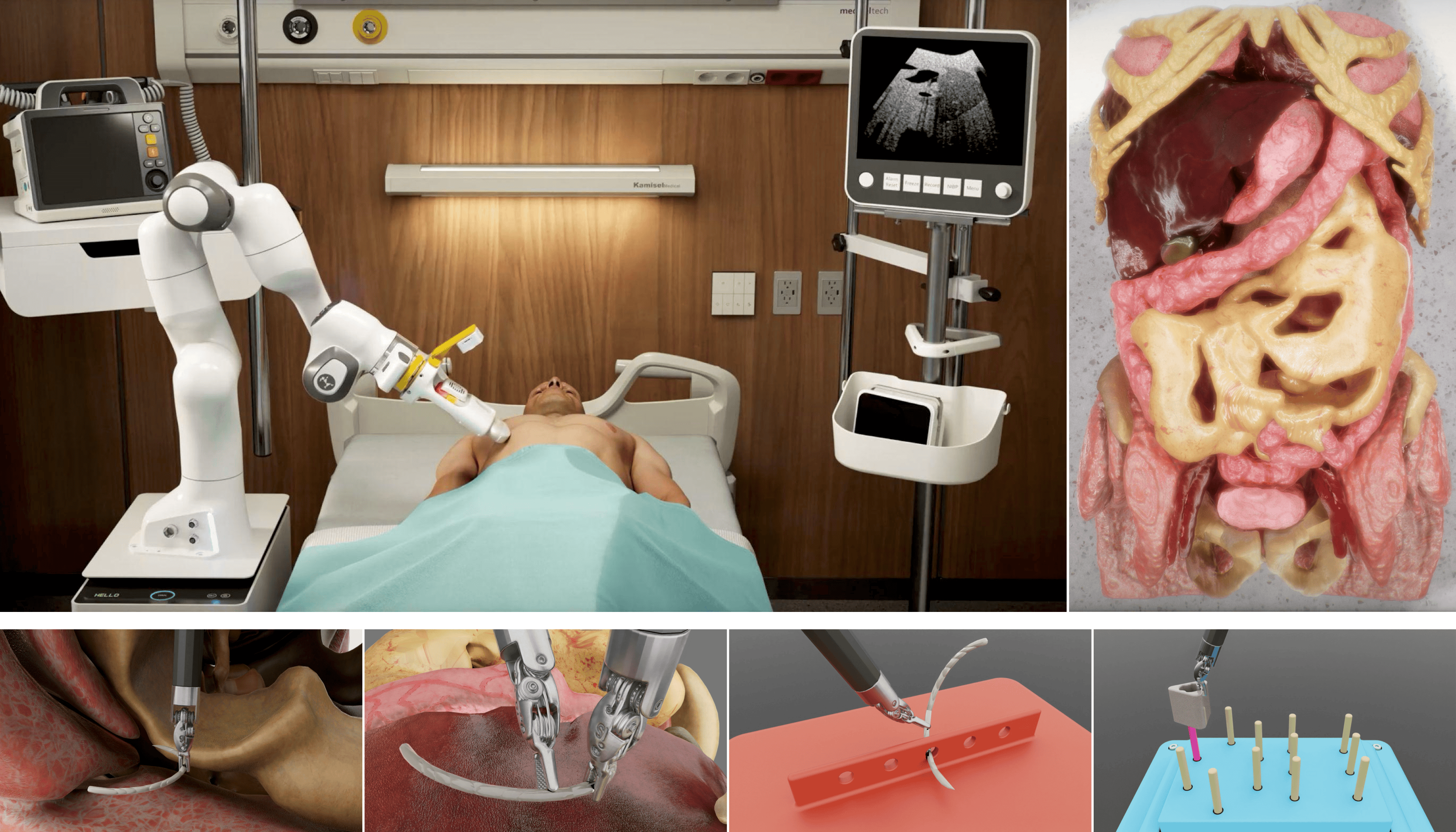}
    
    \captionof{figure}{{Top Row: (Left) Ultrasound simulation in Isaac for Healthcare. (Right) Digital twin of human anatomy in NVIDIA Omniverse. Bottom Row: Training fine-grained robotic maneuvers performed during surgery and the hands-on training exercises used in tabletop surgical curricula.}}
    \label{fig:medical}
\end{figure}

Isaac Lab’s simulation infrastructure is actively enabling key Isaac for Healthcare workflows in robotic surgery, ultrasound, and telesurgery (\cref{fig:medical}). In robotic surgery \citep{moghani2025sufia}, it enables photorealistic, physics-based simulation of surgical tasks such as needle manipulation and tool control. In ultrasound, GPU-accelerated ray tracing simulates acoustic wave propagation to produce realistic B-mode images. For telesurgery, Isaac Lab supports the development and deployment of remote systems across varying network conditions.

\citet{yu2024orbit} build a simulation benchmark for surgical tasks that represent core subtasks in surgical training.
The platform demonstrates successful sim-to-real transfer of learned policies for zero-shot deployment of needle manipulation skills on a physical surgical robot, highlighting its effectiveness in accelerating the development of autonomous surgical systems.
Similarly, \citet{ao2025sonogym} introduce a scalable simulation platform for advancing autonomy in orthopedic surgery. It features anatomically realistic, CT-derived 3D patient models and supports ultrasound simulation using both physics-based and generative methods.
The framework provides a robust testbed for ultrasound-guided navigation and anatomical reconstruction.

\subsection{Generalist Foundation Models}

Foundation models for robotics aim to generalize across tasks, embodiments, and environments by leveraging large-scale, diverse data during pretraining — similar to advances in vision and language domains. GR00T N1~\citep{gr00tn1_2025} is an open foundation model for generalist humanoid robots, built as a vision-language-action model using NVIDIA's Eagle VLM and a Diffusion Transformer (DiT) to integrate visual, textual, and action data. It is pre-trained on a mixture of real robot demonstrations, Internet-scale video, and large-scale synthetic data generated using the Mimic pipeline in Isaac Lab. Synthetic data plays a key role in overcoming the scarcity of real-world data, enhancing robustness and generalization. The model can be post-trained with new demonstrations,  collected or synthesized in Isaac Lab, for task specialization, and can be further improved via sim-and-real co-training~\citep{maddukuri2025simandreal}. The upgraded GR00T N1.5 improves language grounding, generalization, and real-world performance using architectural refinements and the FLARE training technique~\citep{zheng2025flare}, which introduces action prediction and implicit world modeling objectives.
     
Given a foundation model, such as GR00T N1.5, we can perform \textit{online} post-training with \ac{RL}. 
While this is difficult to perform in the real world, Isaac Lab’s high-fidelity rendering and multi-environment setup enables sample-efficient online post-training to be done safely and efficiently in simulation. 
This post-training process can be scaled to a wide range of new tasks with sample-efficient RL techniques~\citep{luo2024serl}, residual RL, and learned reward models for dense feedback during online fine-tuning~\citep{zhang2025rewind}. 
As the field advances towards real-world RL fine-tuning, this simulation-first approach within Isaac Lab is a crucial part of the pipeline for ensuring policies are robust and safe before being deployed on physical hardware.

\section{Future Work and Discussion} %

\begin{figure}[H]
    \centering
    \begin{subfigure}[t]{0.48\textwidth}
        \centering
        \includegraphics[width=\linewidth]{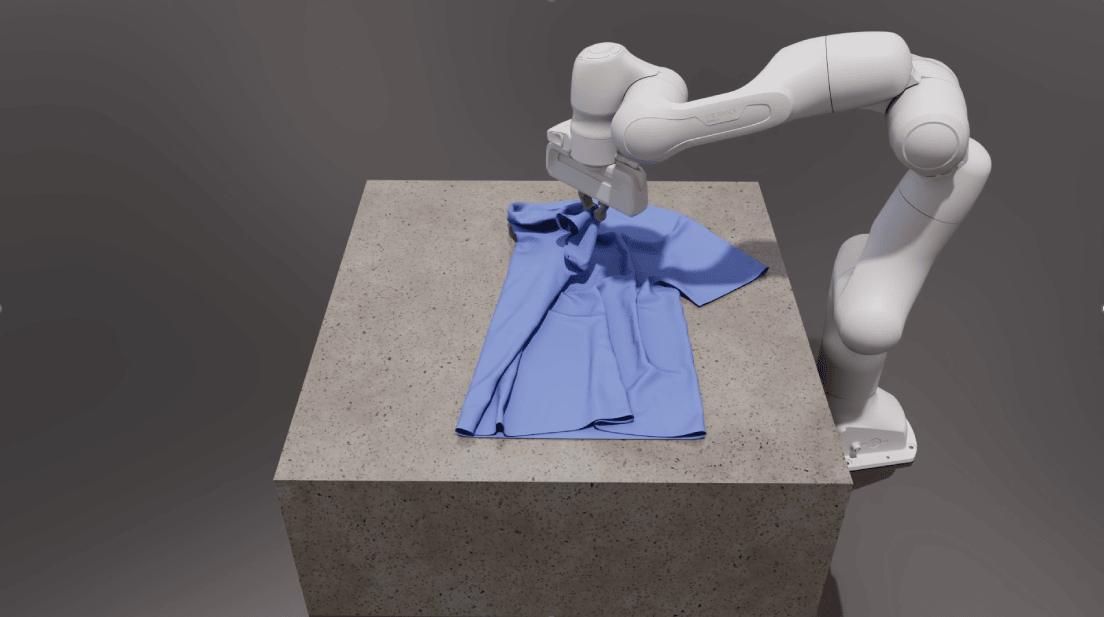}
        \label{fig:newton_cloth}
    \end{subfigure}%
    \quad
        \begin{subfigure}[t]{0.498\textwidth}
        \centering
        \includegraphics[width=\linewidth]{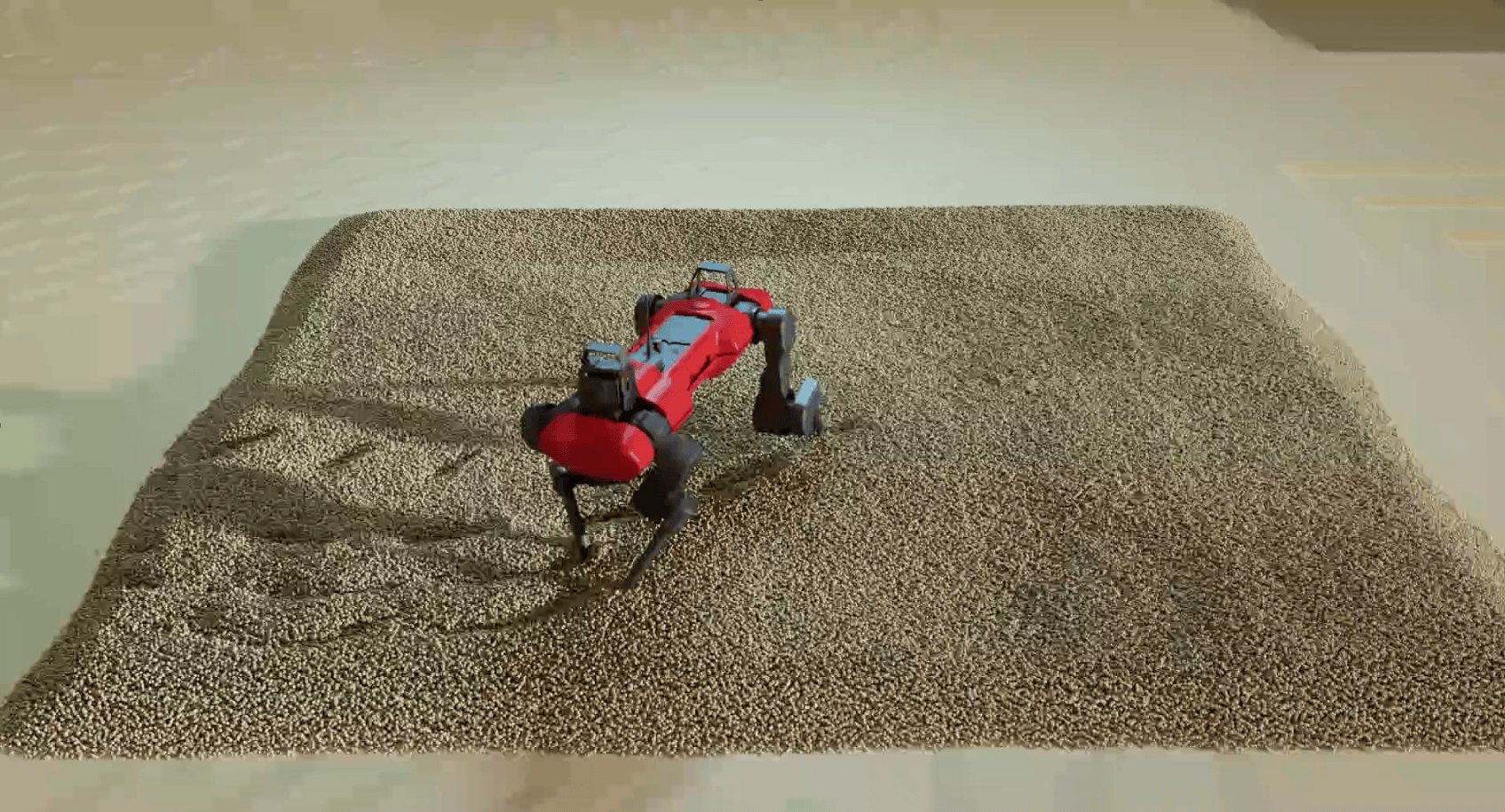}
        \label{fig:newton_mpm}
    \end{subfigure}%
    \caption{%
         Newton supports multi-physics environments by coupling solvers specialized for different types of dynamics. Left: A Franka arm simulated using MuJoCo folds cloth simulated with \ac{VBD} solver\citep{chen2024vertexblockdescent}. Right: An ANYmal robot simulated with PhysX maneuvers through a non-rigid terrain simulated with \ac{MPM} solver\citep{daviet2016mpm}.
         }
    \label{fig:newton_header}
\end{figure}

To further advance Isaac Lab, a major upcoming milestone is integration with the \href{https://github.com/newton-physics/newton}{Newton physics engine} — a GPU-accelerated, extensible, and differentiable simulator offering state-of-the-art solvers for rigid, articulated, and deformable body dynamics (\cref{fig:newton_header}). This addresses key limitations of existing engines in complex robotic scenarios. In addition to physics improvements, future releases will enhance rendering capabilities through deeper integration with NVIDIA’s RTX real-time ray tracing technologies, allowing more photorealistic and physically accurate visuals for vision-based learning. Efforts will also focus on boosting performance and scalability to support large-scale perception training. Additionally, we aim to build a standardized platform for policy evaluation and benchmarking across diverse robotic domains. By incorporating a broad suite of learning-friendly environments, the platform will support rigorous, reproducible evaluation and promote widespread adoption as a unified framework for advancing generalizable robotic policies.

\subsection{Newton Engine and Isaac Lab Integration}
\label{subsec:newton}

Newton is an open-source, GPU-accelerated physics simulation engine explicitly designed for roboticists and simulation researchers. Developed through a collaborative effort by NVIDIA, Google DeepMind, and Disney Research, Newton aims to advance robot learning and development by providing a robust, scalable, and extensible platform for physical AI. Built upon \href{https://github.com/NVIDIA/warp}{NVIDIA Warp}~\citep{warp2022}, a developer framework for accelerating simulation and spatial computing, Newton extends and generalizes Warp's existing simulation functionality, integrating \href{https://github.com/google-deepmind/mujoco_warp}{MuJoCo Warp} as a primary backend. It emphasizes GPU-based computation, differentiability, and user-defined extensibility, facilitating rapid iteration and scalable robotics simulation.

\subsubsection{Key Characteristics and Features}

\begin{figure}[t]
    \centering
    \includegraphics[width=\linewidth, trim={300 100 300 200}, clip]{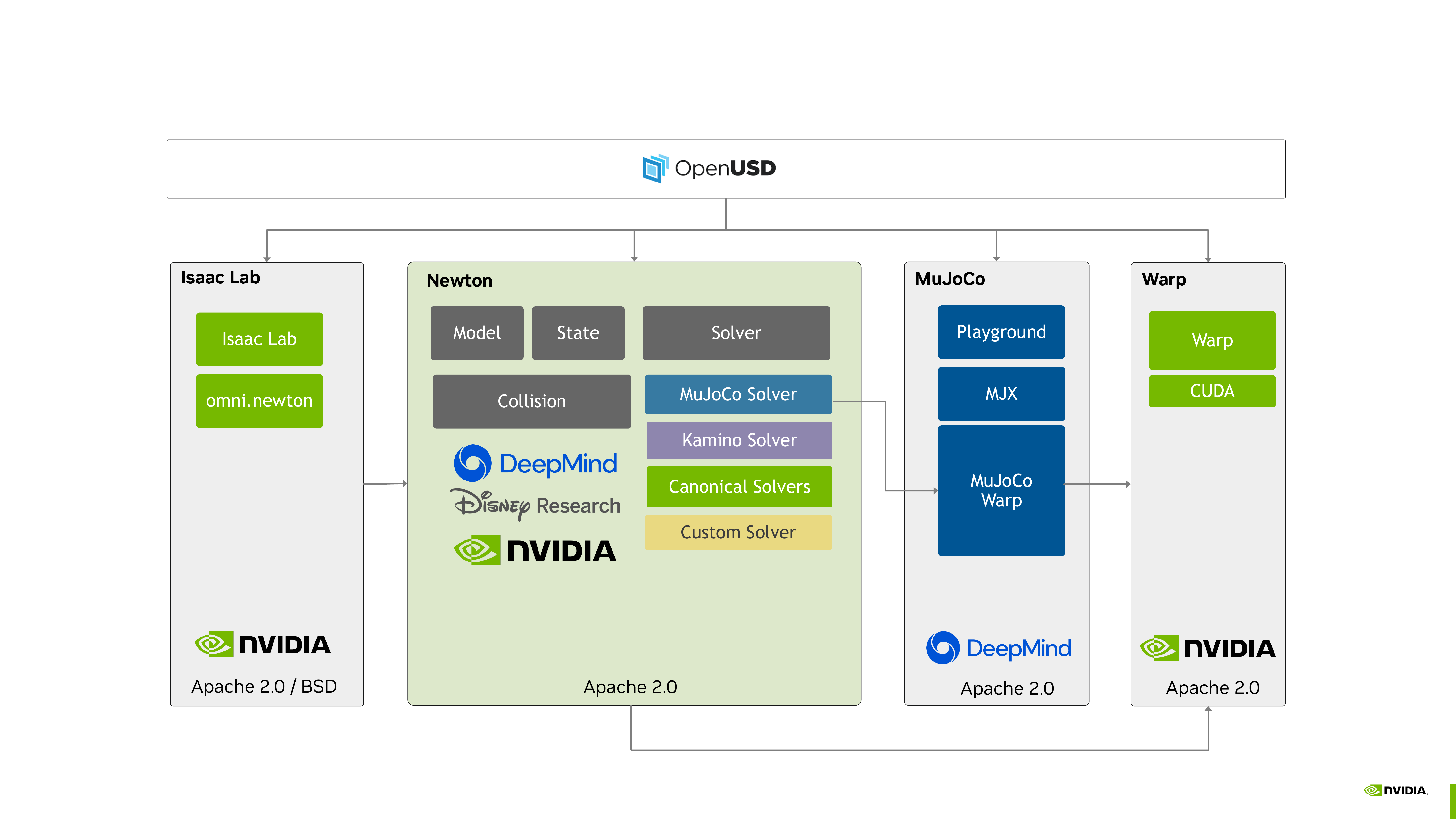}  
    \caption{%
         The Newton Physics Engine is an open-source, GPU-accelerated simulation engine built upon NVIDIA Warp, designed for roboticists and simulation researchers. Its architecture separates the \textbf{Model} (non-time-varying system definition) and \textbf{State} (time-varying physical configuration) for clarity and flexibility, while various \textbf{Solvers} advance the simulation over time by integrating physics. This design, emphasizing flat data structures, enables seamless integration with Deep Learning (DL) frameworks like PyTorch and JAX, and platforms such as Isaac Lab via the \texttt{omni.newton} extension, with key solvers like the MuJoCo Warp built on the same GPU-accelerated foundation. 
    }
    \label{fig:newton_arch}
\end{figure}
Newton is distinguished by several characteristics that cater to the requirements of modern robotics research:

\begin{itemize}
    \item \textbf{Open-Source and Community-Driven}: 
    Newton is an open-source project with source code available on \href{https://github.com/newton-physics/newton}{GitHub} and distributed via PyPI.
    \item \textbf{GPU-Accelerated Performance}: 
    Leveraging NVIDIA Warp and CUDA graphs, Newton delivers high-performance, end-to-end GPU simulations without low-level programming, eliminating CPU bottlenecks common in older physics engines.
    \item \textbf{Extensibility}: 
    Newton’s modular design enables easy integration of custom solvers and models, supporting realistic multiphysics simulations with diverse materials like food, cloth, soil, and cables.
    \item \textbf{Differentiability}: 
    Newton’s automatic differentiation of inputs, controls, and parameters accelerates policy training, design optimization, and system identification for efficient, gradient-based robot learning.
    \item \textbf{Unified API}: Newton offers a single, consistent programming interface for interacting with various physical simulations — including rigid bodies, soft bodies, granular material, and cloth — within the same framework and across multiple solvers.
    \item \textbf{OpenUSD Integration}: The engine utilizes the OpenUSD framework, benefiting from its flexible data model and composition engine to aggregate data describing robots and their environments. 
\end{itemize}

\subsubsection{Architecture and Design Principles}

Newton’s architecture is structured around a clear separation of concerns, and is designed for flexibility and interoperability with DL frameworks.

\begin{itemize}
    \item \textbf{High-level architecture}: Newton organizes simulation around a few core components: \texttt{newton.ModelBuilder} for constructing models, \texttt{newton.Model} for physical structure, \texttt{newton.State} for dynamic data, and \texttt{newton.Solver} for advancing simulation. It supports importing URDF, MJCF, and USD assets via \texttt{newton.Importer} and integrates easily with platforms like Isaac Lab.
    \item \textbf{Separating physical model from numerical method}: Newton separates the time-invariant physical model from the time-stepping solver. This allows different solvers to be applied to the same model, optimizing for various dynamic regimes.
    \item \textbf{Flat data over Object Oriented Programming (OOP)}: Simulation data is represented as tensors and flat arrays, not deep class hierarchies. This design aligns with ML frameworks like PyTorch and JAX, enabling efficient vectorization, JIT compilation, zero-copy interoperability where possible and easy integration into learning workflows.
    \item \textbf{No hidden state}: All internal state is exposed, giving users full control over memory and computation. Solvers act like pure functions — data in, data out — with any state mutation made explicit for clarity and composability.
    \item \textbf{Modular, "take what you need"}: Newton is modular, from low-level geometry up to full solvers. Users can integrate only the components they need, supporting both lightweight prototypes and full production systems.
    \item \textbf{Flexible Selection API}: Similar to PhysX's Tensor API, Newton's Selection API allows creating specialized views of entities in the scene and enables batched, zero-copy access to all attributes in \textbf{Model}, \textbf{State}, and \textbf{Control} objects of scene subsets, such as specific robots or manipulators. It supports named attribute access, joint/link filtering, and custom ordering — making it easy to extend and integrate with ML pipelines and custom solvers.
\end{itemize}

\subsubsection{Solver Implementations}

Newton supports a diverse array of solver implementations including a mix of explicit and implicit methods, as well as reduced and maximal coordinate approaches, each offering unique trade-offs in terms of accuracy, memory usage, and performance.

\begin{itemize}
    \item \textbf{Canonical Solvers}: Out-of-the-box, Newton includes classical solvers such as \textbf{XPBD}~\citep{xpbd2016}, \textbf{Featherstone}~\citep{featherstone1984robot}, and \textbf{Semi-Implicit Euler}. These solvers are generally lightweight implementations of well-established methods as reference for implementers or solver developers.
    \item \textbf{New Solvers}:
    An important new development in Newton is the \textbf{MuJoCo Solver} based on the MuJoCo Warp library, which is a full re-implementation of the MuJoCo~\citep{todorov2012mujoco} algorithms built using NVIDIA Warp. This provides significant performance gains over previous JAX-based MuJoCo XLA (MJX) implementations, particularly for complex scenes involving numerous contacts (\eg dexterous manipulation, humanoid locomotion), without requiring manual contact pruning. In addition, Newton will include a dedicated maximal coordinate solver called the \textbf{Kamino Solver} from Disney Research designed to robustly handle systems with closed loops~\citep{tsounis2025solvingdynamicsconstrainedrigid}.
    \item \textbf{Solver Extensions}: In addition to built-in solvers, many partners are building their simulators on Newton, such as specialized IPC~\citep{Li2020IPC} solvers for tactile manipulation~\citep{li2025taccel}, and specialized solvers for cloth dynamics, \eg\textbf{ Style3D Solver}.    
\end{itemize}

Newton's solver abstraction supports mixed systems encompassing rigid bodies, cloth, and particle  simulations. Multiple solvers can run independently to manage these diverse dynamics, with ongoing development focused on achieving two-way coupling for more intricate interactions.

\subsubsection{Newton USD as a Staging Schema for USD Physics Standardization}
As part of the \ac{AOUSD}'s initiative to advance \ac{USD} as a descriptive language for Physical AI, new schemas are being developed to represent the physical properties of robotic systems in an engine-agnostic manner. This effort is aligned with Newton's solver-abstraction API, which also aims to generalize across simulation backends. Developed alongside the Newton API, the Newton \ac{USD} schema serves as a staging ground where generalized simulation parameters are identified and formalized, with the express goal of proposing them for future inclusion in the \ac{USD} Physics standard.

\subsubsection{Newton Integration with Isaac Lab}

Integration of Isaac Lab with the Newton physics engine is currently underway and accessible on Isaac Lab repository in an \href{https://github.com/isaac-sim/IsaacLab/tree/feature/newton}{experimental feature branch}. This early-stage integration supports a subset of robotics environments, including reinforcement learning tasks for flat-terrain locomotion, manipulation, and vision-based workflows.
\begin{wrapfigure}{R}{0.45\textwidth}
    \vspace{-10mm}
    \includegraphics[width=\linewidth, trim={30, 50, 20, 10}, clip]{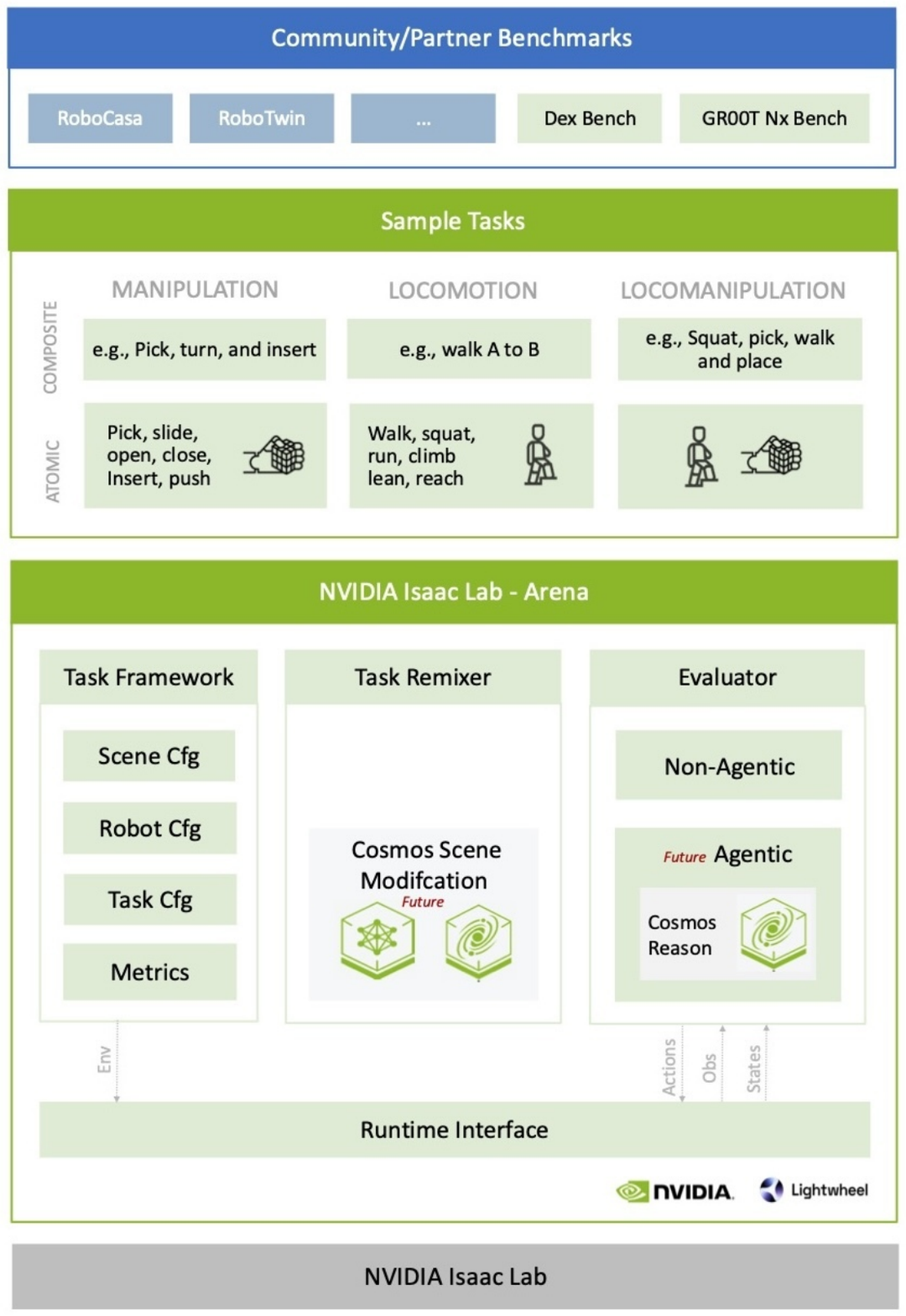}
    \caption{The architecture diagram of \textbf{Isaac Lab - Arena}, a framework and collaborative ecosystem for accessible and scalable policy evaluation in simulation.}
    \label{fig:isaac_lab_policy_eval_framework}
    \vspace{-14.75mm}
\end{wrapfigure}
These implementations serve as testbeds to evaluate Newton’s performance, stability, and physical accuracy through representative robotic workflows.
As part of this integration effort, we have conducted both sim-to-sim comparisons against the existing PhysX backend, as well as sim-to-real validation on physical hardware platforms. These tests aim to characterize policy transferability between simulation engines and real-world deployment.

Ongoing development will focus on expanding Newton support across the full Isaac Lab feature set, including additional sensor and actuator simulation and learning environments, with the objective of achieving feature parity with the existing PhysX backend in the coming months.
This work represents a critical step toward establishing Newton as the default physics engine for high-fidelity and scalable robotics simulation within Isaac Lab. Community engagement is encouraged as we refine this integration and continue to ensure robust support for both research and industrial workflows.

\subsection{Policy Evaluation and Benchmarks}

\subsubsection{Isaac Lab - Arena}

Isaac Lab – Arena is designed to streamline and scale robotic policy evaluation.
While Isaac Lab’s manager-based workflow offers powerful configuration capabilities, achieving end-to-end simulation, from asset preparation to environment setup and large-scale policy evaluation, often requires significant manual effort. This leads to fragmented setups with high overhead, limited scalability, and a steep entry barrier.
Arena introduces a systematic, scalable approach to evaluation, built on Isaac Lab. It provides a robust framework for setting up and executing complex experiments with minimal infrastructure overhead, serving as a launchpad to make advanced simulation-based experimentation more accessible and efficient.

The framework supports simplified, composable task definitions for easy customization and scene diversification, as outlined in \cref{fig:isaac_lab_policy_eval_framework}. It includes extensible libraries for metrics, data collection, and evaluation — starting with rule-based methods and soon expanding to include neural and agent-based approaches. Arena enables parallel, GPU-accelerated evaluations and integrates seamlessly with data generation, training, and deployment frameworks to support closed-loop workflows. It also includes a growing library of sample tasks across manipulation, locomotion, and loco-manipulation.

NVIDIA is actively collaborating with policy developers, benchmark authors, and simulation partners such as Lightwheel to co-develop Arena, use it to accelerate their evaluations, and enable the contribution of their methods and benchmarks back to the community.
Isaac Lab – Arena will be open-sourced on GitHub soon.
    
\subsubsection{Assembly Benchmark}

Isaac Lab provides optimized environments for evaluating and benchmarking robotic manipulation policies on contact-rich tasks. These tasks correspond to a subset of the National Institute of Standards and Technology (NIST) Assembly Task Board 1 \citep{kimble2020benchmarking}. Previous works~\citep{tang2023industreal,tang2024automate,noseworthy2025forge} have demonstrated that the policies trained and evaluated in these environments can be effectively transferred to real robots.
A natural next step is to extend these contact-rich environments to cover the full set of NIST benchmark tasks, further broadening their applicability and impact.

\subsubsection{Dexterous Manipulation Suite}
Isaac Lab currently provides a basic manipulation task suite, comprising of lifting, grasping, and reorienting tasks with the KUKA Allegro hand. These tasks build on prior works from DextrAH \citep{DextrAH-G24, DextrAH-RGB24} and DexPBT \citep{DexPBT_RSS23}. Looking ahead, we plan to significantly expand this suite to include more complex and realistic scenarios across industrial, logistics, and domestic-service domains. In particular, we aim to target humanoid platforms equipped with multi-fingered hands, operating in unstructured home environments — thereby broadening the applicability of Isaac Lab to real-world, high-dexterity tasks.

\section{Conclusion}

Isaac Lab represents a significant advancement in simulation tooling for robotics, unifying high-fidelity physics, scalable rendering, and modular architecture into a single, extensible framework. Built on NVIDIA Isaac Sim and using PhysX and RTX rendering, Isaac Lab delivers high-throughput simulation with support for both direct and manager-based learning workflows. Its key contributions include accurate and performant sensor simulation and actuation modeling, an extensive collection of environments, and tools for motion generation and imitation learning, enabling workflows for training foundation models, RL fine-tuning, and large-scale perception systems. The framework integrates seamlessly with popular learning libraries and adheres to industry-standard APIs, facilitating rapid experimentation and deployment.

Ongoing development of Isaac Lab includes integration with the Newton physics engine, which will introduce even greater physical realism, performance, and flexibility for challenging tasks. Combined with continued improvements to rendering, data generation, and scalability, Isaac Lab is positioned to become a central platform for the next generation of robotics innovation, supporting research at scale and accelerating the path from simulation to real-world deployment.

\clearpage
\appendix
\section{Contributors and Acknowledgments}
\label{sec:contributors}

{
\footnotesize
\setlength{\columnsep}{12pt}
\setlength{\parindent}{0pt}
\linespread{0.96}\selectfont

\begin{multicols}{3}

\subsection*{\normalsize{Core Contributors}}

Mayank Mittal, \textit{NVIDIA, ETH} \\
Yunrong Guo, \textit{NVIDIA} \\
Pascal Roth, \textit{NVIDIA, ETH} \\
David Hoeller, \textit{NVIDIA, Flexion Robotics} \\
James Tigue, \textit{RAI} \\
Antoine Richard, \textit{NVIDIA} \\
Octi Zhang, \textit{NVIDIA} \\
Peter Du, \textit{NVIDIA} \\
Antonio Serrano-Muñoz, \textit{NVIDIA} \\
Xinjie Yao, \textit{NVIDIA} \\
Ren\'{e} Zurbr\"{u}gg, \textit{ETH} \\
Nikita Rudin, \textit{NVIDIA, Flexion Robotics}

\textbf{Physics} \\
Lukasz Wawrzyniak, \textit{NVIDIA} \\
Milad Rakhsha, \textit{NVIDIA} \\
Alain Denzler, \textit{NVIDIA} \\
Eric Heiden, \textit{NVIDIA} \\
Ales Borovicka, \textit{NVIDIA}

\subsection*{\normalsize{Contributors}}

Ossama Ahmed, \textit{NVIDIA} \\
Iretiayo Akinola, \textit{NVIDIA} \\
Abrar Anwar, \textit{NVIDIA, USC} \\
Mark T. Carlson, \textit{NVIDIA} \\
Ji Yuan Feng, \textit{NVIDIA} \\
Animesh Garg, \textit{NVIDIA, GaTech} \\
Renato Gasoto, \textit{NVIDIA} \\
Lionel Gulich, \textit{NVIDIA} \\
Yijie Guo, \textit{NVIDIA} \\
M. Gussert, \textit{NVIDIA} \\
Ankur Handa, \textit{NVIDIA} \\ 
Alex Hansen, \textit{RAI} \\
Mihir Kulkarni, \textit{NTNU} \\
Chenran Li, \textit{NVIDIA} \\
Wei Liu, \textit{NVIDIA} \\
Viktor Makoviychuk, \textit{NVIDIA} \\
Grzegorz Malczyk, \textit{NTNU} \\
Hammad Mazhar, \textit{NVIDIA} \\
Masoud Moghani, \textit{NVIDIA, UofT} \\
Adithyavairavan Murali, \textit{NVIDIA} \\
Michael Noseworthy, \textit{MIT} \\
Alexander Poddubny, \textit{NVIDIA} \\
Nathan Ratliff, \textit{NVIDIA} \\
Welf Rehberg, \textit{NTNU} \\
Clemens Schwarke, \textit{NVIDIA, ETH} \\
Ritvik Singh, \textit{NVIDIA, UC Berkeley} \\
James Latham Smith, \textit{RAI} \\
Bingjie Tang, \textit{NVIDIA, USC} \\
Ruchik Thaker, \textit{NVIDIA} \\
Matthew Trepte, \textit{NVIDIA} \\
Karl Van Wyk, \textit{NVIDIA} \\
Fangzhou Yu, \textit{RAI}

\textbf{Arena} \\
Alex Millane, \textit{NVIDIA} \\
Vikram Ramasamy, \textit{NVIDIA} \\
Remo Steiner, \textit{NVIDIA} \\
Sangeeta Subramanian, \textit{NVIDIA} \\
Clemens Volk, \textit{NVIDIA}

\textbf{Mimic} \\
CY Chen, \textit{NVIDIA} \\
Neel Jawale, \textit{NVIDIA} \\
Ashwin Varghese Kuruttukulam, \textit{NVIDIA} \\
Michael A. Lin, \textit{NVIDIA} \\
Ajay Mandlekar, \textit{NVIDIA} \\
Karsten Patzwaldt, \textit{NVIDIA} \\
John Welsh, \textit{NVIDIA} \\
Huihua Zhao, \textit{NVIDIA}

\textbf{Omniverse} \\
Fatima Anes, \textit{NVIDIA} \\
Jean-Francois Lafleche, \textit{NVIDIA} \\
Nicolas Moënne-Loccoz, \textit{NVIDIA} \\
\begin{CJK}{UTF8}{mj} %
Soowan Park (박수완), \textit{NVIDIA}
\end{CJK} \\
Rob Stepinski, \textit{NVIDIA} \\
Dirk Van Gelder, \textit{NVIDIA}

\textbf{Physics} \\
Chris Amevor, \textit{NVIDIA} \\
Jan Carius, \textit{NVIDIA} \\
Jumyung Chang, \textit{NVIDIA} \\
Anka He Chen, \textit{NVIDIA} \\
Pablo de Heras Ciechomski, \textit{NVIDIA} \\
Gilles Daviet, \textit{NVIDIA} \\
Mohammad Mohajerani, \textit{NVIDIA} \\
Julia von Muralt, \textit{NVIDIA} \\
Viktor Reutskyy, \textit{NVIDIA} \\
Michael Sauter, \textit{NVIDIA} \\
Simon Schirm, \textit{NVIDIA} \\
Eric L. Shi, \textit{NVIDIA} \\
Pierre Terdiman, \textit{NVIDIA} \\
Kenny Vilella, \textit{NVIDIA} \\
Tobias Widmer, \textit{NVIDIA} \\
Gordon Yeoman, \textit{NVIDIA}

\textbf{XR Teleoperation} \\
Tiffany Chen, \textit{NVIDIA} \\
Sergey Grizan, \textit{NVIDIA} \\
Cathy Li, \textit{NVIDIA} \\
Lotus Li, \textit{NVIDIA} \\
Connor Smith, \textit{NVIDIA} \\
Rafael Wiltz, \textit{NVIDIA}

\subsection*{\normalsize{{Leadership}}}

Kostas Alexis, \textit{NTNU} \\
Yan Chang, \textit{NVIDIA} \\
David Chu, \textit{NVIDIA} \\
Linxi “Jim” Fan, \textit{NVIDIA} \\
Farbod Farshidian, \textit{RAI} \\
Ankur Handa, \textit{NVIDIA} \\
Spencer Huang, \textit{NVIDIA} \\
Marco Hutter, \textit{ETH, RAI} \\
Yashraj Narang, \textit{NVIDIA} \\
Soha Pouya, \textit{NVIDIA} \\
Shiwei Sheng, \textit{NVIDIA} \\
Yuke Zhu, \textit{NVIDIA, UT Austin}

\textbf{Physics} \\
Miles Macklin, \textit{NVIDIA} \\
Adam Moravanszky, \textit{NVIDIA} \\
Philipp Reist, \textit{NVIDIA}

\subsection*{\normalsize{{Core Leadership}}}
Yunrong Guo, \textit{NVIDIA} \\
David Hoeller, \textit{NVIDIA, Flexion Robotics} \\
Gavriel State, \textit{NVIDIA}

\end{multicols}
}
\vspace{-10pt}
\noindent\hrulefill

We use the following abbreviations for affiliations:
\begin{itemize}[noitemsep, topsep=0pt, leftmargin=*]
\item \textbf{ETH} – Swiss Federal Institute of Technology, Z\"{u}rich, Switzerland
\item \textbf{RAI} – Robotics and AI Institute, Cambridge, Massachusetts, United States and Z\"{u}rich, Switzerland
\item \textbf{MIT} – Massachusetts Institute of Technology, Cambridge, Massachusetts, United States
\item \textbf{UC Berkeley} – University of California, Berkeley, United States
\item \textbf{USC} – University of Southern California, Los Angeles, United States
\item \textbf{UT Austin} – The University of Texas at Austin, Austin, United States
\item \textbf{UofT} - University of Toronto, Toronto, Canada
\item \textbf{GaTech} - Georgia Institute of Technology, Atlanta, United States
\item \textbf{NTNU} - Norwegian University of Science and Technology, Trondheim, Norway
\end{itemize}

\noindent\hrulefill

The roles are defined as follows:

\textbf{Core Contributor:} Individuals who had a very significant and sustained impact throughout the project, across multiple subsystems, with accountability for key design and implementation decisions. Core contributors are listed in order of merit.

\textbf{Contributor:} Individuals who made important contributions to the project. Contributors are listed in alphabetical order. Contributions include:
\begin{itemize}[noitemsep, topsep=0pt]
    \item Major code and documentation contributions to the main codebase
    \item Substantial engineering or production work on key underlying simulation technologies, such as the PhysX and Newton physics engines, the Omniverse RTX rendering system and related technologies, or Isaac Sim
    \item Authorship of research papers referenced in this white paper whose individual contributions significantly influenced the design and implementation of Isaac Lab
    \item Authorship of significant portions of this white paper
\end{itemize}

\textbf{Leadership:} Individuals who contributed to the organizational and technical direction of efforts related to Isaac Lab. Leadership members are listed in alphabetical order.

\textbf{Core Leadership:} Individuals with primary responsibility for the organizational and technical direction of Isaac Lab. Core leadership members are listed in alphabetical order.

\noindent\hrulefill

\subsection*{Acknowledgments}

The development of Isaac Lab initiated from the Orbit framework~\citep{mittal2023orbit}. We gratefully acknowledge the authors of Orbit for their foundational contributions.

Isaac Lab relies heavily on core technology from NVIDIA Omniverse and Isaac Sim, such as the Omniverse RTX Renderer and USD conversion tools for robot description formats.
While certain NVIDIA staff members have contributed directly to the Isaac Lab project and are recognized above as contributors, we express our deep appreciation for everyone involved, including members of the open-source community working on connected projects such as Open USD.

We further thank the open-source community for their active engagement in the development of Isaac Lab. The project has benefited from code contributions, issue reports, documentation improvements, and valuable feedback from users worldwide. A complete and regularly updated list of contributors is available on \href{https://github.com/isaac-sim/IsaacLab/graphs/contributors}{Isaac Lab GitHub repository}.

We are especially grateful to Philip Arm, Arjun Bhardwaj, Filip Bjelonic, Nicola Burger, Rafael Cathomen, Ioannis Dadiotis, Clemens Eppner, Oliver Fischer, Per Frivik, Caelen Garrett, Ayush Ghosh, Pulkit Goyal, Tairan He, Matthias Heyrman, Jason Jingzhou Liu, Victor Klemm, Chenhao Li, Zichong Li, Tyler Ga Wei Lum, Zhengyi Luo, Gary Lvov, Denys Makoviichuk, Takahiro Miki, AJ Miller, Kyle Morgenstein, \"{O}zhan \"{O}zen, Aleksei Petrenko, Tifanny Portela, Jean-Pierre Sleiman, Jonas Stolle, Balakumar Sundaralingam, Lorenzo Terenzi, David Tingdahl, and Fan Yang for their research contributions and constructive feedback.

Finally, we acknowledge the use of Large Language Models (such as OpenAI GPT-5), which helped to refine the manuscript language.
All technical content, analyses, and citations were generated and verified by the authors.
We note that large language models may exhibit biases, errors, or omissions, and we take full responsibility for the accuracy and appropriateness of the manuscript.

\clearpage
\section{Acronyms}
\label{sec:acronyms}

\begin{acronym}[AOUSD] %
    \acro{AEC}{Agent Environment Cycle}
    \acro{AOUSD}{Alliance for OpenUSD}
    \acro{API}{Application Programming Interface}
    \acro{CPU}{Central Processing Unit}
    \acro{DLSS}{Deep Learning Super Sampling}
    \acro{DoF}{Degrees-of-Freedom}
    \acro{DR}{Domain Randomization}
    \acro{FEM}{Finite Element Method}
    \acro{IK}{Inverse Kinematics}
    \acro{IL}{Imitation Learning}
    \acro{GPU}{Graphics Processing Unit}
    \acro{LiDAR}{Light Detection and Ranging}
    \acro{MARL}{Multi-Agent Reinforcement Learning}
    \acro{MDL}{Material Definition Language}
    \acro{MJCF}{MuJoCo Modeling XML File}
    \acro{MPM}{Material Point Method}
    \acro{PBD}{Position-Based Dynamics}
    \acro{PBT}{Population-Based Training}
    \acro{PCA}{Principal Component Analysis}
    \acro{RL}{Reinforcement Learning}
    \acro{SDF}{Simulation Description Format}
    \acro{SDG}{Synthetic Data Generation}
    \acro{URDF}{Unified Robot Description Format}
    \acro{USD}{Universal Scene Description}
    \acro{VBD}{Vertex Block Descent}
    \acro{XR}{Extended Reality}
\end{acronym}

\clearpage
\setcitestyle{numbers}
\bibliographystyle{plainnat}
\bibliography{main}

\end{document}